\documentclass{article}
\usepackage{arxiv}
\usepackage[utf8]{inputenc}
\usepackage[T1]{fontenc}  
\usepackage{amssymb,mathtools}
\usepackage{algorithm}
\usepackage[noend]{algpseudocode}
\usepackage{bookmark}
\usepackage{subcaption}
\usepackage{lineno,hyperref}
\usepackage{url}            
\usepackage{booktabs}       
\usepackage{amsfonts}       
\usepackage{nicefrac}       
\usepackage{microtype}      
\usepackage{cleveref}       
\usepackage{lipsum}         
\usepackage{graphicx}
\usepackage{doi}
\usepackage{longtable}

\DeclareMathOperator{\ODESOLVE}{\textit{ODESolve}}

\def\BibTeX{{\rm B\kern-.05em{\sc i\kern-.025em b}\kern-.08em
		T\kern-.1667em\lower.7ex\hbox{E}\kern-.125emX}}

\title{Recent Trends in Modelling the Continuous Time Series using Deep Learning: A Survey.}

\author{ \href{https://orcid.org/0000-0001-9051-1370}{\includegraphics[scale=0.06]{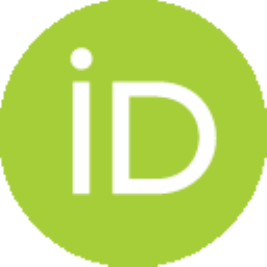}\hspace{1mm}Mansura Habiba}
	PLatform Archietct\\
	IBM in Ireland\\
	Dublin, Ireland \\
	\texttt{mansura.habiba@gmail.com} \\
	\And
	\href{https://orcid.org/0000-0003-0521-4553}{\includegraphics[scale=0.06]{orcid.pdf}\hspace{1mm}Barak A. Pearlmutter}\\
	Professor \\
	Department of Computer Science \& Hamilton Institute\\
	Maynooth University\\
	Maynooth, Ireland \\
	\texttt{barak@pearlmutter.net} \\
	\AND
	\href{https://orcid.org/0000-0002-4029-8409}{\includegraphics[scale=0.06]{orcid.pdf} \hspace{1mm}Mehrdad Maleki}\thanks{Maleki was supported by the Irish Research Council (IRC) grant Goipd/2019/803.}\\ 
	Postdoctoral Research Fellow\\
	Department of Computer Science\\
	Maynooth University\\
	Maynooth, Ireland \\
	\texttt{mehrdad.maleki@mu.ie} 	\\
}

\begin{document}
	\maketitle

\begin{abstract}
Continuous-time series is an essential component for different modern days application areas, e.g. healthcare, automobile, energy, finance, Internet of things (IoT) and other related areas. Different application needs to process as well as analyse a massive amount of data in time series structure in order to determine the data-driven result, for example, financial trend prediction, potential probability of the occurrence of a particular event occurrence identification, patient health record processing and so many more. However, modeling real-time data using a continuous-time series is a very challenging task since the dynamical systems behind the data could be a differential equation. Several research works have tried to solve the challenges of modeling the continuous-time series using different neural network models and approaches for data processing and learning. The existing deep learning models are not free from challenges and limitations due to diversity among different attributes, behaviour, duration of steps, energy, and data sampling rate. In this paper, we have described the general problem domain of time series and reviewed the challenges in terms of modelling the continuous time series. We have presented a comparative analysis of recent developments in deep learning models and their contribution in order to solve different challenges of modeling the continuous time series.  We have also identified the limitations of the existing neural network model and open issues. The main goal of this review is to understand the recent trend of neural network models used in a different real-world application with continuous-time data.

\end{abstract}



\section{Introduction}
Deep learning algorithms have demonstrated impressive performance in terms of the continuous-time series modelling. At the same time, the sequence-based data model, for example, time series has become trendy in different areas of industry and science.  Recent deep learning model architectures are focused on learning time sequence modelling problem in order to provide an optimised, efficient and accurate result for real-time data in several applications, such as financial trading, utility such as energy as well as water distribution, natural language processing, electronic health record analysis, healthcare, human activity recognition and prediction, robotics and other modern research areas for artificial intelligence. The current trend shows significant popularity of deep learning in order to solve continuous-time series learning problems. However, there is not enough research tackling the problems of dealing sparse and asynchronous time sequence, processing multi-variate time series with a massive number of parameter and dealing with dynamic, heterogeneous and uncertain nature of the time series. Besides, there is still scope for improvement in terms of optimisation and accuracy precision. 

Time series modelling is very challenging due to the unique nature of the problem domain as well as the data structure. Some of the challenges for continuous-time series processing using deep learning are that (i) the data can be collected at a different rate, (ii) this variable sampling rate needs a very complex deep learning architecture to learn. Based on the problem domain, the data collection as well as data sampling rate can be too irregular. For example, in the case of collecting data from different sensors in the Internet of Things (IoT) framework, the sampling rate is irregular. Additionally, often, the continuous-time series problem has a considerable number of parameters in order to provide a real-time result. Some of these parameters are even unknown, uncertain as well as dynamic; therefore, sometimes, they are even inaccessible. An efficient neural network model architecture needs to be well equipped to deal with the numerous amounts of parameters and high dimensional data. Another limitation of time series modelling is that existing models are suitable for the static problem domain, but real-time time series is noisy as well as dynamic. Section~\ref{sec:challenges} describes the main challenges for processing time series using deep learning algorithms.

In most cases, existing deep learning models fail to collect real-time data due to the complex as well as higher-order temporal dynamics. In order to avoid the challenges of dynamic input size and variable sampling rate, most deep learning research works mainly focus on fixed-size inputs along with consistent data format. However, to get a better as well as an efficient result for a dynamical system with uncertain nature in the era of Internet-of-Things (IoT), it is essential to improve the performance of deep learning algorithm. Due to these limitations, a recent survey \cite{langkvist2014review} has revealed that the trend in future research is mainly focused on developing a Deep model for static data rather than continuous-time series data. 

Time series is complicated and challenging to process, but it is essential for essential research fields, such as finance, health, IoT, weather, climate, cloud and others. Besides, the recent increase in software application with event-driven architecture, time series is a mandatory component for our everyday software. Continuous-time data hold extreme potential. For example, \cite{helbing2015saving} describes the design of a complex system to understand the data-flow of crime, terrorism, wars, and epidemics and other crowd disasters using continuous-time data. This work demonstrates the potential of real-time data analysis to address a number of serious issues that are continuously disrupting human life over the last decades.  Let's consider the example of the automobile industry. Self-driving cars are already on the street. Alexa \cite{10.5555/3137257} is analysing our daily activity in real-time. Systems surrounding us are getting increasingly smart and autonomous. Our dependence on this autonomous system is also increasing significantly. Its no longer only using a robot vacuum to clean the floor, people are commuting on self-driven cars on a busy road or undergoing surgery performed by robots. As the applications of autonomous objects are increasing widely, and scientists can no longer settle with the result of approximate accuracy; they need real-time as a well accurate result.        

However, recent research works focused on learning unlabelled real-time data has achieved a better result with more accuracy. Learning with unlabelled data also helps to learn much more information about the data and can be used to draw a better outcome. Therefore, learning data using a continuous-time series can achieve a better result. However, as for different unique feature in time series data, the data have to process, and the corresponding model needs specific features in order to adjust the time series along with its characteristics. This article discusses the pipeline of designing the model for the continuous-time data. 
Recent research trends are focusing on all these limitations in pursuit of finding an efficient and optimised solution. A recent work, Phased-LSTM \cite{neil2016phased} and some of its variant \cite{ehr,timelstm}, Skip RNN \cite{skiprnn}, Spatio-Graph-based RNN \cite{structuralrnn,fcnrlstm} have achieved substantial improvement in result in terms of overcoming the challenges of missing pattern and irregular sampling rate. Table \ref{tab:time-series} lists several recent research works with excellent performance in solving different challenges. A new family of the neural network \cite{chen2018neural} leveraged the capabilities of the differential equation to compute instant changes for any dynamical system.  

Different types of deep learning neural network model have been researched and analysed in order to solve different challenges in terms of modelling time series modelling.  Among all different models, the RNN is a pioneer of choice for modelling temporal process \cite{neil2016phased, ehr, timelstm}, however, at the same time, RNN has its pitfalls when modelling time-series. Each model differs in structure, design, performance and learning mechanism. Each of them has it's respective strengths and weaknesses for learning continuous-time data. Besides, the same model is not suitable for all kinds of time series. Some neural network models provide a better result for irregular data points where other models are more suitable for fixed time rate time series. Again, a supervised learning model works better with CNN, AWNN neural network models, while unsupervised learning mechanism is best-suited with RNN models. 

This survey is an attempt to identify the challenges in different research fields with continuous-time data and how different aspects of Deep learning algorithm can approach these challenges to overcome.   This paper focuses on the following questions, 
\begin{enumerate}
\item What are the challenges for continuous-time data processing evolve with emerging technologies and research field, e.g. IoT, healthcare and others?
\item How the state of the art deep learning algorithms are closing the gap to mitigate existing challenges?
\item How is continuous-time data processing using deep learning affecting research in different real-life applications?
\item What are some recent promising neural network for processing continuous-time data?
\item How ordinary and partial differential equations are changing the behaviour of neural network?
\end{enumerate}

We review the recent trend in using deep learning algorithms for modelling continuous sequential data. We also taxonomise different challenges and identified the state of art solutions for existing limitations in terms of modelling a continuous-time series. We studied the problem domain in order to determine the main challenges in modelling the continuous-time series. Another significant contribution of this paper is that it discusses a comparative analysis of some of the promising recent researches corresponding to address challenges regarding computing efficiency in processing continuous-time series. We also discuss the current trends of deep learning algorithms in different real-time applications.
The rest of the paper is organised as follows. §\ref{sec:characteristics} describes the properties of time series and different aspect of the problem domain of continuous-time series learning using deep learning. Section  §\ref{sec:domain} discusses different types  of a continuous series of data in. Comparative analysis of some existing neural network architectures and algorithms is described in §\ref{sec:models}. §\ref{sec:challenges} analyses the existing challenges for modelling time sequences along with several recent research in order to overcome those challenges. In §\ref{sec:application}, a wide range of applications of the continuous-time series is discussed. Some future research area regarding time series modelling as exhibited  in Section §\ref{sec:future}.

\section{Time Series Analysis} 
\label{sec:characteristics}
Due to the dynamic nature of the time series, learning the continuous-time data with deep learning algorithms is a very complex task. Moreover, efficient and precise computation is even more challenging. One of the fundamental behavior is that continuous time series vary in length, and the complexity of learning such a long time series increases with the increase of the length. Besides, there is an implicit dependency between different time states in a series. The previous input and past computation have a significant role in calculating the current or any future state in a time series. Therefore, most of the time, there is a requirement for memorizing the previous state in the current state. As a result, a traditional feed-forward neural network, where input at each state is independent of each other is not capable of learning time series. Therefore, it is essential to understand the different dynamic and specific nature of the continuous-time series to design a neural network model. In this section, we discuss those unique characteristics of a time series.
\subsection{What is the time series?}
\begin{figure}[h]
\centering\includegraphics[width=1\linewidth]{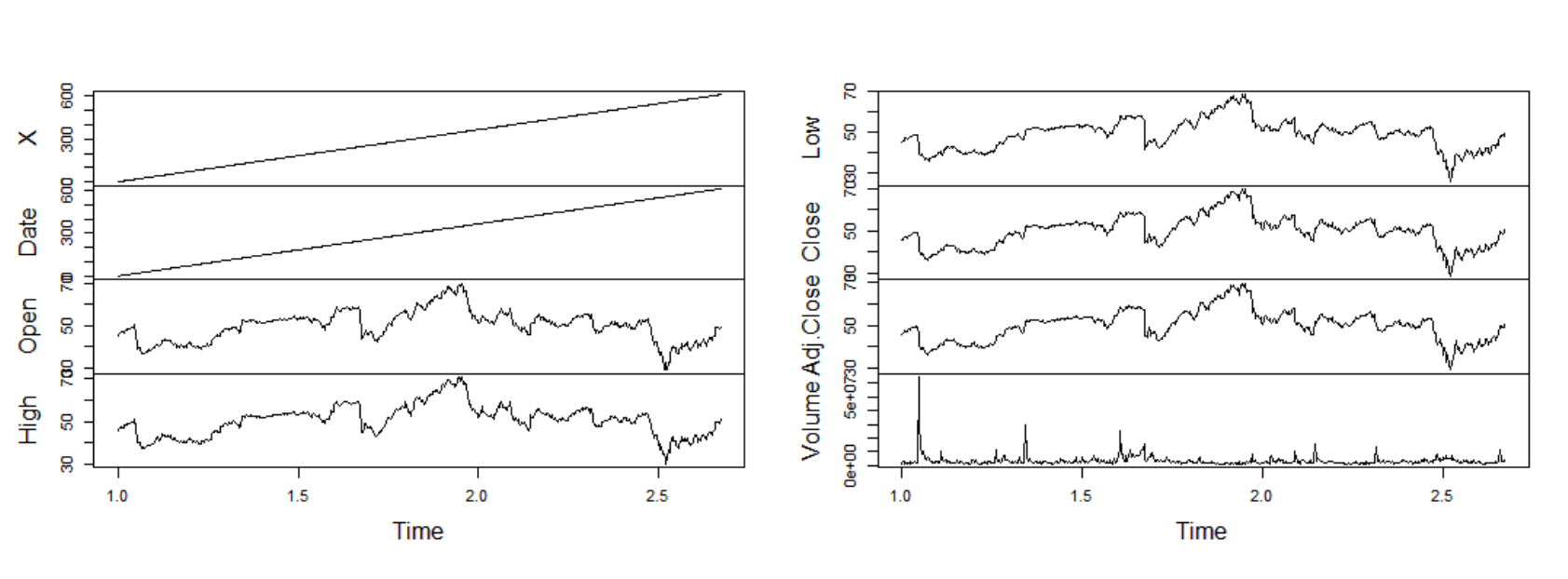}
\caption{Continuous time series for stock price of Dell company from 2 January 2018 until 5 June 2020} \label{fig:continuous-time-series}
\end{figure}

Time series is as a sequence of numeric observations of a variable at a continuous time. For example, Fig.~\ref{fig:continuous-time-series} shows the observation of certain variable, i.e.  stock price of Dell over a definite time period. More precisely, a time series is a function $x:T\to \mathbb{R}^N$ , where $T\subseteq \mathbb{R}$, with a probability distribution $\mathcal{P}$, i.e., $(x_t)_{t\in T}\sim \mathcal{P}$. If $t\in T$ then $x(t)$ or $x_t$  refers to the observed value at time $t$. If $T$ is a finite set we have definition of discrete time series. If $T=\{t_1,\dots,t_n\}$ with $t_{i+1}-t_i=h$ then $(x_t)_{t\in T}$ generates a discrete time time-series. If $T=[a,b]$ then a continuous time time-series \cite{beran2018mathematical} is produced. 

For example, any variable $x$ can have value as $x_1, x_2, \ldots, x_n$ at a different time step, such  as $t_1, t_2, \ldots, t_n$. Eq~(\ref{eq:Xn}) shows  time series as a vector \cite{doucoure2016time}, where each value $x_t \in \mathbb{R}$. Each entry in the $X$ vector is the real-time value of a time-dependent variable measured at a corresponding time $t$ of a continuous or discrete-time set $T$, i.e., $x_1=x(t_1),x_2=x(t_2),\dots$, where $t_1,t_2,\dots \in T$.

\begin{equation} 
  X_n=(x_1,x_2,\ldots,x_n)
  \label{eq:Xn}
\end{equation}

\subsection{What are the properties of a continuous-time series?}
Continuous-time data has it’s unique properties. Some of them are listed as follows

\begin{itemize}
\item  \textbf{Length}: Different research application such as automobile, health, they need to observe the value for any variable for a very long period to get correct data. Therefore, most continuous Real-time data is usually very long. For example, weather data analysis for ten years, financial trend analysis for the last decade, unemployment statistics analysis, event data from sensors for a self-driving car for one hour. For accurate analysis in any type of time series problem including classification, anomaly detection, synthetic generation, it is important to train data for a long duration. 
\item	\textbf{Higher Sampling Frequency}: For a continuous time series, the intermediate time step is short, and the sampling frequency is much higher. For example, existing recurrent neural network (RNN) models use a fixed time step, and instead of modeling a continuous time series, these models first convert the time data to discrete-time time series data and then process it with a higher sampling rate. As a result, for a long time series, the model chooses a comparatively small fixed time step but a high data sampling rate, for better accuracy. 
\item	\textbf{Pathological dependencies}: One of the main goals for time series modeling using deep learning algorithms is to identify underline temporal relation of consecutive data points in order to detect the pattern. In the case of most time-series, there are dependencies between consecutive data points.  This dependency can be either implicit or explicit. Another related property of continuous-time series is that the length of mutually dependent sequence length is unknown. For example, $x_{t}$ can have a dependency on previous observations at a very distant past($x_{t-n}$)  or a very short distance past($x_{t-1}$)  .
\item 	\textbf{Higher Dimension}: Every data point $x_{t}$ in Eq.~\eqref{eq:Xn} consists of thousands of attributes. Multiple parameters are a common feature for continuous real-time data. For example, $x_{t}$ in Eq.~\eqref{eq:Xn} can be rich in dimensionality with a large number of parameters in order to describe an individual observation. Multi-variate time series shows very high dimensionality. The value for x at time $t$ does not only depend on time $t$ but also depend on other variables(multivariate time series), i.e., if $\mathbf{X}(t)=[x_1(t),\dots,x_N(t)]$ then $\mathbf{X}(t)=A_1\mathbf{X}(t-1)+\dots +A_p\mathbf{X}(t-p)+\mathbf{b}+\mathbf{\varepsilon}(t)$ in its simplest form, where $A_i$ are $N\times N$ autoregressive matrices and error term $\mathbf{e}$ and $(x_i(t))_{t\in T}$ for $i=1,\dots, N$ are different time series \cite{box2011time}.
\item 	\textbf{Additional Noise}: Along with the real-time value of any observations $x_t$ at a time t, continuous-time also contains excessive noise components for the entire series. 
\item	\textbf{Higher Energy Consumption}: The dimension of a continuous-time series is often very long, which requires extensive computation and thereby energy.
\item	\textbf{Uncertainity}:  Continuous-time series is often very dynamic and uncertain with the missing pattern as well as irregularities. Therefore, it is often impossible to model the complete time sequence. There are different well-known data imputation techniques are used to replace the missing values in a time series.
\item	\textbf{ Irregular Sampling Rate }:  Besides having a higher sampling rate, continuous-time series also exhibits an irregular sampling rate. The irregular sampling rate, high complexity, and massive length of real-world continuous time series often cause missing data point that ultimately affects the result of neural network negatively. It is not possible to process an incomplete model precisely.  An irregular sampling rate often causes missing observations. The value of observation x can be missing at any time step t in the continuous-time series. Not every time step has enough information describe the observed value for x at time t.
\item	\textbf{ Memory}:  Time-series needs memory to encapsulate information from previous steps. Sometimes, the interrelation between different steps in the time sequence. 
\end{itemize}

In short, continuous-time data is dynamic, enormous in amount, unique, sophisticated, and high dimensional. The characteristics of different time series are extensively dynamic. All these properties of continuous-time series make it very challenging to process. However, in the recent era, several research works have set in and achieved a significantly improved result with continuous-time data. Therefore, there is a recent trend to design the deep learning models to process continuous-time data without converting the continuous-time series to discrete-time series.  In this paper, we discussed how recent deep learning models are designed to adjust to the above-mentioned characteristics of continuous-time data. The architecture of deep learning models has to consider all these unique behaviors of time series in order to achieve a  better result with higher precision of accuracy. The complexity, training data requirement, computation power all can vary a lot based on the characteristics of time series and the architecture of the neural network model.

\section{Problem Domain Analysis} 
\label{sec:domain}
\begin{figure}[h]
\centering\includegraphics[width=0.9\linewidth]{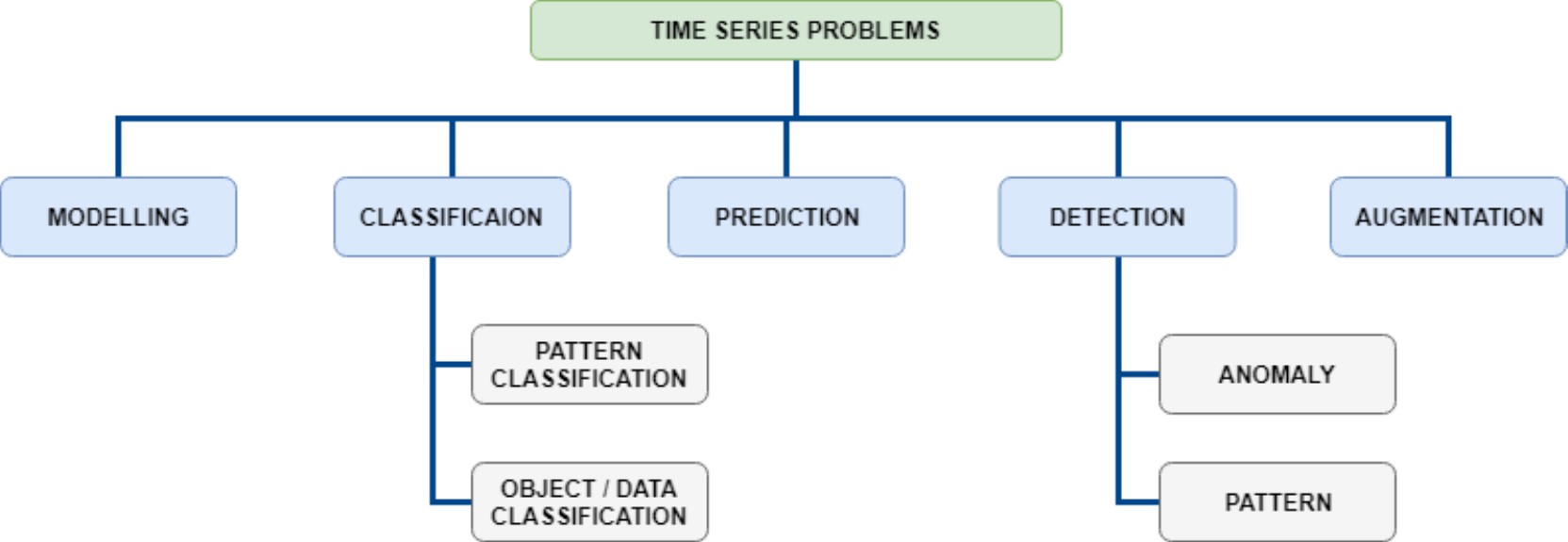}
\caption{Different category of time series problems}\label{fig:category} 
\end{figure}

Time series related problems are not new, and the categories of these problems are wide. Primarily, these problems can be  classified into four categories \cite{ref13},as shown in figure [\ref{fig:category}]. Again Modeling can be either sequence-to-sequence mapping or augmentation. Time series classification has a wide range of applications, as shown in Table \ref{tab:time-series-problems}. Classification problem can be either related to the classification of a pattern embedded in the data or classification of continuous-time series data such as video, text, wavelet and others. Prediction probably the most popular problem type in the domain of continuous time series. Besides these, sometimes detection of embedding pattern in the continuous-time data are also essential problem category. Another kind of problem in time series is anomaly detection. There is no abundant number of work in this area of time series problem domain. \cite{feng2015novel}  proposed an anomaly detection mechanism using low-level tracking algorithms and  \cite{gershenfeld1993future} highlights the following questions that need to be solved for deep learning to be ready to provide accurate results.

\begin{itemize}
	\item	How does understanding (explicitly extracting the geometrical structure of a low-dimensional system) relate to learning (adaptively building models that emulate a complex system)? 
	\item	When a neural network correctly forecasts a low-dimensional system, it has to have formed a representation of the system. 
	\item	What is this representation? 
	\item	Can the representation be separated from the network’s implementation? 
	\item	Can a connection be found between the entrails of the internal structure in a possibly recurrent network, the accessible structure in the state-space reconstruction, the structure in the time series, and ultimately the structure of the underlying system? 
\end{itemize}

It is fascinating to say that most of the questions are now known. As an answer to the third question, the time series is represented using a graph or  N $\times$ M vector as the input of a neural network. The neural network proposed in  \cite{structuralrnn,fcnrlstm} describes a neural network model where the structure of space, as well as the structure of the time series, are considered for neural network model architecture. These works are practical examples to answer the fifth question mentioned above. Spatio-temporal-graph \cite{structuralrnn} has become a popular representation of space and time sequence.

The unique behaviour of continuous-time series attracts the attention of deep learning researchers in the early ’90s.  Table \ref{tab:time-series-problems} shows some related works in every type of continuous-time series-based problem. In this paper, we have evaluated recent works in mainly the last decade. \\

There is no single solution for each of the above-mentioned problem category. Different types of neural network models are suitable for different problems types. For example, Artificial wavelet neural network (ANN) \cite{doucoure2016time} is suitable for modelling and prediction of continuous time series. On the other hand,  RNN pioneer in time series modelling, classification and prediction problem domains due to it’s a suitable set of properties. For time series classification, CNN is mostly used, model. CNN can learn classes from a continuous-time series through unsupervised learning with minimum human interaction.
In most cases, CNN, along with for time series modelling, is not a practical choice. CNN is often used in a hybrid model along with other kinds of the neural network such as  AWNN or RNN, which can improve the performance significantly. A recent trend in using hybrid neural network models becoming very popular in a different application of deep learning, where the model has the attribute from both RNN and CNN is promising in order to achieve a better result. Table \ref{tab:models-for-problems} describes the usage of the different neural network models in the case of solving different categories of problems.

\subparagraph{Classification:} Time series classification task is a complex task for deep learning fraemwork. Time series can be either univariate as described by Eq~\eqref{eq:Xn} or M-Dimentional as described in Eq.~\eqref{eq:md}, where $X_{i}$ itself is a univariate time series. Sometimes time series can be even more complex where $X_{i}$ in Eq.~\eqref{eq:md} can be a multi-variate time series instead of being uni-variate.   

\begin{equation}
\label{eq:md}
X ={X_{1}, X_{1} \dots X_{n}}
\end{equation}

For Time series classification task, time series is described as collection of tuple in \cite{fawaz2019deep} as shown in Eq.\eqref{eq:tuple}. Here time series is a data set, D , consists of a collection of pair $(X_{i}, Y_{i})$, where $X_{i}$ can be either uni-variate or multi-variate time series and $Y_{i}$ is a label. 

\begin{equation}
\label {eq:tuple}
D=\left\{\left(X_{1}, Y_{1}\right),\left(X_{2}, Y_{2}\right), \ldots,\left(X_{N}, Y_{N}\right)\right\}
\end{equation}

Time series classification task becomes more challenging as the number of dimention increases.

\begin{figure}[h]
	\centering\includegraphics[width=\textwidth]{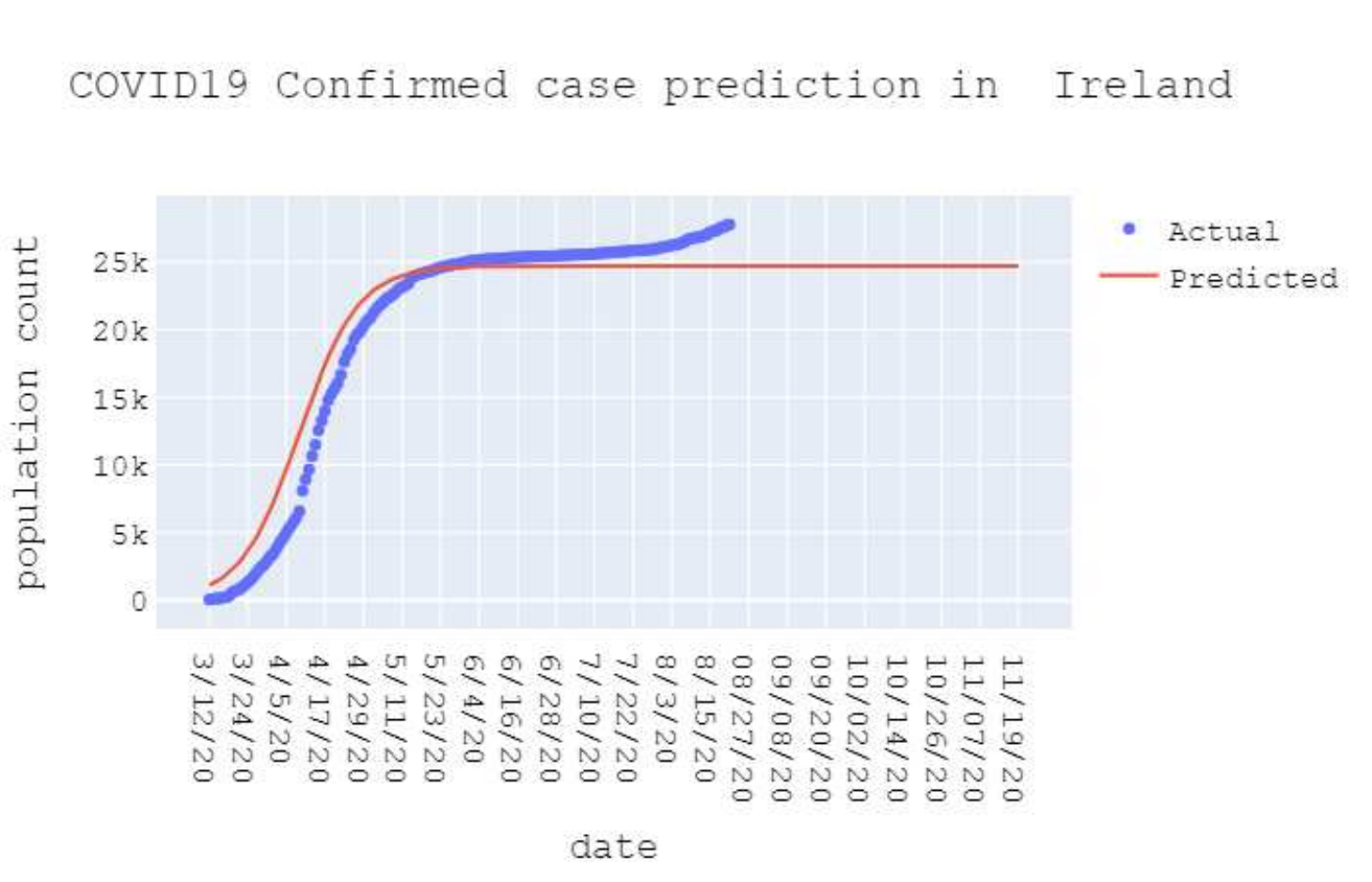}
	\caption{Time series prediction for confirmed covid-19 case in Ireland }\label{fig:prediction} 
\end{figure}

\subparagraph{Prediction}: Prediction tasks uses previous oversavation for any variable such as stock price, rain fall, house price, energy consumption and others; and forecast a future oversavation value for corresponding variable. \cite{lim2020time,doucoure2016time} describes the time series prediction using neural network identifies the relation between previous value and current value of a variable. For example, in Eq.~\eqref{eq:xt}, $f$ is a neura network that can identify the relation between past values of $x$ from time $t-n$ to time $t-1$ with it's current value $x(t)$ at time t. This is an exmaple of simplest time series prediction with one step. Similarly Eq.~\eqref{eq:xt2} shows multi-step ahead prediction, where the next h-th value for variable $x$ is predicted using neural network $f$.

\begin{subequations}

\label{eq:prediction}
\begin{align}
\label{eq:xt}
x(t) = f \{ x(t-1), x(k-2), \ldots, x(t-n-1)\} 
\\
\label{eq:xt2}
x(t+h) = f \{ x(k), x(k-1), x(k-2), \ldots, x(k-n-1)\}
\end{align}
\end{subequations}

In Fig~\ref{fig:prediction} the predicted number of confirmed Covid-19 cases in Ireland are platted.

\subparagraph{Detection:} Detection task is often used for classification and prediction. For example , to undertsnad the difference or anomaly in Actual value and Predicted value shown in Fig\ref{fig:prediction}, time series anomaly detection is essential.

\begin{figure}[h]
	\centering\includegraphics[width=0.9\linewidth]{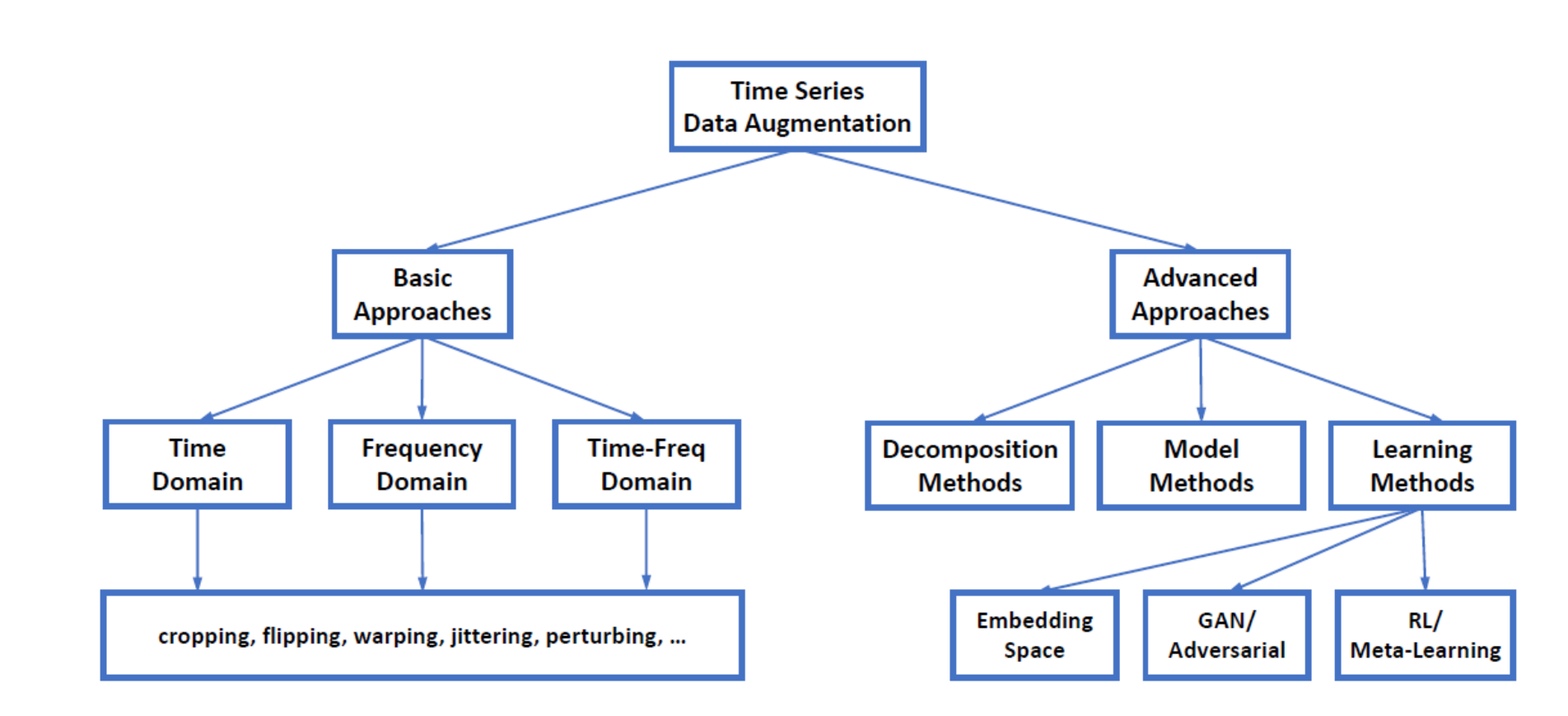}
	\caption{ Taxonomy of time series data augmentation proposed by \cite{wen2020time}.}\label{fig:ts-augmentation} 
\end{figure}

\subparagraph{Augmentation:} A recent survey paper, \cite{wen2020time} describes different techniques for successful time series augmentation used for generating synthetic time series data.  \ref{fig:ts-augmentation} shows the taxonomy of  time series augmentation for deep learning algorithms proposed by \cite{wen2020time}. This review finds the most available time-series augmentation technique. However, due to the complex nature of different fields, the time series data has particular characteristics that need additional treatment. For example, finance multi-variate time data often shows intrinsic probabilistic patterns as a relationship among different variables; health record data shows a direct relationship among different variable at any time point which can be represented as a graph. In this section, these additional time-series augmentation techniques are analysed at length.  

\begin{figure}[h!]
	\label{fig:augmentation}
	\centering
	\begin{subfigure}[b]{0.5\textwidth}
		\centering
		\includegraphics[width=\textwidth]{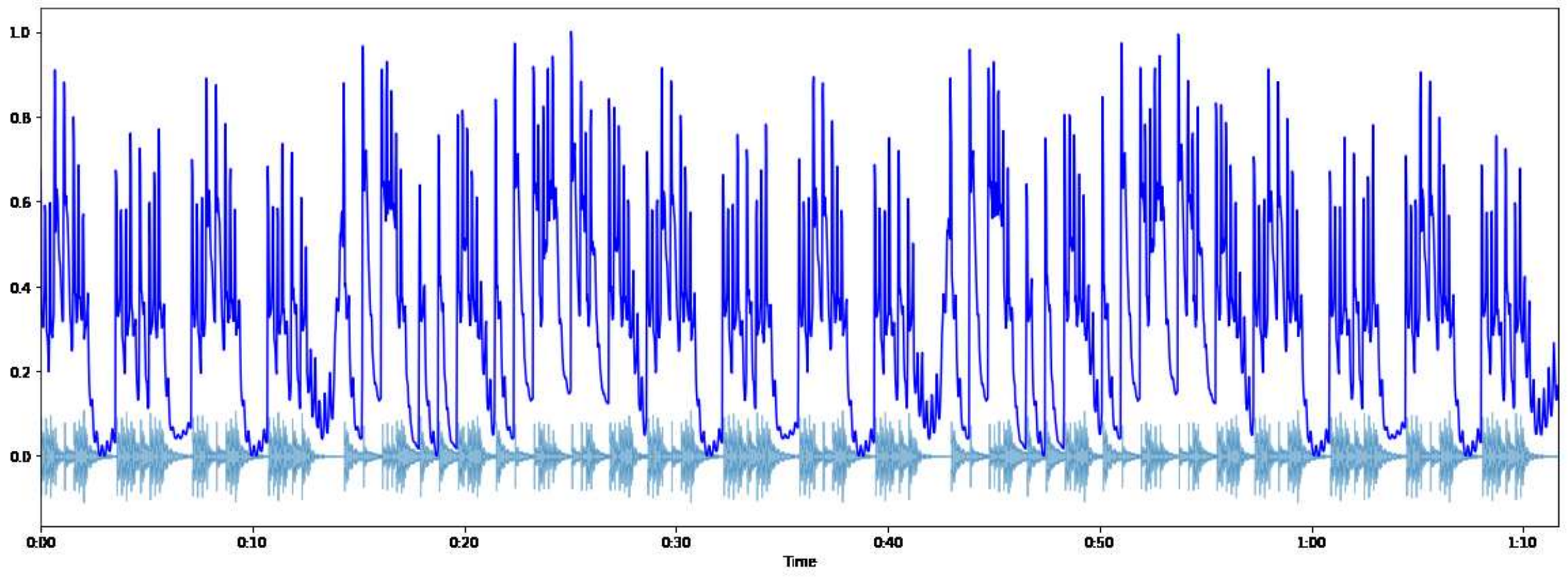}
		\caption{Time Domain}
		\label{fig:type_augmentation_time}
	\end{subfigure}
	\hfill
	\begin{subfigure}[b]{0.5\textwidth}
		\centering
		\includegraphics[width=\textwidth]{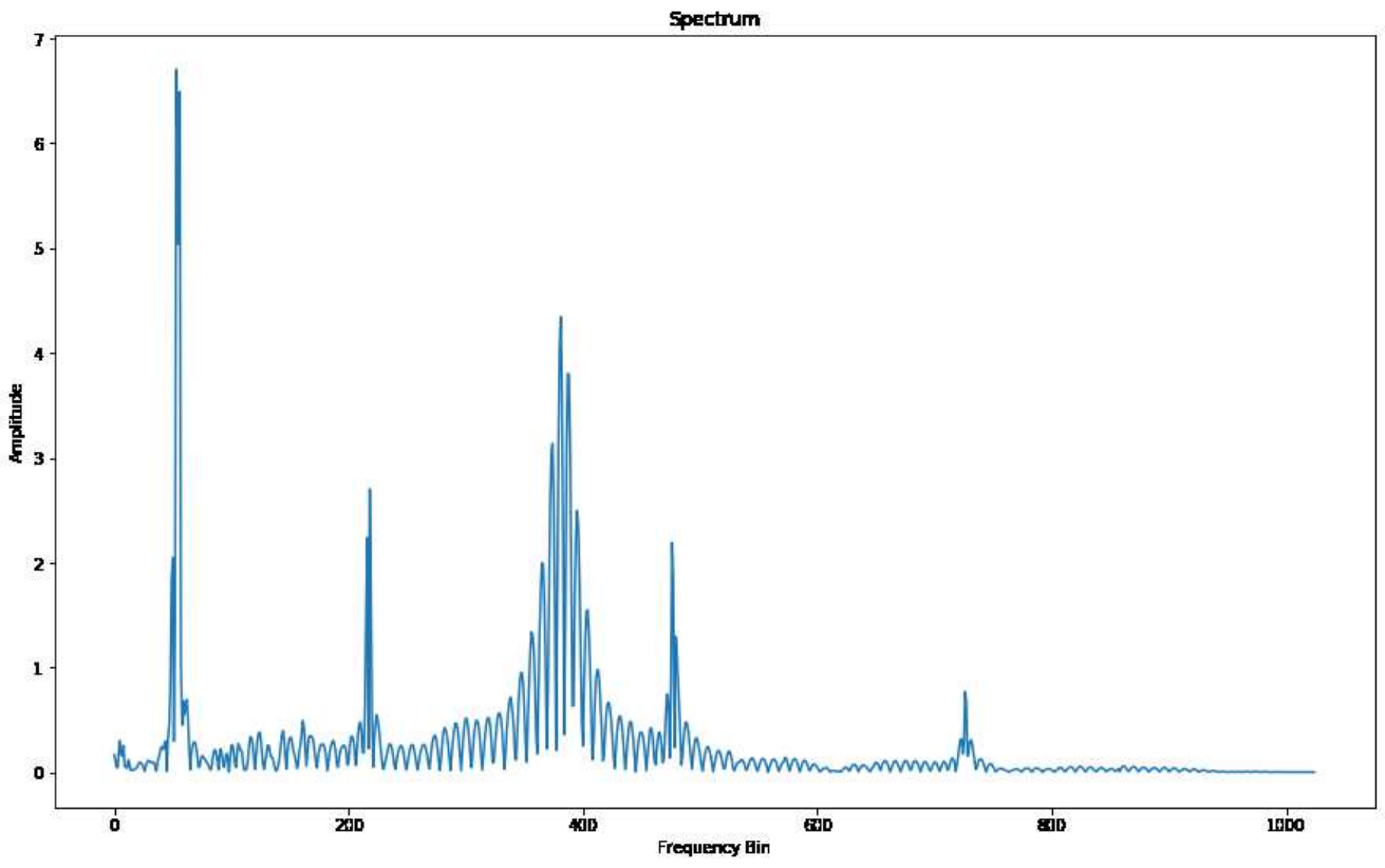}
		\caption{Frequency Domain}
		\label{fig:type_augmentation_frequency}
	\end{subfigure}
	\hfill
	\begin{subfigure}[b]{0.45\textwidth}
		\centering
		\includegraphics[width=\textwidth]{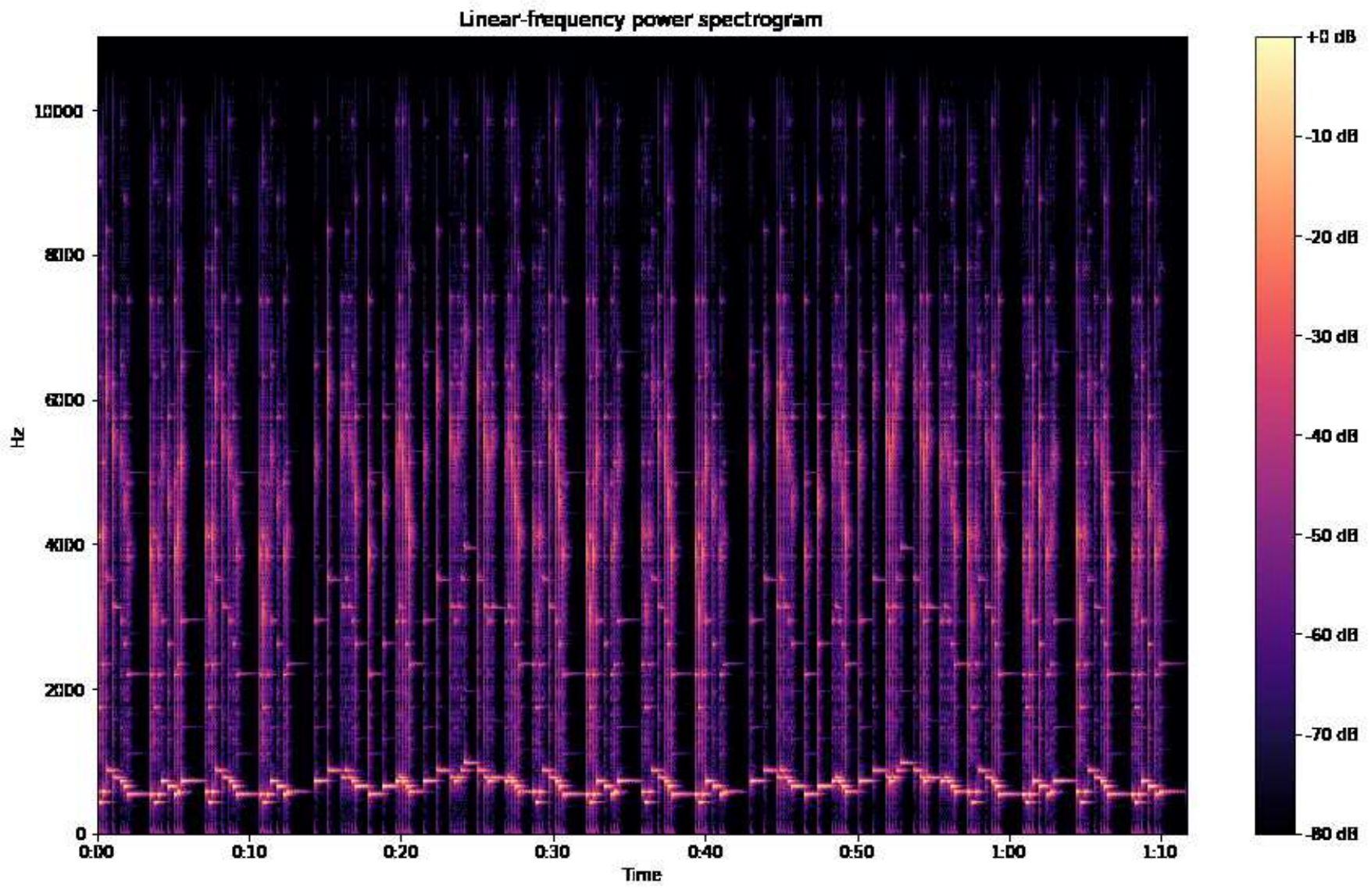}
		\caption{Time-Frequency Domain}
		\label{fig:type_augmentation_time-frequency}
	\end{subfigure}
	\caption{Different types of augmentation for time series}	
\end{figure}

Fig.~\ref{fig:type_augmentation_time}, \ref{fig:type_augmentation_frequency} and \ref{fig:type_augmentation_time-frequency} show different kind of augmentation for a continous audio file \textit{Pokemon - pokecentre theme.wav} from the Pokemon-midi \cite{corynguyen19} dataset.

Besides, these time series data augmentation techniques, there are some other advanced augmentations; such as (i) Graph Augmentation  (ii) Embedding Entropy with Ordinal patterns to describe the intrinsic patterns; are increasingly getting attention among researchers.

\begin{table}[h]
\begin{center}
\caption{DIFFERENT NEURAL NETWORK MODEL FOR DIFFERENT TIME SERIES PROBLEM} \label{tab:models-for-problems}
\begin{tabular}{|p{.6in}|p{0.7in}|p{0.9in}|p{0.7in}|p{0.7in}|p{1.0in}|} \hline 
Deep Model & Modelling /Augmentation & Classification & Prediction & Detection   \\ \hline 
\textbf{ANN } & \cite{doucoure2016time, liu2019time} & & \cite{doucoure2016time,ali2012neural,zhang2019applying} &  \\ \hline 
\textbf{CNN} & & \cite{fawaz2019deep} & &   \\ \hline 
\textbf{RNN} & & & \cite{deo2016identification,bianchi2017overview} & \\ \hline 
\textbf{LSTM} & \cite{neil2016phased, sutskever2009recurrent, hochreiter1997long,timelstm}  &  & \cite{bianchi2017overview} &   \\ \hline 
\textbf{GRU} & \cite{grud} &  & \cite{ref13, bianchi2017overview} &  \\ \hline 
\textbf{DNN} & \cite{hinton2012deep}  &  &  &  \\ \hline 
\textbf{Hybrid} & \cite{chen2018neural,raissi2017physics,yildiz2019ode} & & \cite{wang2012hybrid, yang2017continuous, koller2016deep} & \cite{canizo2019multi} \\ \hline 
\textbf{AWNN} & \cite{cohen2010statistical} & \cite{bhardwaj2020comparative}  &  & \cite{ahn2020efficient} \\ \hline 
\textbf{Graph} &\cite{structuralrnn,fcnrlstm} &&& \\ \hline
\end{tabular}
\end{center}
\end{table}

\subsection{Different dataset}
\label{sec:dataset}
This section discuss some well-known datasets used for different research. Different dataset focuses on different characteristics of continuous-time data as discussed in \ref{sec:domain}. Table \ref{tab:dataset} shows the feature for some well known datasets.

\begin{table}[htb]
\centering
\caption{Well-known datasets for continuous-time data research }
\label{tab:dataset}
\resizebox{\textwidth}{!}{%
\begin{tabular}{|l|l|l|l|l|} \hline
\textbf{Dataset} & \textbf{Data type} & \begin{tabular}[c]{@{}l@{}}\textbf{Application}\\ \textbf{Field}\end{tabular} & \textbf{Nature} & \textbf{Source} \\ \hline

MIMIC-III& \begin{tabular}[c]{@{}l@{}}Electronic Health Record \\ (EHR)\end{tabular} & Health Care  & \begin{tabular}[c]{@{}l@{}}Missing pattern,\\ irregular sampling rate \end{tabular}&  \\ \hline

\begin{tabular}[c]{@{}l@{}}Global Man-made Impervious Surface (GMIS)\\ DMSP/OLS night-time stable light (NSL)\\ Global Population Density Grid \\Time Series Estimates\\ British  Petroleum (BP) Statistical \\ Review of World Energy \end{tabular} &  & Energy &  & NASA(SEDAC) \\ \hline

BreizhCrops & satellite image time series  & Agriculture &\begin{tabular}[c]{@{}l@{}} Supervised classification \\of field crop \end{tabular}& \\ \hline 

Multivariate Time Series Search& sensor generated data for media and event& Aerospace  & Multi-variate & NASA(SEDAC)  \\ \hline
MNIST \cite{lecun-mnisthandwrittendigit-2010}& image & image processing, object detection and classification & complex series & YANN \\ \hline

\end{tabular}}
\end{table}

Different libraries and packages are developed to generate time targeting different algorithms for time series problems. \cite{faouzi2020pyts} is a python package which provides the implementation along with example dataset for different time series classification algorithms such as  Dynamic Time Warping, Shapelet Transform, Markov Transition Field, Time Series Forest and others. 

For healthcare, the challenges of availability of dynamic dataset is sever than other research fields. Due to data privacy issue, a limited number of the dataset is publicly available, which focuses on a limited number of aspects of health care. Most researches in health care with continuous-time data are biased towards these same EHR datasets, which has a significant impact on overall research for health care. There are some efforts \cite{dash2019synthetic} to generate synthetic health data can simulate the temporal relations of multiple variables in a patient’s health record.

Similar to health care, some other research area also suffers from the absence of enough labelled data. These fields generate their own data. Using automated tools to generate a synthetic dataset for unexplored input space in case of various time series problems, including time series classification, forecasting and anomaly detection is very popular. A successful tool requires the following feature 

\begin{enumerate}
\item can generate very long time series with multiple variable 
\item can generate high-quality continuous-time data
\item uses efficient data augmentation technique.
\end{enumerate}

Table \ref{tab:dataset2} shows some well-known dataset used in different problem-solving in the problem domain of continuous time series. 

\begin{table}[htb]
\centering
\caption{Some well-known dataset for time series problems}
\label{tab:dataset2}
\resizebox{\textwidth}{!}{%
\begin{tabular}{|l|l|l|l|} \hline
\textbf{Classification}  & \textbf{Modelling} & \textbf{Prediction} &\textbf{ Regression }\\
\toprule

\begin{tabular}[c]{@{}l@{}}Human Activity \\ Recognition\cite{anguita2013public}\end{tabular} & \begin{tabular}[c]{@{}l@{}}SPSS with USDA feed\\ grains \cite{spss}\end{tabular} & \begin{tabular}[c]{@{}l@{}}Daily Demand\\  Forecasting \cite{ferreira2016study}\end{tabular} & Air Quality \cite{vito2008onfield}\\ \hline

\begin{tabular}[c]{@{}l@{}}Gesture Phase\\ Segmentation \cite{gesture}\end{tabular} & EEG \cite{bay2000kibler} & \begin{tabular}[c]{@{}l@{}}Google Web Traffic \\ Time Series Forecasting\end{tabular} &  \\ \hline

MINST \cite{lecun-mnisthandwrittendigit-2010} & \begin{tabular}[c]{@{}l@{}}Activities modelling as \\ Fuzzy Temporal Multivariate\\ model\end{tabular} & Population Time Series &  \\ \hline

CIFAR10 & \begin{tabular}[c]{@{}l@{}}Free Music \\ Archive (FMA)\end{tabular} & \begin{tabular}[c]{@{}l@{}}Climate Change \\ Earth Surface Temperature\end{tabular} &  \\ \hline

\begin{tabular}[c]{@{}l@{}}PhysioNet \\ Challenges dataset\cite{johnson2019mimiccxr} \end{tabular} &  & \begin{tabular}[c]{@{}l@{}}Hourly Energy \\  Consumption\end{tabular} &  \\ \hline

MMIC-III \cite{johnson2019mimiccxr}&  & Financial Distress Prediction &  \\ \hline

Time Series \cite{UCRArchive2018} &&& \\ 

\bottomrule
\end{tabular}%
}
\end{table}

\section{ Different Applications of Continuous Time Series}
\label{sec:application}
There are several applications of sequential data modelling, for example, speech recognition, bioinformatics and human activity recognition. In speech recognition, the input is a continuous or discrete audio clip $P$, which needs to be mapped to corresponding text transcript $Q$. In this example, the input and the output are both sequential data containing temporal information. Input, $P$, is an audio clip and so that plays out over time and output, $Q$, is a sequence of words. Therefore, deep learning models are suitable for sequential data such as recurrent neural networks and its other variations, have become promising for speech recognition—another example of the continuous-time series modelling in music generation. In the case of music generation, only the output $Q$ is a sequence of music notes, where the input can be an empty set, or a single integer, just representing the genre of music. The input can also be a set of the first few notes. The output $Q$ is a sequence of data. This sequence also has implicit temporal information. A third example is sentiment classification, where the input $P$ is a sequence of information and output is a  discrete set of values.
Even in bio-informatics, sequential data modelling is extensively important for the different fundamental research area, such as DNA sequence analysis. In DNA analysis, the input $P$ is a sequence of the alphabetic letter, and the corresponding output is a corresponding label ($Q$) such as protein for each part of the input sequence. Here input, $P$ is a sequence of data and output, $Q$ is a set of labels.
Human activity recognition or video activity recognition is also related research fields where sequential data model like time series can be beneficial where the input $P$ is a sequence of video frames, and the output is corresponding activity.
Future trend prediction in financial data or event prediction from a continuous sequence of sensory information is some recent area where sequence modelling can bring a revolution in accuracy as well as efficiency.
These are just a small subset of sequence modelling problem domains. There are a whole lot of different types of sequence modelling problems. These kinds of problems have some common structure and challenges. Following are some features from different sequence modelling problem 
\begin{itemize}
	\item Both input ($P$) and output ($Q$) can be sequenced
	\item Only one of the Input ($P$) and Output ($Q$) can be sequenced, either input or output
	\item 	$P$ and $Q$ can have different length
	\item 	$P$ and $Q$ can have the same length
	\item 	$P$ and $Q$ can be either continuous or discrete
\end{itemize}

Time sequences can be different in length, characteristics, nature and behaviour. On top of that, the corresponding output can also vary in types. Therefore, the deep learning model needs to be coherent, robust, dynamic and efficient to achieve a better result.

Different applications and problem domains are prone to different challenges of the continuous time series. For example, in \cite{neil2016phased} the main challenges  of event prediction using sensor data have been described are as follows 

\begin{table}[h]
	\begin{center}
		\caption{DIFFERENT APPLICATION OF DIFFERENT TIME SERIES PROBLEM}\label{tab:time-series-problems}
		\begin{tabular}{|p{1.5in}|p{1.0in}|p{1.2in}|p{0.8in}|p{0.8in}|} \hline 
			\textbf{Application}            &  \textbf{Modelling / Augmentation} &  \textbf{Classification} &  \textbf{Prediction}&  \textbf{Detection} \\  \hline
			Time series behaviour & \cite{jimenez2020feature, singh2019multi} & \cite{le2016data,ahn2020efficient, fawaz2019deep, bahadori2019temporal,wang2016effective} & \cite{bhardwaj2020comparative,kuremoto2019training,jiang2016time}  & \cite{canizo2019multi, chavez2019detecting}\\ \hline

			\begin{tabular}[c]{@{}l@{}} Climate \end{tabular}  &  \cite{timelstm, skiprnn, ref14}&   &   &   \\ \hline
			
			\begin{tabular}[c]{@{}l@{}}Financial trend\\ forecasting \end{tabular} &  \cite{ehr, structuralrnn,henry2019permutation} &   & \cite{mohapatra2019financial,wang2016financial, ferreira2016study}  &   \\ \hline

			Event prediction  & \cite{neil2016phased} &  & \cite{connor1994recurrent, zhang2017deep}  &  \\ \hline
			
			Human Activity Tracking & \cite{humantracking,timelstm}   &   &   & \cite{gesture}   \\ \hline
			
			Image Processing &  & \cite{brunel2019cnn} & & \\ \hline
			
			Helath Data & \cite{dash2019synthetic} & \cite{michielli2019cascaded,chambon2018deep} & \cite{zhang2019predicting,ballinger2018deepheart,kissas2020machine, jimenez2020feature} & \cite{sosa2017real} \\ \hline
			
			Speech recognition  & \cite{hinton2012deep, mandarin, 10.5555/3137257}  &   &   &  \\ \hline

			Character detection   & \cite{mandarin} & \cite{yang2017continuous,koller2016deep} &  & \cite{10.5555/3137257} \\ \hline
			\begin{tabular}[c]{@{}l@{}}Human activity\\ recognition \cite{humantracking}\end{tabular}   &   &   &   &   \\ \hline
			
			Signal processing  &  \cite{liu2019neural, yao2019stfnets, zhou2018recover, yao2017deepsense} &   & \cite{ruta2007neural, livieris2020cnn}  &   \\ \hline
			
			Climate  &  \cite{nourani2019rainfall,liu2019time} &   & \cite{khaledian2020simulating,ayazpour2018combined, wu2019hybrid, byakatonda2019prediction}  &  \cite{vito2008onfield}  \\ \hline
			
			Robotics & & & & \cite{chen2019intelligent} \\ \hline 
		\end{tabular}
	\end{center}
\end{table}


\section{Different challenges in modelling time sequence}
\label{sec:challenges} 
Continuous-time dataset possesses several unique behaviours, which makes modelling data as well as learning the hidden dynamics or pattern extensively challenging. This section elaborates different challenges of continuous-time data processing using neural networks, and highlight the recent trend for solving those challenges.

\subsection{Modelling hidden dynamics of dynamic temporal system}

Modelling continuous time series is the fundamental task for each of the continuous-time series problem categories mentioned in section \ref{sec:domain}. Different deep learning models demonstrate a wide range of techniques to overcome challenges and design seamless dynamical system with continuous-time data. Among different deep learning models, ANN, RNN, CNN, DNN, DBN and some other hybrid neural network models exhibit higher precision of accuracy. However,  their performance varies based on the system architecture and characteristics of underline data. No single model is suitable for all different continuous-time dynamics system. For example, for a continuous-time dynamical system with regular as well as irregular data sampling rat, LSTM, and different variations of LSTM show promising results than other neural networks \cite{hochreiter1997long, neil2016phased}. LSTM can capture the long-term dependency in continuous-time series, which increases the success rate, efficiency of the model and precision of the accuracy. Still, LSTM has some weakness as well. One of the limitations of LSTM based neural network models is that it cannot explicitly model the pattern in the frequency distribution, which is a critical component in terms of solving time series prediction problem. To solve this limitation of LSTM, a recent novel work \cite{hu2017state} which decomposes the memory states of an input sequence into a set of frequency set. \cite{hu2017state} uses the time sequence of length of T represented by Eq.~\eqref{eq:Xn} in section \ref{sec:characteristics} as the input, where each observation ($x_{t}$) belongs to an N-dimensional space $\mathbb{R}^N$, i.e., $x_{t} \in \mathbb{R}^N$. Similar to LSTM, this model uses a sequence of the memory cell, with length same as the time duration (T), to model the dynamics of the input sequence, but each of the memory cell states is decomposed into a set of frequency components as shown in Eq.~\eqref{eq:5.1}. Here, F represents a state- frequency decomposition of the input sequence across different states and frequencies. 

\begin{equation}
\label{eq:5.1}
F=\ w_1,w_2,\ldots, w_K\ 
\end{equation}

To update the cell state, this model uses a State Frequency Memory (SFM) matrix, $S_{t} \in \mathbb{C}^{D \times K}$, where D is the number of dimensions and K is the number of frequency states as shown in Eq.~\eqref{eq:5.2}.  Forget gate ($f_{t}$) and input gate ($g_{t}$) control the current as well as previous memory state and frequency to update the SFM at time t using a modulation ($i_{t}$) of the current input. Eq.~\eqref{eq:5.8}  shows that ($i_{t}$) combines the current input and the output vector from previous state $ z_{t} $.  SFM combines memories from the previous cell to the current input similar to LSTM. 

\begin{equation} \label{eq:5.2}
\mathbf{S}_{t}=\mathbf{f}_{t} \circ \mathbf{S}_{t-1}+\left(\mathbf{g}_{t} \circ \mathbf{i}_{t}\right) \left[ \begin{array}{c}{e^{j \omega_{1} t}} \\ {\cdots} \\ {e^{j \omega_{K} t}}\end{array}\right]^{T} \in \mathbb{C}^{D \times K}
\end{equation}

\begin{equation}\label{eq:5.8}
\mathbf{i}_{t}=\tanh \left(\mathbf{W}_{i} \mathbf{z}_{t-1}+\mathbf{V}_{i} \mathbf{x}_{t}+\mathbf{b}_{i}\right) \in \mathbb{R}^{D}
\end{equation}

This work proposed an improved forget gate which is a combination of state forget gate ($\mathrm{f}_{t}^{\mathrm{ste}}$), described in Eq~\eqref{eq:5.3}, and frequency forget gate($\mathrm{f}_{t}^{\mathrm{fre}}$),described in Eq~\eqref{eq:5.4}. Therefore the forget gate as shown in Eq.~\eqref{eq:5.5}) is decomposed over memory states and frequency of the states and control the input. 

\begin{equation}
\label{eq:5.3}
\mathrm{f}_{t}^{\mathrm{ste}}=\sigma\left(\mathbf{W}^{\mathrm{ste}} \mathbf{z}_{t-1}+\mathbf{V}^{\mathrm{ste}} \mathbf{x}_{t}+\mathbf{b}^{\mathrm{ste}}\right) \in \mathbb{R}^{D}
\end{equation}

\begin{equation}\label{eq:5.4}
\mathrm{f}_{t}^{\mathrm{fre}}=\sigma\left(\mathbf{W}^{\mathrm{fre}} \mathbf{z}_{t-1}+\mathbf{V}^{\mathrm{fre}} \mathbf{x}_{t}+\mathbf{b}^{\mathrm{fre}}\right) \in \mathbb{R}^{K}
\end{equation}

\begin{equation}\label{eq:5.5}
\mathbf{f}_{t}=\mathbf{f}_{t}^{\mathrm{ste}} \otimes \mathbf{f}_{t}^{\mathrm{fre}}=\mathbf{f}_{t}^{\mathrm{ste}} \cdot \mathbf{f}_{t}^{\mathrm{fre}^{\prime}} \in \mathbb{R}^{D \times K}
\end{equation}

Here, W and V  in Eq~\eqref{eq:5.3} and \eqref{eq:5.4} are the weight vectors and $z_{t}$ is an output vector computed from the amplitude ($ \mathbf{A}_{\mathbf{z}}^{k} $) of SFM at time t and the output of the output gate for each frequency component (k), of the previous cell as shown in Eq.~(\eqref{eq:5.6}). The multi-frequency aggregated output($\mathrm{z}_{t}$) as shown in Eq.~(\eqref{eq:5.7}) controls the input for forget and input gate. 

\begin{equation}\label{eq:5.6}
\mathbf{z}_{t}^{k}=\mathbf{o}_{t}^{k} \circ f_{o}\left(\mathbf{W}_{\mathbf{z}}^{k} \mathbf{A}_{\mathbf{z}}^{k}+\mathbf{b}_{\mathbf{z}}^{k}\right), \text { for } k=1, \cdots, K
\end{equation}

\begin{equation}\label{eq:5.7}
\mathrm{z}_{t}=\sum_{k=1}^{K} \mathrm{z}_{t}^{k}=\sum_{k=1}^{K} \mathrm{o}_{t}^{k} \circ f_{o}\left(\mathbf{W}_{\mathrm{z}}^{k} \mathrm{A}_{\mathrm{z}}^{k}+\mathrm{b}_{\mathrm{z}}^{k}\right) \in \mathbb{R}^{M}
\end{equation}
Finally, the output of the output gate as shown in Eq.~(\ref{eq:5.9}) uses a weight matrix  $ \mathbf{U}_{\mathbf{o}}^{k} $

\begin{equation}\label{eq:5.9}
\mathbf{o}_{t}^{k}=\sigma\left(\mathbf{U}_{\mathbf{o}}^{k} \mathbf{A}_{t}^{k}+\mathbf{W}_{\mathbf{o}}^{k} \mathbf{z}_{t-1}^{k}+\mathbf{V}_{\mathbf{o}}^{k} \mathbf{x}_{t}^{k}+\mathbf{b}_{\mathbf{o}}^{k}\right) \in \mathbb{R}^{M}
\end{equation}

The frequency state helps to learn the dependencies of both low and high-frequency patterns. Learning frequency pattern ultimately improves the performance in the case of time series prediction problem. Time-frequency analysis is most famous for noisy time series prediction problems. This is a common feature for multivariate time series data, e.g. weather data, financial time series, pattern recognition from continuous-time data and others.   

Table \ref{tab:timefrequency} shows several works, where time-frequency analysis is used for time series prediction 
The most popular neural network for time-frequency analysis is based on ANN, LSTM, FFT and back-propagation neural network (BPNN). A few one-dimensional CNN based works also exhibit better performance in case of lack of sufficient training data  
 
\begin{table}[htp]
\centering
\caption{Application of  Time–frequency analysis}
\label{tab:timefrequency}
\begin{tabular}{|p{2.0in}|p{1.5in}|p{1.3in}|} \hline 
\textbf{Application} & \textbf{Base Neural network} & \textbf{Proposed work}\\ \hline 

Rain-fall prediction & ANN & \cite{byakatonda2019prediction} \\ \hline
Multi-variate time series forecasting(e.g. wind, financial, energy consumption) & LSTM &  \cite{shih2019temporal} \\ \hline
Sea surface temperature forecasting & BPNN & \cite{wu2019hybrid} \\ \hline
Intelligent Fault diagnosis & CNN & \cite{chen2019intelligent} \\ \hline
Sleep stage classification & Cascade LSTM  &  \cite{michielli2019cascaded} \\ \hline
Signal processing & Short-Time Fourier (STFNet)&  \cite{yao2019stfnets} \\ \hline
\end{tabular}
\end{table}

\subsubsection{ Irregular data sampling rate and fixed time step}
Existing deep learning models for solving continuous-time data usually use fixed data sampling rate, but in real-time time rate is mostly irregular. For example, event stream data in the automobile industry, patient record in healthcare, weather data in climate applications, all these real-time data demonstrate irregular sampling rate. So far, RNN based model pioneer among all state-of-art neural network models for modelling irregularly sampled data by considering the continuous-time data as a sequence of discrete fixed-step data which often suffer from loss inaccuracy. 

\begin{table}[htp]
\centering
\caption{Potential deep learning-based solutions for irregular sampling rate}
\label{tab:irregularsamplingrate}

\begin{tabular}{|p{1.5in}|p{2.9in}|} \hline 
\textbf{Proposed works} & \textbf{Neural Network Model} \\ \hline 
 \cite{grud} & GRU-D   \\ \hline 
 \cite{neil2016phased} & Phased LSTM   \\ \hline 
 \cite{chang2017dilated} & Dilated RNN  \\ \hline 
 \cite{timelstm} & Time- LSTM   \\ \hline 
 \cite{skiprnn} & SKIP -RNN   \\ \hline
 \cite{rubanova2019latent} & Latent-ODE \\ \hline
 \cite{bahadori2019temporal}& Temporal clustering \\ \hline
 \cite{singh2019multi} & Multi-resolution Flexible Irregular Time series Network (Multi-FIT) \\ \hline

 \cite{kidger2020neural}& Neural Differential equation \\ \hline
\cite{2020arXiv200509807H} &  RNN based Neural Differential equation   \\ \hline
\end{tabular}
\end{table}

The variable data sampling rate is another significant challenge for continuous-time data.  Most of the time, in a continuous-time series, the state value does not update at every time step. Therefore, the value of the selected parameter cannot be found at every state of the time sequence, which may lead to an inaccurate result. Therefore, solving irregular data sampling is very important. More or less every continuous-time series suffers from the problem of having an irregular sampling rate, where the interval between consecutive data collection point can vary in length. An excellent example of such a scenario in the healthcare industry is a patient health record. The interval of two consecutive hospital visits for a patient can be a few days or even a few years.
Similarly, the voice command for Alexa can be sampled within a few minutes or even days. An efficient deep learning model needs to understand the irregularity in the data sampling rate. Temporal clustering invariance\cite{bahadori2019temporal} is a unique technique to solve irregular healthcare data by grouping regularly spaced time-stamped data points together and then cluster them, yielding irregularly-passed time-stamps. Another recent neural network called Multi-resolution Flexible Irregular Time series Network (Multi-FIT) \cite{singh2019multi} fights against irregularly-paced observation of a multivariate time series. Instead of using data imputation technique to replace missing observation used by \cite{grud}, \cite{singh2019multi} uses a FIT network to create informative representation at each time step using last observed value or time interval since the last observed time stamp and overall mean for the series. Some recent works, as shown in Table [\ref{tab:irregularsamplingrate}], mainly focus on sampling irregular data sampling.  

\subsubsection{Informative missingness}
 One of the serious outcomes of the irregular sampling rate is informative missingness \cite{grud}. Informative missingness is a major challenge in terms of processing time series. Problems in the prediction category, suffer missingness challenge the most. Missing values in the time series and their missing patterns are often correlated. Understanding this correlation can lead to better prediction result. Although due to some useful attributes in the design, RNN is well equipped to capture long-term temporal dependencies and variable-length observations.  There are some relevant models based on RNN, as shown in table \ref{tab:missing}. Among these works, \cite{grud} impressively improve RNN structures to incorporate the patterns of missingness for time series classification problems. Besides, the sparse and asynchronous nature of the data sampling rate causes missing observation. For each missing observation, the modelling of time series gets interrupted and can never recover again. The primary way to battle this irregular data sampling rate is to impute the missing data to provide value for missing observations similar as proposed in GRU-D\cite{grud} model. Another efficient mechanism, usually used by RNN based models, is to let the model know when there is data available and take action accordingly, as demonstrated in Phased-LSTM \cite{neil2016phased} model.  Over the last decades, several research works motivate to fix the informative missingness due to missing observation in the time series. Here are some conventional approaches to solve this problem as follows: 

\begin{enumerate}
\item  To ignore the missing data point and to perform the analysis only on the observed data. The limitation of this approach is that if the missing rate is high, and the sampling rate is too few, the performance decreases significantly. 

\item  To substitute in the missing values using data imputation. Data imputation mechanism \cite{kreindler2006effects, bai2018convolutional, grud} are widely popular for solving missing data. However, data imputation does not always capture variable correlations, complex pattern and other important attributes. GRU-D \cite{grud} is based on the idea that. This model combines two different representation of missing patterns, such as masking and time interval in the deep learning architecture to detect the long-term pathological time dependencies in the time series and also uses the missing pattern to achieve better prediction result.  
\item To Leverage ordinary as well as neural network based on partial differential equation helps to learn the change in data over time and use the instant derivative of the state of any dynamical system over time to replace the missing observation state value. 
\end{enumerate}

\begin{table}[htp]
\centering
\caption{ Different NN models as solution for informative missingness}
\label{tab:missing}
\begin{tabular}{|p{2.0in}|p{2.9in}|} \hline 
\textbf{Proposed work} & \textbf{Underline Neural network model} \\ \hline 
 \cite{grud} & GRU-D \\ \hline 
\cite{ehr} & Phased -LSTM- D \\ \hline 
\cite{neil2016phased} & Phased LSTM   \\ \hline 
 \cite{sun2019neupde} & partial Differential equation based  \\ \hline
 \cite{2020arXiv200510693H} &  GRU-D based Neural Differential equation   \\ \hline
\end{tabular}
\end{table}

Recovering Missing data from the asynchronous at a heterogeneous sampling rate is very challenging work but essential or the efficiency of the model. The missing entries in a continuous-time series can result in large and random distribution, the phased-LSTM \cite{neil2016phased} has introduced the time gate to overcome complexity with the asynchronously sampled data, and it can sample data at any continuous-time within its open period. Several recent works emphasise on hybrid LSTM to recover missing data. For exaple,  J. Zhou et al. \cite{zhou2018recover} proposed a neural network consists of the standard LSTM \cite{hochreiter1997long} for regularly sampled data and the Phased-LSTM \cite{neil2016phased} for irregularly sampled data as shown in Eq.~\eqref{eq:lstms}.  The benefit of the model learning is that it utilises not only the information from collected observations but also previous input's missing values and thereby, deal with the data sparsity caused by missing data.

\begin{equation}
\label{eq:lstms}
\begin{aligned}
h_{s}^{f} &=L S T M^{f}\left(h_{s-1}, x_{s}\right) \\
h_{u}^{b} &=L S T M^{b}\left(h_{u+1}, x_{u}\right)
\end{aligned}\end{equation}

The most commonly used mechanism in order to overcome the informative missingness is to use interpolation and data imputation.  Here is a list of widely used imputation mechanism
\begin{itemize}
	\item PCA (Principle component analysis)
	\item	MICE (Multiple Imputation by chained equation)
	\item	MF (Matrix Factorization)
	\item	Miss Forest
	
\end{itemize}

One well-known technique is to sample the data only when available. Several neural networks \cite{neil2016phased, timelstm, skiprnn} are designed with this principle. A discrete-time series can be defined as Eq.~\eqref{eq:Xn}, which is a sequence of observed data points at consecutive time steps. For the time series X, $x_i$ is a set of observations at time step $t_i$. Usually, the time sequence is long, and it needs sampling in order to optimise the observation data set at different time steps. Some step can have data while others may have no data for the underlined subject of observation. Therefore, neural networks require learning a very long time series where data is collected at an arbitrary sampling rate. Missing data pattern in time series impacts the final outcome significantly. As a result, it is mandatory for neural network models to be designed to overcome the informative missingness. 

\subsubsection{ Temporal and Spatial Coherence}
 Real-world continuous-time data dynamically evolve continuously with noise as well as statistical properties, both temporal and spatial coherence influence the efficiency of the neural network used for time series modelling. Temporal coherence refers to the correlation between $z(t_{1})$  and $z(t_{2})$, $z$ is a vector representing any dynamical system, and $z(t_{1})$  and $z(t_{2})$ are the state value of $z$ in two different data points $t_{1}$  and $t_{2}$. Different real-world applications \cite{sun2012seasonality} are tightly dependent on temporal coherence. The recent trend of data-driven neural network \cite{2020arXiv200509807H,holl2020learning} supports the importance of temporal coherence in most time series problems.
For some domain-specific continuous-time data, such as prediction of wind, energy consumption, climate phenomenon \cite{clemson2016reconstructing, sheppard2016changes}, exhibits spatial coherent pattern in corresponding multivariate time series. Wavelets analysis \cite{cazelles2014wavelet, ng2012geophysical, cohen2010statistical} is one of the most popular mechanisms to learn transient coherent patterns in time series.  Dynamic spatial correlation for multivariate time series \cite{chavez2019detecting} combines wavelet analysis along with non-stationary Multifractal surrogate-data generation algorithm to detect short-term spatial coherence in multivariate time series. The surrogated data are generated by the stochastic process, with the amplitude and time-frequency distributions of original data being preserved. An LSTM based neural network \cite{fcnrlstm} demonstrate the usage of spatial coherence in neural network modelling for time series with Spatial Coherence.

\subsubsection{ High Dimensionality}
Real-world time series contains noise components which may add additional complexity to the modelling and processing of time series \cite{ruta2007neural, chavez2019detecting}. Besides, multivariate time-series data samples exhibit high dimensionality. For time-series data processing, it is essential to optimise the noise as well as Dimension. There are several mechanisms to overcome high dimensionality problem for multivariate time series, such as restricted Boltzmann Machine \cite{sutskever2009recurrent}, feature extraction, wavelet analysis, filtering feature and others. One of the solutions is to use  Wavelet analysis which removes some portion of noise and reduces the dimensionality. Different kind of optimisation functions can be used to optimise the weight matrix. Peng et al. \cite{jiang2016time} have transformed the weight matrix optimisation problem into a Lagrangian dual problem to overcome the high dimensionality. Another popular solution is feature extraction. In feature extraction, neural network models only focus on a set of features from the long time series for computation. However, this can impose a negative effect on the time series process as an essential or relevant feature can be excluded, and the result can be misleading. On the other hand, if the number of selected feature is large, the associated computation, time, memory and energy would be extensive as well, and that would influence the performance negatively.

\subsubsection{ Over-fitting}

For any deep learning model, it is crucial to provide a mechanism for avoiding over-fitting. In the case of modelling a continuous-time data, it is mandatory and challenging. Some common approach for avoiding over-fitting are listed as follows

\begin{enumerate}
\item  Bach normalization \cite{srivastava2014dropout}

\item  Dropout \cite{hinton2012deep}

\item  Optimizing weight matrix \cite{jiang2016time}

\item  Using un-tuned weight matrix

\item  Reducing the Dimension of the representation of the input vector by shifting a row or column 

\item  Using Gaussian or similar process for feature extraction

\item  Using comparatively less memory size so the model cannot memorise the input sequence

\item  ANN-based model uses early stopping procedure to avoid the over-fitting problem. In the case of early stopping a predetermined number of step controls an automated stopping procedure. 

\item  Using a broad set of training and test datasets can reduce the chances of over-fitting.

\item Regularisation is a well-known procedure for overcoming over-fitting. Sampling noise is a very excellent solution for Regularisation. 

\item  Fine-tuning of the neural network model prone to stop early, thereby, reduce over-fitting.

\item  A unique method generative pre-training \cite{hinton2006reducing}, where representation mapping of input data is done using feature detectors before the actual training.  In this process, multiple layers of feature detectors are for a discriminative fine-tuning phase along with adjusting the weight found in the pre-training phase. This process significantly reduces the chances of over-fitting. 
\end{enumerate}

\subsubsection{ Length of Sequence}

Recently, several works, as shown in Table \ref{tab:accuracy}, argue that the long time series is more efficient for learning the continuous-time data. Usually, a long time series has shorter time steps which improve the accuracy of the result, but it still needs to overcome the higher sampling rate complexity. At the same time, the longer the sequence is, the more complex the neural network needs to be.  

A minimal number of works demonstrate efficient processing of lengthy continuous-time series. The usual approach is to normalise the long continuous-time data and train it in multiple batches with a subset of data for each batch. The Wavenet \cite{oord2016wavenet} is one of the most popular works on long sequence time series classification.  Dilated convolutional neural networks are often used to overcome the sequence length problem. Another new addition to this solution is phased-LSTM, a number of variations \cite{timelstm} of Phased-LSTM \cite{neil2016phased} have been proposed since 2016 to combat the larger sequence length problem.

\subsubsection{ Memory size }

Memory size is another problem that may arise due to the sequence length of the time series. For memory related problem, RNN is more popular than ANN and CNN due to its ability to recover memory from long past events. However, recently Temporal Convolution Networks (TCNs) has achieved better performance in terms of maintaining a more extended history than RNN.

\begin{table}[htp]
\centering
\caption{Accuracy on the copy memory task} 
\label{tab:accuracy}
\begin{tabular}{|p{0.7in}|p{1.0in}|p{0.9in}|} \hline 
\textbf{Model} & \textbf{Sequence Length} & \textbf{Accuracy} \\ \hline 
LSTM & 50 & $\boldsymbol{\mathrm{<}}$ 20\% \\ \hline 
GRU & \textbf{200} & \textbf{$\boldsymbol{\mathrm{<}}$20\%} \\ \hline 
TCN & 250 & 100\% \\ \hline 
\end{tabular}
\end{table}

Table \ref{tab:accuracy} shows that for the same data set as described in \cite{bai2018convolutional} TCNs maintains 100\% accuracy for all sequence lengths. On the other hand, the accuracy level falls below 20\% for LSTMs and GRUs after the sequence length reaches 50 and 200, respectively. For LSTM and GRU, the model quickly degenerates to random guessing as the sequence length grows. TCN is an excellent example of a hybrid deep learning model, where the attributes of CNN and RNN are combined efficiently. Large memory size is not a feasible choice for modelling neural network as that may cause to over-fitted training model.

\subsection{Pathological long-term dependencies}
Another sequence-related limitation of deep learning architecture for time series problem is pathological long-term dependencies \cite{chakraborty1992forecasting}. Some of the data points in a long time series dataset, as shown in \ref{eq:Xn}, can be related while some others are entirely independent. An excellent example described in \cite{ehr}, where N consecutive visits of a single patient can be related to the same physical problem where the range between two consecutive visits is comparatively short, while N consecutive visits of a single patient can be for entirely independent of each other, but in this case, the range between two consecutive visits is relatively broad. Usually, problems in long-term time series prediction problems have a long sequence of inputs. The future prediction for such input series can be only dependent on a few time-stamps at the end of the input series, and there is a long irrelevant part of random input in the sequence.  Healthcare shows this kind of problems more often than in other sectors. This kind of pathological long -term dependency makes the problem even harder to solve. The challenges increases relative with the length of the sequence (L) as longer sequences exhibit a broader range of dependencies.

\subsubsection{Influential feature selection}

Time series usually have a massive number of parameters. In order to model the time series, some of the parameters are more significant than others. Also, domain-specific feature strongly influences the accuracy of the result generated by different neural network model. As a result, different models demonstrate different performance for similar tasks with the same dataset and domain. Selection of the correct feature for each task is critical for time series modelling, especially in case of the time series classification problem. However, it is not feasible to design a domain-specific neural network for different tasks. Therefore, it is essential to distinguish the essential parameters and removes irrelevant parameters to reduce noise. If the length of the time series is L and M is the number of features to describe any problem, the scope of  M features is M$\times$L. For this reason, influential feature selection is crucial. Many methodologies are available for feature extraction from different static domain-specific data, for example, static images and computer vision. However, extracting features from continuous-time data still an open problem. One of the solution to improve the efficiency of a neural network model in terms of feature selection is that the design of the models needs to be data-driven. Several data-driven neural networks, as shown in Table \ref{tab:datadriven} demonstrate promising result for feature selection and optimising the performance of neural network models.

\begin{table}[htb]
\centering
\caption{Data driven neural network}
\label{tab:datadriven}

\begin{tabular}{|p{2.0in}|p{2.5in}|} \hline 
\textbf{Proposed work} & \textbf{Underline Neural Network}   \\ \hline 
\cite{grathwohl2018ffjord} & Neural ODE  \\ \hline 

\cite{sun2019neupde} & Neural PDE  \\ \hline 
 \cite{holl2020learning} & PDEs with Differentiable Physics \\ \hline 

 \cite{holl2020learning} & PDEs with Differentiable Physics  \\ \hline

 \cite{zhong2019symplectic} & Symplectic ODE-Net  \\ \hline 
\end{tabular}
\end{table}
   
Another approach to overcome the challenges of right feature selection is to use external methodologies to select features during the data collection and cleaning phase before training the neural model and later uses that selected feature during the training phase. Some of the efficient methodologies for feature selection are mentioned in Table \ref{tab:featureselection}.

\begin{table}[htb]
\centering
\caption{Efficient methodologies for feature selection}
\label{tab:featureselection}
\begin{tabular}{|p{1.0in}|p{2.0in}|p{2.0in}|} \hline 
\textbf{Proposed work} & \textbf{Methodology} & \textbf{Remark} \\ \hline 
\cite{ahn2020efficient} & Generic & time series classification. \\ \hline 

 \cite{jimenez2020feature} & multi-objective evolutionary algorithms (MOEA) along with evaluators based on the most efficient state-of-the-art regression algorithms & This method improves the performance of multivariate time series forecasting  \\ \hline 

\end{tabular}
\end{table}

\subsection{ Challenges faced by different Applications }

 There are several applications of sequential data modelling, for example, speech recognition, bioinformatics and human activity recognition. Different real-world application data suffers from different challenges mentioned in this section. Table~\ref{tab:ch} shows how different characteristics of continuous-time data affect different applications. 
 
\begin{table}
	
	\caption{Challeneges for proecessing continous-time data in different applications}
	\label{tab:ch}
	\centering
	\resizebox{\textwidth}{!}{
	\begin{tabular}{|p{3.8cm}|p{2cm}|p{2cm}|p{2cm}|p{2cm}|p{2.2cm}|}
	\toprule
	Challenges & HealthCare & Finance & Event Based IoT & Frequency Anasis \& Prediction & Classification \\ \hline
	Informative Missingness & $\checkmark$ && $\checkmark$&&  $\checkmark$\\ \hline
	Irregular sampling rate &  $\checkmark$&& $\checkmark$&& \\ \hline
	Higher sampling frequency & &&  $\checkmark$&& \\ \hline
	High Dimensional Data & && $\checkmark$&&  $\checkmark$\\ \hline
	Pathological  dependencies & $\checkmark$ & $\checkmark$&&& \\ \hline
	High Memory Consumption & && $\checkmark$&& \\ \hline
	Complex Computation &  $\checkmark$&  &  $\checkmark$ &  $\checkmark$&  $\checkmark$\\ \hline  
	\bottomrule
	\end{tabular}} 
\end{table}

As shown in the table~\ref{tab:ch}, healthcare data significantly suffers from informative missingness,  sampling irregularity and Pathological dependencies.  Table [\ref{tab:healthcare}] depicts the current trends of using continuous time series as the data model for health care problems. As a result, most research in the Healthcare industry uses RNN based neural network. Due to privacy, Healthcare industry suffers from a lack of real clinical data. Some researches use GAN based neural network to generate sample clinical data for further research. 


\begin{table}
	\centering
	\caption{Different Projects in Health care with time series}
	\label{tab:healthcare}
	\begin{tabular}{|p{2.5in}|p{2.9in}|} \hline 
		\textbf{Health care projects} &\textbf{Deep learning models}    \\ \hline 
		Heart failure prediction & Doctor AI \cite{doctorai}     \\ \hline 
		Patient Visit analysis & Time LSTM \cite{timelstm}, Phased--LSTM-D \cite{ehr}\newline Choi, Edward, et al.~[2]     \\ \hline 
		Multi-outcome Prediction  & DeepPatient \cite{miotto2016deep}     \\ \hline 
		Sleep stage classification & \cite{chambon2018deep}    \\ \hline 
	\end{tabular}
\end{table}  
 
Continous-time data is the primary block for the Internet of Things(IoT). The work presented  \cite{neil2016phased} has significantly contributed to opening new areas of investigation for processing asynchronous sensory events that carry timing information. This work has been extended and used in several applications \cite{kuremoto2014time, kuremoto2014forecast,yao2017deepsense, chen2018neural}. Some core characteristics of sensor-based data are explained by\cite{neil2016phased}, such as  \textit{Irregular sampling rate, Higher sampling frequency and High Dimension}. High Dimension is an unavoidable outcome of multivariate time series. Most sensor-generated data for weather, climate, automobile and other applications are generally multivariate. For example, sensor data from wearable consists of multiple channels of data, such as, step-count in smart mobile devices use the accelerometer and optical heart rate sensor. This sensor collects data of heart rate and step count. There are several use cases where data is coming from multiple channels.   For a more specific example, the heart rate monitor in smart mobile devices. In recent days, most of the research fields deal with a temporal sequence where the input channel is more than one. In addition, another significant limitation is that it requires long time series with higher frequencies and short time steps to predict an asynchronous future event from n previous events from the sequence. This processing possesses huge computation load and energy consumption. For multivariate sensor-generated time series, RNN and CNN based hybrid models \cite{fcnrlstm,hu2017state,grud,neil2016phased} are comparatively better in performance as well as accuracy.

The early and most popular application of time series is Time series frequency analysis for future prediction. Deep learning state of artworks for predicting future events have achieved impressive accuracy and performance. However, the existing works mainly based on an underlying assumption that the test data and the train data share a similar scenario \cite{zhang2017deep}. However, in reality, that is hardly true. Therefore, only a few relatively simple practical sequence models with a fixed data rate and some semi-supervised learning algorithms can provide satisfactory result under specific circumstances with the state of art research works.  In addition, the performance for these deep learning models for processing time series do not provide expected performance in real-world problem-solving.

Besides, time-series frequency analysis, continuous-time data has become very popular in terms of event prediction. Both regular as well as rare event prediction task, deep learning algorithms use time series analysis. However, this is a very challenging task as the time series is mainly a multi-variant series consists of heterogeneous variables. In addition, the dependency between variables is very complicated. Data are sparse and not sampled uniformly. The irregular sampling rate is very common in case of event prediction.
Furthermore, the series is asynchronous, which makes the task even more complicated. So far, the most used solution is to divide the problem up to a homogeneous variate. Therefore, several works\cite{neil2016phased, ehr, grud, timelstm} focus on dealing with the irregular data sampling rate and asynchronous time series.

\begin{figure}[h]
	\centering
	\begin{subfigure}[b]{0.4\textwidth}
		\includegraphics[width=\textwidth]{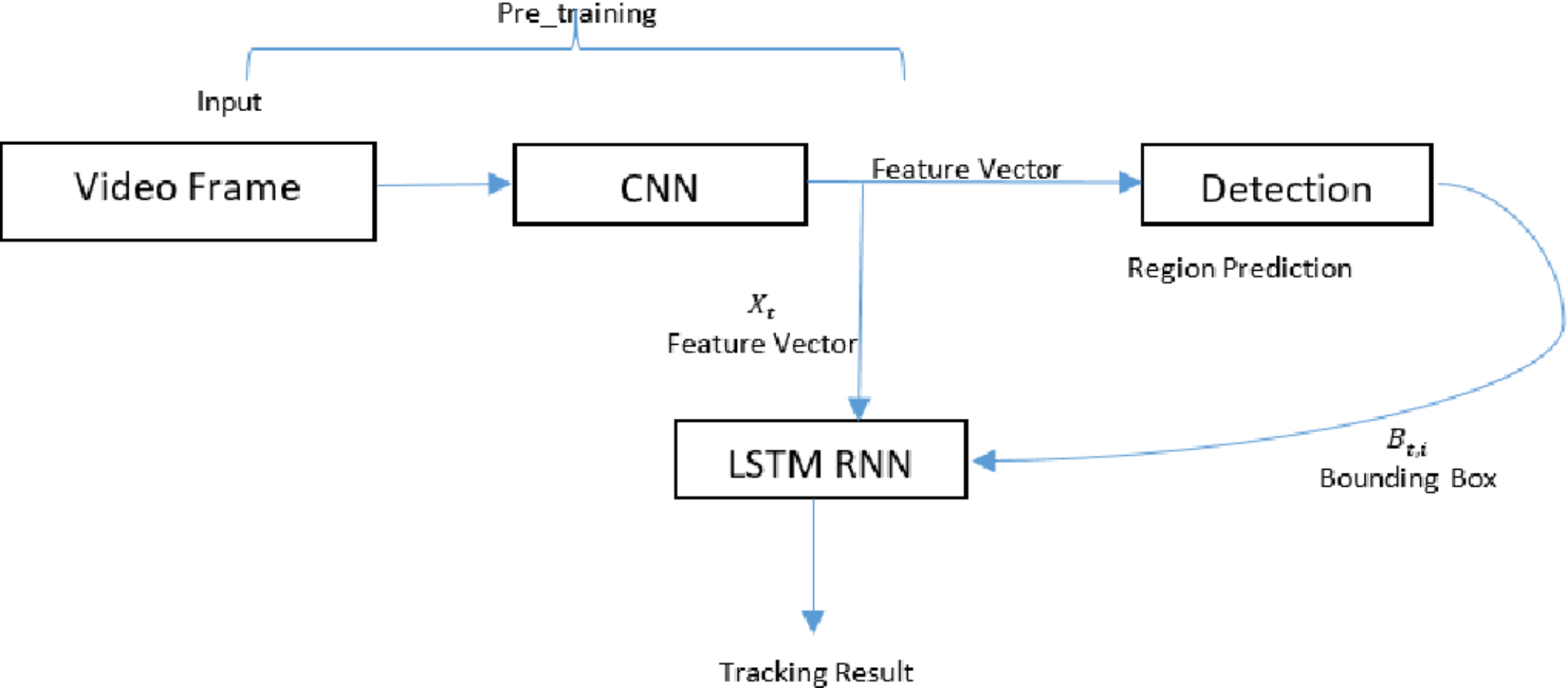}
		\caption{Hybrid CNN-LSTM Model}
		\label{fig:humantracking1}
	\end{subfigure}
	\begin{subfigure}[b]{0.4\textwidth}
		\includegraphics[width=\textwidth]{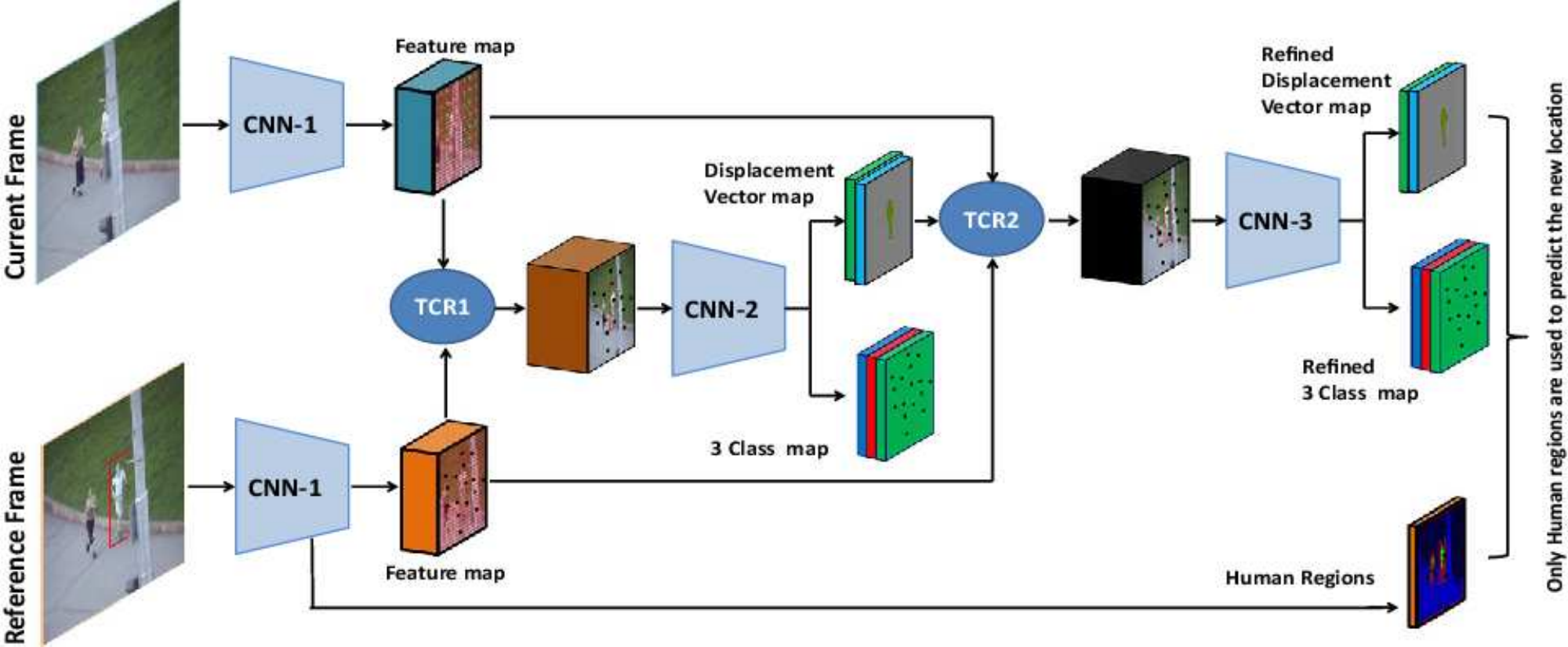}
		\caption{Cascaded CNN Model}
		\label{fig:humantracking2}
	\end{subfigure}
\end{figure}

Visual recognition is a perfect example of time series classification or detection from continuous-time data. Recently,  human activity recognition from the video has become a trendy field as an example of a temporal sequence learning using a deep learning algorithm. The main challenge in processing time sequence from the video is that the data is spatially imbalanced with irregular sampling rate. The quality of the image and the place of the object in the image make this even more challenging. In a single video frame, there can be a large number of parameter.
On top of that, additional noise in the data set makes it harder for a deep learning algorithm.   Hybrid neural network models are more successful than traditional vanilla models in this area. Combination of CNN and LSTM networks can successfully overcome the problem of spatially imbalanced data. Most cases, CNN models outperform other neural network models for visual recognition and image classification task for continuous-data. For example, a recent deep learning model \cite{humantracking} has proposed a similar architecture, where a CNN model is used for feature extraction phase. Later the feature vector an LSTM has been placed in the model architecture for human tracking detection and result tuning as shown in Fig[\ref{fig:humantracking1}]. Fig[\ref{fig:humantracking2}] shows a similar deep learning model \cite{Kumar2019ACC} where multiple CNN models are used as a cascaded CNN model.

\section{Different Neural Networks Models For Time Series Processing}
\label{sec:models}
\begin{figure}[h]
\centering\includegraphics[width=\textwidth]{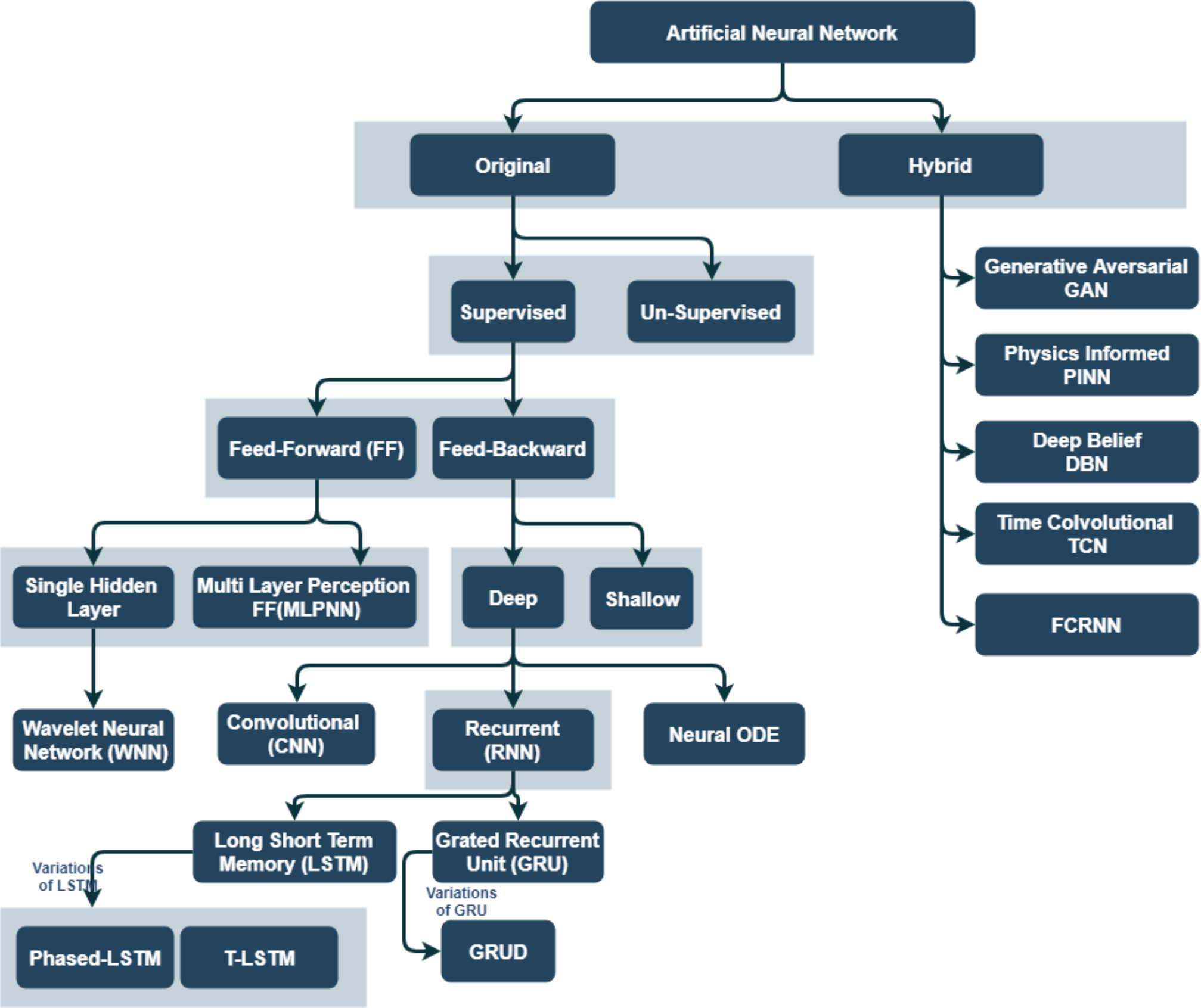}
\caption{Different types of neural network used in continuous time series modelling}
\label{fig:models} 
\end{figure}

In this section, we present different neural network architectures, which are generally used to learn a continuous time series. The different neural network models have their strengths and weaknesses in terms of modelling continuous time series. Fig.~\ref{fig:models} shows the primary types of neural network for modelling continuous-time series over decades. 

A single type of neural network is not suitable for all different solution. Based on the nature of the time series, the different model performs better than others.

\subsection{Artificial Neural Network (ANN)}

Wind speed prediction, energy prediction, financial time series forecasting,  and so many other continuous series prediction problems have been solved using the Artificial neural network (ANN). In early 2000, along with the CNN and the RNN based solution, ANN is a widely popular neural network for modelling non-linear time series. ANN does not need any previous assumption or linearity. The ANN successfully provides a mechanism to determine the number of neurons to be fired in the hidden layer. Some other characteristic features of ANN are listed as below

\begin{itemize}
\item	ANN can derive its computing power through massively parallel distributed structure and learn the corresponding model.
\item	The architecture of the model is straightforward. It consists of units. These units are connected using the symmetric weighted connection. This connection can be either one-directional or bi-directional. The weight of the connection can be inhibitory or excitatory.
\item	Usually a sigmoid function is used as the activation function, which is also known as the squashing function. Besides other activation function such as Linear, Atanh, Logistic, Exponential and Sinus are used as the activation function in the hidden as well as the output layer.
\item	As the ANN is simple in terms of design and system structure, it is possible to use different artificial intelligence algorithms, such as particle swarm optimisation algorithm (PSO) for learning as well as adjusting the parameters. Due to the simplicity of the underlying design of ANN, most ANN-based models are hybrid, where other algorithms are integrated to improve the performance of ANN. Some of the most popular methods used with ANN for time series prediction, classification and detection problems are as follows:

\begin{itemize}
\item Hidden Markov model (HMM)
\item Genetic Algorithms (GA)
\item Generalised regression neural networks model (GRNN)
\item Fuzzy regression models
\item Simulated Annealing algorithm (SA)
\item Echo State Network (ESN)
\item Particles Swarm Optimisation algorithm (PSO). \cite{egrioglu2016hybrid}
\item Elman Recurrent Neural Networks (ERNN) \cite{wang2016financial}
\item Hydrodynamic Neural Network 
\item Back Propagation Neural Network (BPNN)
\item Recurrent Multiplicative Neuron Model (RMNM)\cite{egrioglu2015recurrent}
\item Wavelet Neural network (WNN) \cite{nourani2019rainfall}
\item Hilbert-Huang transform (HHT) \cite{liu2019time}
\item  Multilayer Perceptron Networks (MLP)
\item Support Vector Machine  (SVM) \cite{ayazpour2018combined, khaledian2020simulating}
\end{itemize}

\item	Mean square error (MSE) and mean absolute percentage error (MAPE) is the loss functions used in ANN-based models. 
\item Artificial neural networks (ANNs) is very suitable for time series prediction, univariate time series forecasting, financial trend detection, wind speed and water fluctuation detection and other similar continuous-time problems. This neural network is also beneficial for pattern classification and pattern recognition \cite{bhardwaj2020comparative}.
\item ANN models are mostly data-driven without any initial assumption. 
\item ANN models usually perform better for time series prediction because of its ability to model any continuous functional relationship between input and output. 
\item ANN models can capture the underlying non-linearity of the system with highly non-linear dynamics by using a non-linear activation function. 
\item ANN model can adapt conditional training quickly. Therefore, the conditional time series forecasting is one of the most used areas for ANN. 
\end{itemize}

\subsubsection{Challenges of ANN}
\label{sec:challenges-ann}
The main challenges of ANN models are as follows

\begin{itemize}
\item \textit{Less Robust}: Noise data influences the optimisation as well as the robustness of the model. if the noise is not handled properly in the design of the network model, it can reduce the robustness for unknown input data. 
\item \textit{Optimum architecture}: The structure of the neural network model controls the performance of the model. therefore, it is crucial to determine the architecture or structure design of the network, including the number of layers, number of neurons and other parameters.
\item \textit{No previous memory}: ANN models can not preserve the previous state in the memory.
\item \textit{Back-Propagation Error}: 
\item \textit{Performance accuracy}: ANN models often suffer from reduced accuracy for multi-variate, missing pattern and other complex problem. So, the feasibility of the proposed methodology needs to be evaluated against a different kind of dataset. The prediction accuracy of ANN models directly depends on the right choice of parameters.
\item \textit{Selection of right related variable and parameters}: Different related variable such as the weight for each connection between neuron controls the accuracy and robustness of the model. Therefore, most ANN models pay attention to determine intelligent techniques to choose the right variable and evaluate them over time. Selection of right variable or parameters also expands to determine data transformation, initial values of the parameters, stop criterion. It is essential to choose the right parameter in order to avoid unnecessary overfitting.  
\item {Time series learning}: ANN models need to capture the main features of the continuous-time series and its generalisation.
\end{itemize}

\subsubsection{Recent ANN models to overcome different challenges }

\begin{figure}[h]
\centering\includegraphics[width=0.9\linewidth]{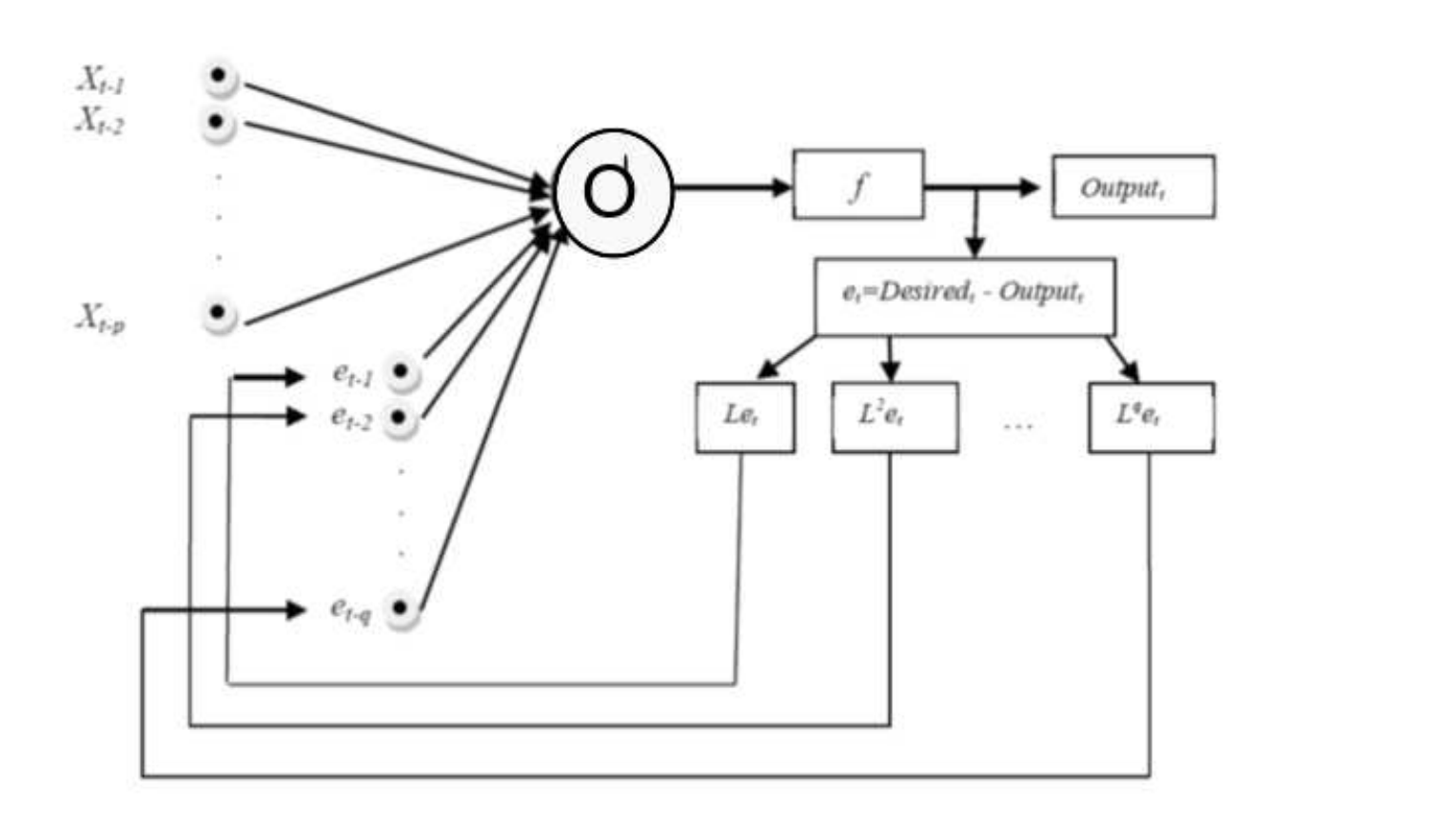}
\caption{The Architecture of RMNM \cite{egrioglu2015recurrent} }\label{fig:RMNM} 
\end{figure}

Recurrent Multiplicative Neuron Model(RMNM) proposed by \cite{egrioglu2015recurrent} is a recent work deal the lagged variable of error as input along with its recurrent structure in the case of learning time series. It overcomes the traditional challenge of ANN of deciding the number of neuron in the hidden layer. Fig~\ref{fig:RMNM} shows the RMNM model architecture. 

\begin{figure}[h]
\centering\includegraphics[width=0.9\linewidth]{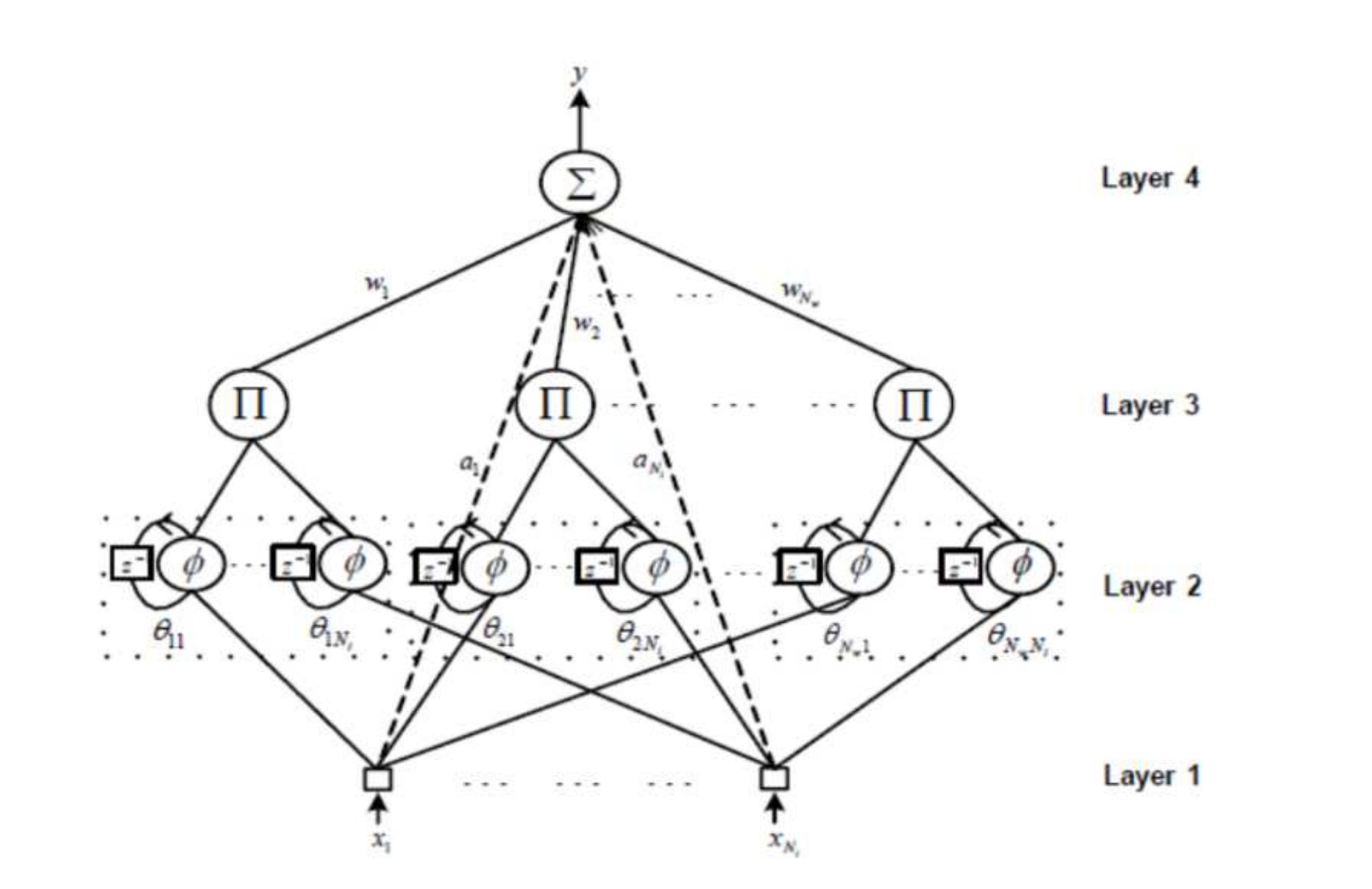}
\caption{Architecture of SRWNN \cite{yoo2005stable}}\label{fig:SRWNN} 
\end{figure}

Another significant limitation of ANNs is that it cannot preserve the previous state in memory. SRWNN \cite{yoo2005stable} proposes an outstanding solution for memory state preservation. Fig~\ref{fig:SRWNN} shows the architecture of SRWNN. SRWNN accepts N inputs but generates only one output. SRWNN usually consists of four different layers. The first layer accepts the input signal. The second layer consists of wavelet neuron (wavelon). Each of the wavelon has its self-feedback-loop which introduce the recurrent nature in traditional Artificial wavelet neural network (AWNN). The wavelon loss function described in \eqref{eq:4.2} compute the wavelet. The input of the second layer contains the memory term, as shown in \eqref{eq:4.3}. In this layer, the current dynamics of the system is conserved for the next step. The third layer is a product layer, and the fourth layer generates the output.
In comparison to AWNN, SRWNN shows that better performance is inefficient in solving the sequential temporal problem. SRWNN can store information temporarily. As a result, SRWNN is suitable for chaotic time series and non-linear system. 

Most of the above mentioned ANN models use hybrid methodologies in order to improve their performance. 
Despite AWNN's performance in sequence modelling, it has some limitations which often set obstacles for achieving a better result. The recent trends of using AWNN along with CNN or RNN, can overcome some of the limitations and achieve better performance. However, AWNN is not famous for its sophisticated time series modelling. 

\cite{tealab2018time} presented a comparative qualitative analysis of different ANNs proposed for forecasting time series in the time period between the year 2006 and 2016. This analysis shows that the most successful ANN model is hybrid and different algorithms are used to improve the performance as well as overcome different challenges of multi-variate time series. Hybrid ANN models demonstrate better performance as compared to basic ANN models, especially for dynamic datasets with a missing pattern, hidden non-linear and non-stationary characteristics.

Choose the right variable, weight, initial value for parameters are very important for artificial neural network modelling. Artificial intelligence algorithm such as Genetic Algorithms (GA), Simulated annealing algorithm and PSO algorithm are usually used in ANN to adjust the parameters of the model.

There are several approaches adopted by different ANN-based model in order to overcome the limitation of back-propagation. For example, \cite{chen2015short} combines three-dimensional hydrodynamic model in conjunction with an ANN to improve prediction accuracy. Integration with reinforcement learning (RL) is another solution for back-propagation. Usually, BP learning algorithm uses the traditional Euclidean distance to determine the error between the output of the model and the training data. However, in RL, the reward can be defined as an error zone. Therefore, the influence of the noise data can be reduced; at the same time, the robustness to the unknown data may be raised \cite{kuremoto2019training}.

There are different ANN-based neural network models, as shown in Table~\ref{tab:ann-models}. These recent models adopt different hybrid mechanism in order to overcome the challenges described in §\ref{sec:challenges-ann}.



\begin{longtable}{|p{.18\textwidth}|p{.26\textwidth}|p{.52\textwidth}|} 
  \caption{DIFFERENT KIND OF ANN}
  \label{tab:ann-models}\\
  \toprule
  \textbf{Proposed work} & \textbf{Challenges To Solve} & \textbf{Characteristics}\\
 
  \midrule
  \endfirsthead 
  \toprule
  \textbf{Proposed work} & \textbf{Challenges To Solve} & \textbf{Characteristics}\\
  
  \midrule
  \endhead 

Recurrent Multiplicative Neuron Model \newline (RMNM) \cite{egrioglu2015recurrent} \newline
ARMA TYPE PI-SIGMA ANN \cite{kocak2019new}& Adjust the parameter as well as related variable such as weights of neuron connection to avoid over-fitting and improve accuracy. & Use a PSO algorithm for learning. \\ \hline 

 Wavelet net \cite{yoo2005stable,nourani2019rainfall} & Support vector machine, Wavelet decomposition  & Usually Wavelet nets are hybrid models combining data-driven least square support vector machine (LSSVM), (ANN) and wavelet decomposition for disaggregation of continuous-time series. Instead of identifying the correlation among feature, \cite{yoo2005stable} model proposed a new parameter named as mutual information (MI)  which can be defined by Eq.~\eqref{eq:mi} 
 \begin{equation}\label{eq:mi}I(X, Y)=\iint d x d y \mu(x, y) \log \frac{\mu(x, y)}{\mu_{x}(x) \mu_{y}(y)}\end{equation}   \\ \hline 

Hybrid \cite{sun2019precipitation} & Learn hidden pattern or dynamics of the system & Uses different methods such as singular spectrum analysis (SSA) to identify hidden oscillation pattern in the data\\ \hline 

Multi-layer artificial neural networks \cite{zhang2019applying}& Multi-variate and interdependent dataset & Uses Mutual Information(MI)based feature selection process \newline In addition to the input layer and output layer of the neuron, this model has multiple layers of hidden units. These hidden units learn features and hidden dynamics of data with multiple layers of abstraction.  \\ \hline

Functional link artificial neural network (FLANN) \cite{mohapatra2019financial} & Multi-variate time series, where along with continuous-time data, location is also an influential parameter for future prediction &  Introduces a low-complexity artificial neural network and employing incremental and diffusion learning strategies \\ \hline 
 
 Hilbert-Huang transforms \cite{liu2019time} & Analysing the spectral and temporal information of non-linear and non-stationary time series & Introduce hybrid ensemble empirical mode decomposition (EEMD)-ANN that uses HHT to identify the time scales and change in continuous data. \\ \hline
 
 Hybrid ANN-(U-)MIDAS\cite{xu2019artificial} & Explore hidden non-linear pattern in raw mixed frequency data without information loss& introduce mixed data sampling to use raw input directly without any latent preprocessing. \newline Use back-propagation and chain rule to derive the gradient vector for optimisation algorithms\\ \hline

SciANN \cite{haghighat2020sciann} & Solution and discovery of partial differential equations (PDE) & Uses Physics Informed NN and Variational Physics-Informed Neural Networks With Domain Decomposition  \\ \ hline

Multiplicative Neuron Model \cite{egrioglu2015recurrent}& Use simple activation function for hidden layer & The output of time series at step $t$ , $y{t}$  for a typical MLP can be defined by Eq. \ref{eq:4.1}.
\begin{equation}
\label{eq:4.1} 
y_t\ =\ G\ \left(\alpha_0+\sum_{j=1}^{h}{\alpha_jF\left(\sum_{i=1}^{p}{\beta_{ij}y_{j-1}}\right)}\right)
\end{equation} Here, $\alpha$, $\beta$ are the weight for the neural network model. F and G is the activation function of the hidden and output layer, respectively, as described by \eqref{eq:4.2} and \eqref{eq:4.3}.

\begin{subequations}
	\begin{align}
	\label{eq:4.2} 
	F\left(x\right)=\frac{1}{1+e^{-x}} \\
	\label{eq:4.3} 
	G\left(x\right)=x
	\end{align}
\end{subequations} \\ \hline

Self recurrent wavelt neural network (SRWNN) \cite{yoo2005identification} & This model is suitable for chaotic time series described in §\ref{sec:characteristics} & The main self recurrent wavelet layer uses gradient descent method, which is derived from stability theorum along with adaptive learning. \\ \hline

\bottomrule

\end{longtable}

\subsection{Recurrent Neural Network(RNN)}

In the field of processing continuous sequence, the recurrent neural network is the pioneer.  As repetitive cell connected with a lateral connection creates a more extensive neural network in an RNN which process data sequentially, which is precisely what a temporal data sequence requires. Therefore, it is very efficient to model sequence structure. Even though RNN has demonstrated significant satisfactory result for modelling variable-length time sequence, as mentioned by \cite{structuralrnn} RNN architecture lack an intuitive spatiotemporal structure. To model time sequence is much more comfortable in RNN due to some novel feature of RNN as follows

\begin{itemize}
\item  RNN processes data sequentially one record at a time. It can model sequence with recurrent lateral connections.  

\item  It extracts the inherent sequential nature of time series.

\item  It can model different length of the series. 

\item  It supports time distributed joint processing.
\end{itemize}

Although RNN is one of the pioneer deep learning models for realistic high-dimensional time-series prediction tasks, it has some of its pitfalls. Some of the main drawbacks of RNN are as follows

\paragraph{\textit{ Vanishing gradient problem:}}\cite{langkvist2014review,sutskever2013training}. Most of the current work mainly focused on Vanishing gradient problem. There is already a significant improvement in the case of dealing with Vanishing gradient. Mostly, the continuous-time series can be extensive in length where the data point in the later of the sequence can have a long-term dependency with the data point at the earlier state of the sequence. So, much earlier time state can affect much later time state in the sequence.  In such a situation as the neural network can be very deep, even for a 100-layer neural network, it is challenging for the gradient to propagate back to affect the earlier state in the sequence.  The process of back-propagating to an earlier state of the sequence to modify the computation in the earlier state is just very much difficult. One of the major limitations of RNN and its different variant in the case of modelling a continuous time series is to overcome Vanishing gradient. Training a complex deep learning architecture with multiple layers of hidden units becomes critical due to the vanishing gradient problem when the error is usually backpropagated. LSTM \cite{hochreiter1997long} is an impressive addition to the RNN family which solves the vanishing gradient problem. Another way to solve the problem is to use unsupervised greedy layer-wise pre-training of each layer. This kind of regularisation helps to achieve better initialisation of a model.

\paragraph{ \textit{ Exploding gradient problems:}} Another significant limitation when using RNNs for time sequence modelling is the exploding gradient problem \cite{ref11}. This problem makes it very difficult to train a continuous time series using RNN.

\paragraph{ \textit{Butterfly Effect:}} For example, due to the dependencies among different time state in a long term time sequence, a minimal change in an iterative process of an RNN can result in a significant change in future time states after n iterations, where $n\rightarrow\infty $ \cite{lorenz1963deterministic}.  The main reason is that the loss function reacts significantly in small change and can be ultimately discontinuous, which would result in a complete failure in terms of predicting real-time continuous time series \cite{sutskever2013training}.  Some of the conventional approaches to deal with the butterfly effect are to use different data imputation methods, e.g., PCI, MissForest, KNN SoftImpute.

In this section, some of the critical architecture design of RNN are discussed.

\subsubsection{Long Short-Term Memory (LSTM)}
Vanilla RNN has been used for time sequence processing for past years. However, processing time sequence using vanilla RNN requires extensive modification in the architecture. Therefore, several works have been presented in the last few years, where Vanilla RNN mainly LSTM has been modified and extended to improve the performance and accuracy of time series processing. In an LSTM, it is possible to pass information from one cell to the next selectively. This model can process long term pathological dependency along with short term dependency, which is one of the most significant challenges in sequence modelling \cite{neil2016phased}.  Long Short-Term Memory (LSTM) \cite{langkvist2014review} has already become popular for both long-term and short-term time steps, for its performance and efficiency.

\begin{figure}[h]
	\centering\includegraphics[width=0.6\linewidth]{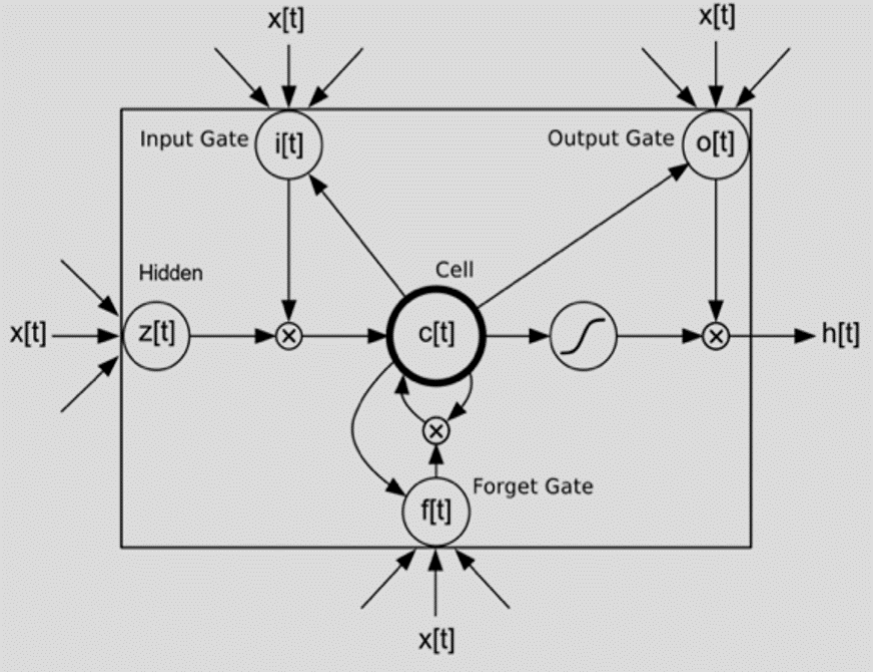}
	\caption{Traditional LSTM}\label{fig:lstm1} 
\end{figure}

As depicted in Fig.~\ref{fig:lstm1}, traditional LSTM cell is composed of 5 different non-linear components described in Eq.~(\ref{eq:4.4}) to Eq.~(\ref{eq:4.8}). They interact with each other through linear interaction, therefore, information can be transferred through back propagation across time. Each cell has an input gate i[t] Eq.~(\ref{eq:4.5}), output gate, o[t], Eq.~(\ref{eq:4.6}) and forget gate f[t], Eq.~(\ref{eq:4.4}). The cell state c[t], Eq.~(\ref{eq:4.7}) represents the current state of the cell and hidden output vector $\left[h\right[$, Eq.~(\ref{eq:4.8}) influence the computation of cell state of the next cell in the RNN sequence.

\begin{equation}\label{eq:4.4}
f\ \left[t\right]= \sigma_f(W_{xf}x\left[t\right]+h\left[t-1\right]W_{hf}+ w_cf \odot c\left[t-1\right]+b_f)
\end{equation}

\begin{equation}\label{eq:4.5}
i\ \left[t\right]= \sigma_i(W_{xi}x\left[t\right]+h\left[t-1\right]W_{hi}+ w_ci \odot c\left[t-1\right]+b_i)
\end{equation}

\begin{equation}\label{eq:4.6}
o\ \left[t\right]= \sigma_o(W_{xo}x\left[t\right]+h\left[t-1\right]W_{ho}+ w_co \odot c\left[t-1\right]+b_o)
\end{equation}

\begin{equation}\label{eq:4.7}
c\ \left[t\right]=f\left[t\right]\odot c\left[t-1\right]+i\left[t\right]\odot{\ \sigma_c(W_{xc}x\left[t\right]+h\left[t-1\right]W_{hc}+b_c)}
\end{equation}

\begin{equation}\label{eq:4.8}
h\left[t\right]=o\left[t\right] \odot \sigma_h(c\left[t\right])
\end{equation}

Here $x_{t}$ is the input vector and $h_{t}$ is the hidden output of LSTM cell at time t. Fig \ref{fig:lstm1} represents an overview of how different components of the different cell in traditional LSTM cell interacts with each other.

Several recent research works \cite{structuralrnn,ref13,kuremoto2014time,liu2014deep} have demonstrated significantly impressive result for modelling as well as processing time series.  The main idea behind the success of LSTM in the case of modelling time-series is its universal approximation ability for the open dynamical system \cite{yoo2005identification} which can learn the dynamics of the system and can generate it at the same time. For a deep LSTM neural model, it is not recommended to use a considerable memory size, as the model can be over-fitted by memorizing the input sequence data.

Several works \cite{neil2016phased, timelstm} have used different systems for time series modelling. For example, Shengdong Zhang et al. \cite{fcnrlstm} has used a set of the temporal sequence of observation collected in $n$ consecutive steps, as shown in Eq.~(\ref{eq:Xn}). They also have considered that the number of observations in each set is not constant. Then they have used a sequence of labels$\mathrm{\ }\mathrm{\{}e_{\mathrm{1}}\mathrm{,\ }e_{\mathrm{2}}\mathrm{,\ \dots .\ ,\ }e_n\mathrm{\ \ }\mathrm{\}}$ for X where $e_{\mathrm{i}}\{\mathrm{0,1\ }\mathrm{\}}$.  $e_{\mathrm{i}}\mathrm{=1}$ indicates the event is observed, on the other hand,  $e_{\mathrm{i}}\mathrm{=0}$ indicates absence of the event.  In  order to define the time gap between time duration between consecutive change in the value of $e_{\mathrm{i}}$, this work has used a target label sequence   y = $\{y_{\mathrm{1}}\mathrm{,\ }y_{\mathrm{2}}\mathrm{,\ }\mathrm{\dots .\ ,\ }y_n\mathrm{\ \ }\mathrm{\}}$ $y_{\mathrm{i}}=1$ refers that an event is observed in the next K time-steps, i.e. j+K j=i+1 ej $\mathrm{>}$ 0, and $y_{\mathrm{i}}=0$ refers otherwise. In this work, K referred to as monitor window. This monitor window helps to build the distribution of positive as well as negative events efficiently as the value of $e_{\mathrm{i}}$ indicates if the event at a particular time step is either positive or negative. Feeding the entire sequence to the network at once is not a recommended practice. This can impact the final outcome negatively. An alternative best practice is to feed the time series by batch. The time sequence of length T is divided into M subsequence; each of them is of maximum length of $\frac{T}{M}$. This technique is known as time sequence chopping. The chopped size is selected very carefully; otherwise, this technique sacrifices temporal dependency longer than $\frac{T}{M}$ time steps. This LSTM based architecture has been used for fault prediction on multi-variant heterogeneous time-series data. One of the most significant contributions of this work is that it handles time sequence modelling for rare events where the distribution of positive and the negative samples are highly imbalanced.

LSTM is the most used neural network model in terms of solving time series problems. Table [\ref{tab:rnn-models}] shows some recent work where LSTM has been used on order model time series. 

\begin{table}[htp]
	\begin{center}
		
		\caption{DIFFERENT TYPES OF RNN FOR MODELLING THE CONTINUOUS TIME SERIES}\label{tab:rnn-models}
		\begin{tabular}{|p{1.9in}|p{2.5in}|} \hline 
			\textbf{Neural network Model} & \textbf{Related work} \\ \hline 
			Vanilla RNN & Skip-RNN \cite{skiprnn}, Structural-RNN\cite{structuralrnn}, State-Frequency-RNN \\ \hline 
			LSTM & Phased LSTM \cite{neil2016phased}, Time LSTM \cite{timelstm}, Phased LSTM-D  \cite{ehr}\\ \hline 
			LSTM + CNN & Fcn-rLSTM \cite{fcnrlstm} \\ \hline 
			GRU & GRU-D \cite{grud} \\ \hline 
		\end{tabular}
	\end{center}
\end{table}

Recent trends are focusing on improving or modifying the architecture to model time series and related problems. Some of the promising models are described here as follows:

\subparagraph{Adaptive time scales Recurrent Units (ARU)}
Adaptive time scales Recurrent Units is a recent variation of RNN \cite{quax2020adaptive}, where, time constraints $\tau_{r}$ is used to drive the rate of firing $-r_{n}$ neuron of a neural network unit as shown in Equation.

\begin{equation}
\label{eq:tau}
\tau_{r} \frac{d r_{n}}{d t}=-r_{n}+f\left(I_{n}\right)
\end{equation} 

Here $f$ is a $sigmoid$ activation fucntion.

Eq~\eqref{eq:aru} shows the complete model for Adaptive time scales Recurrent Units (ARU). The input  signal $x$ is processed using weight matrices $W, U$ and $V$. N-dimensional Intermediate state variable $I$ and $r$ are used to capture synaptic couplig between neurons. $\alpha_{s}=\Delta t / \tau_{s} $ and $\alpha_{r}=\Delta t / \tau_{r}$ are rate constants to define the relation between state variable and time constraints $\tau_{r}$  and $\tau_{s}$. 
\begin{equation*}
\label{eq:aru}
\begin{array}{l}
	\left.\mathrm{I}_{t}=\left(1-\alpha_{s}\right) \mathrm{I}_{t-1}+\alpha_{s}\left(\mathrm{W} \mathbf{r}_{t-1}+\mathrm{U} \mathbf{x}_{t}\right)\right) \\
	\mathbf{r}_{t}=\left(1-\alpha_{r}\right) \mathrm{r}_{t-1}+\alpha_{r}\left(\mathrm{f}\left(\mathrm{I}_{t}\right)\right) \\
	\mathrm{y}_{t}=\mathrm{f}\left(\mathrm{V} \mathbf{r}_{t}\right)
\end{array}
\end{equation*}

The significant contribution of this paper is to identify the impact of slower rate constants can lead to longer memory retention, which enables to maintain memory for more extended periods. ARU is a suitable solution for vanishing gradients problem.

\subparagraph{Phased-LSTM}

\begin{figure}[h]
	\centering
	\begin{subfigure}[b]{0.4\textwidth}
		\includegraphics[width=\textwidth]{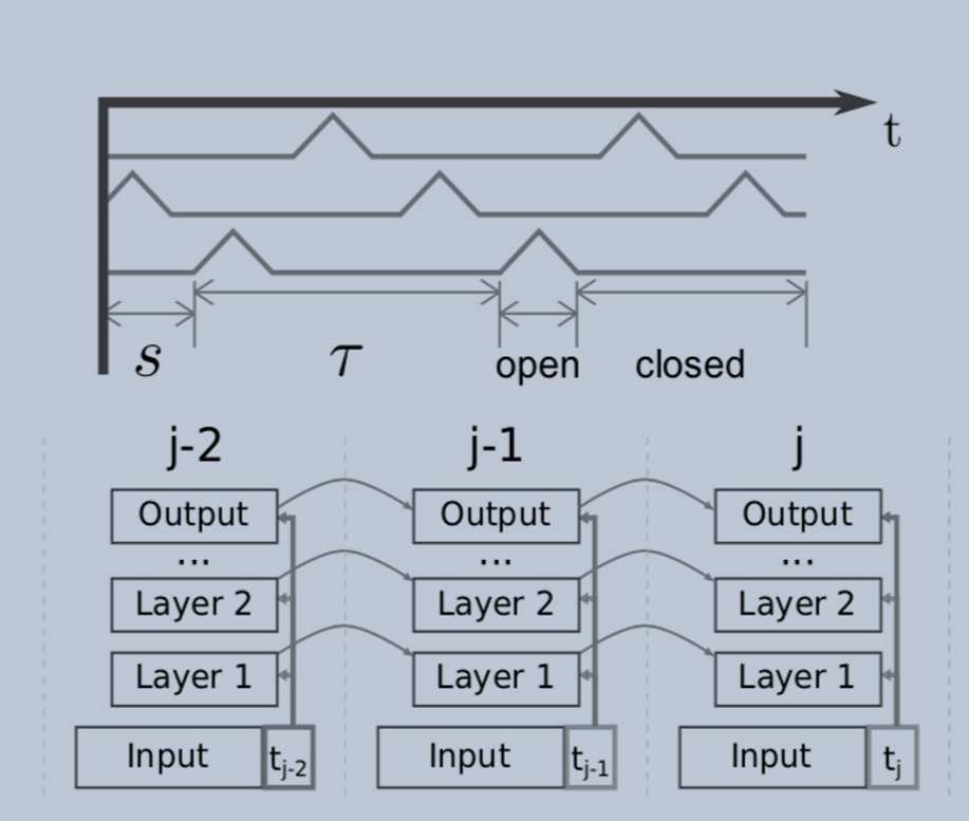}
		\caption{The openness of Phased LSTM}
		\label{fig:plstm1} 
	\end{subfigure}
	\begin{subfigure}[b]{0.4\textwidth}
		\includegraphics[width=\textwidth]{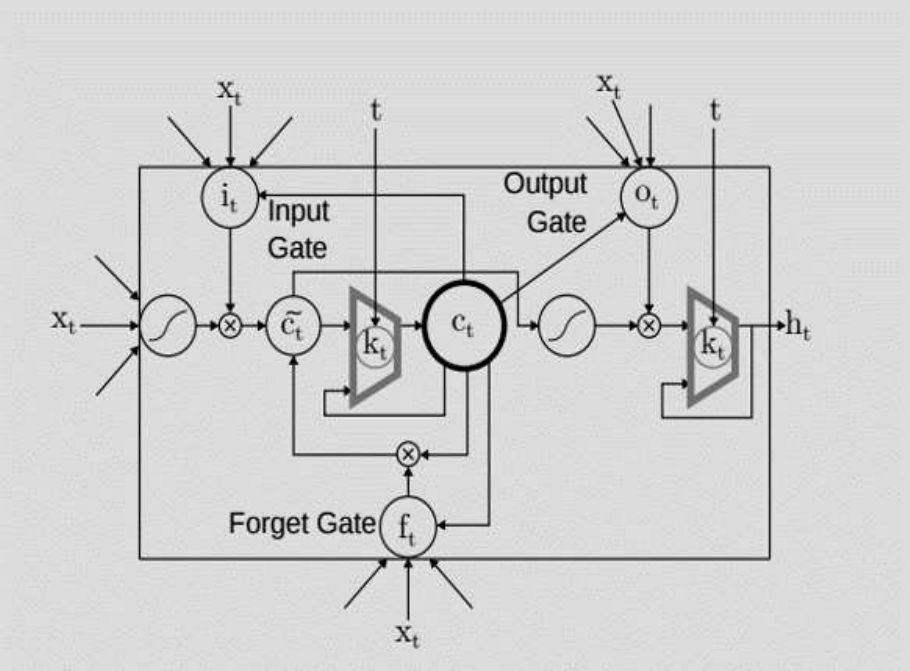}
		\caption{Phased LSTM Model Cell}
		\label{fig:plstm2} 
	\end{subfigure}
\end{figure}

Neil et al. \cite{neil2016phased} have added a new time gate to the existing LSTM and the timestamp which is the input for the new time gate controls the update the of the cell state, hidden state and the final output.  For improved performance, this model of LSTM has been designed to sample time rates captured in the model's active state only. Therefore, only when the time gate is open, associated sample time rates are captured. As a result, Phased-LSTM can attain a faster learning convergence in the training phase. The phased LSTM described in Fig [9] shows that the time gate decides the phase as either openness when the time gate value raises from 0 to 1 or closed when the corresponding time gate value drops from 1 to 0. The cell state is only updated when the time gate is open. Eq.~(\ref{eq:4.9}, \ref{eq:4.10} and \ref{eq:4.11}) describes the time gate in three different phases shown in Fig [\ref{fig:plstm1} and \ref{fig:plstm2}]

\begin{equation}\label{eq:4.9}
k_{topen1}=\ \frac{2\emptyset_t}{r_{on}}\ \ if\ \emptyset_t<\ \frac{1}{2}\ r_{on}\ \ and\ \emptyset_t=\frac{\left(t-s\right)\ mod\ \tau\ }{\tau}\ 
\end{equation}

\begin{equation}\label{eq:4.10}
k_{topen2}=2-\ \frac{2\emptyset_t}{r_{on}}\ \ if\ \frac{1}{2}\ r_{on}<\ \ \emptyset_t<\ \ r_{on}
\end{equation}

\begin{equation}\label{eq:4.11}
k_{tclose}=\ {\alpha\emptyset}_t
\end{equation}

The leakage rate ($\alpha$) is active when the time gate is closed.  The ratio ($r_{on}$) of time duration when the gate is open to the full period helps to control the open phase and close phase of the gate to fine-tune the sequence. Another important controlling parameter is $s$, which controls the shit between the open and close phase of the time gate ($\tau$) in order to determine the duration for each oscillation period.

The main advantages of this model are that it uses the time gate to convert the continuous-time series to a discrete sequence of sensor input with a minimum loss. Therefore existing LSTM can deal with the discretized sensory event. One of the limitations is that this work is still converting the continuous-time series to a discrete sequence; therefore, the accuracy is still not 100 percent.
Another limitation of this work is that it is not capable of de-sparsifying sparse sensor data. This model cannot reverse back to the original continuous time series from the discretized sequence, therefore, there are still missing input, and the input is not completely asynchronous. 

\subparagraph{ Phased LSTM-D}

For continuous sequence such as Electronic Health Records (EHR), where each feature is asynchronously sampled, it is challenging to apply Phased-LSTM straightforwardly. EHR data sampling can be used for disease diagnosis and prediction. In the case of EHR related problems for an individual patient, the time gap between two consecutive visits is irregular, and also all the clinical data about the patient collected during different visits are not the same. Some feature may be collected the first time and may not be collected in next time. Therefore, continuous data sampling is not only irregular but also has missing parameters. Therefore, S. J. Bang et al. \cite{ehr} has proposed Phased-LSTM-D based on Phased-LSTM \cite{neil2016phased} to achieve better prediction. Phased-LSTM-D is mainly designed to solve the irregularity of the sampling interval and the asynchronous of the feature.

\begin{figure}[h]
	\centering\includegraphics[width=0.8\linewidth]{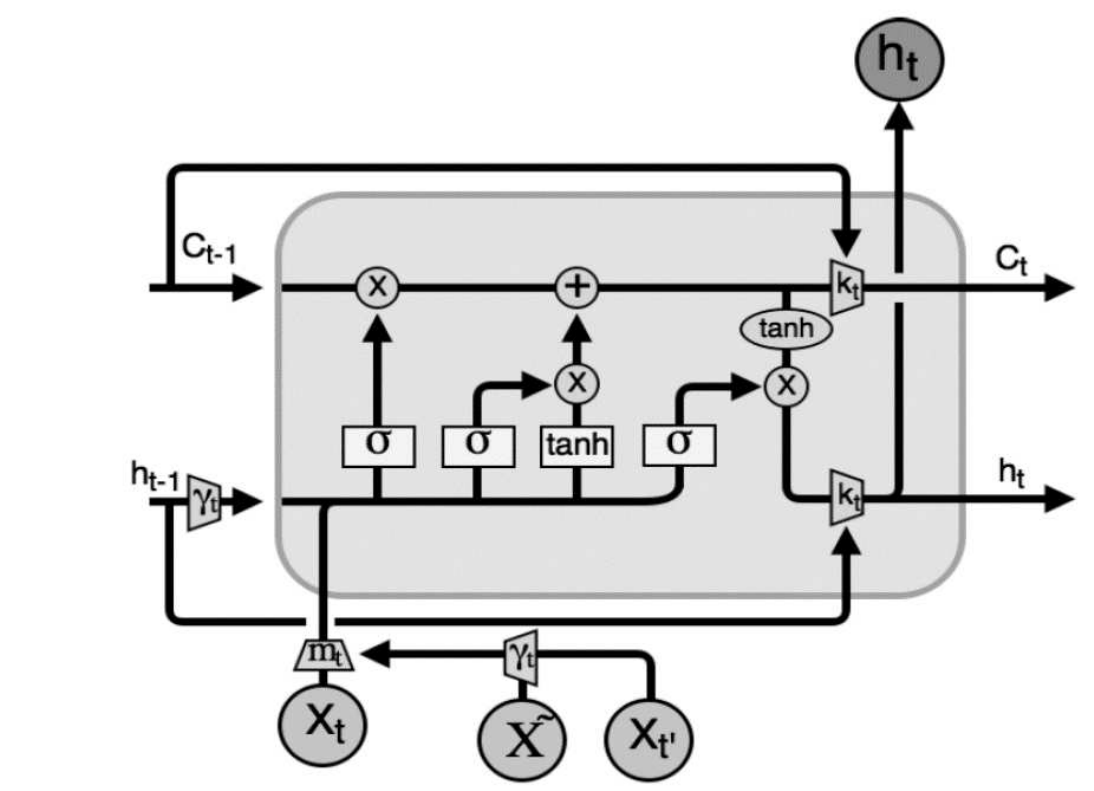}
	\caption{Cell of a Phased LSTM-D model}\label{fig:plstmd} 
\end{figure}

Phased-LSTM-D, as shown in Fig [\ref{fig:plstmd}], is a new predictive deep learning model that can simultaneously however individually deal with two missing patterns are as follows

\begin{itemize}
	\item  The irregularity of sampling interval 
	
	\item  Asynchronous of sampling features 
\end{itemize}

The new feature introduced in Phased-LSTM-D is the decay rater ($\gamma_t$) in existing Phased-LSTM. This additional parameter introduces a decay mechanism. Input and hidden states have corresponding decay rater $\gamma_x$ and $\gamma_h$ for each event-driven time point t. Therefore, if at some time t, the inputs are missing, the missing features are replaced with a weighted sum of the last measurement and average measurement. The decay rate at time t controls these. Similarly, the previously hidden state is also updated with a fraction of the previously hidden state-controlled with the decay rate. 

Phased-LSTM-D is designed to take advantages of both Phased-LSTM and GRU-D\cite{grud}. In this work, the new vector j which keeps track of the timestamps when the time gate of Phased-LSTM \cite{neil2016phased} was open, and the cell state was updated. Therefore, j and j-1 represent the timestamp when the cell was updated consecutively.  Phased-LSTM-D adopts Phased-LSTM to solve the irregularity and the additional requirement for Electronic Health Records (EHR) is to solve the missing feature at every update time point. In this work, in terms of EHR, the state of an individual patient (n) is updated at each time sequence for j =1, 2,3 {\dots}..Tn and only then the time gate is opened. At the same time, at each timestamp, the missing feature $x_j$ is replaced by a weighted sum of previous measurement ($x_{j-1}$) and an average measurement ($x_{j}$) of the same feature, controlled by decay rate ($\gamma_{j}$). According to Phased-LSTM-D, along with cell state which is the state of the patient, the previous hidden state ($h_{j-1}$) is also replaced by a fraction of the previous hidden state controlled by decay rate ($\gamma_{j}$). Therefore, when the time gate is open, the missing feature can be described as Eq.~(\ref{eq:4.12})

\begin{equation}\label{eq:4.12}
x_j^{(d)}=m_j^{(d)}x_j^{(d)}+\left(1-m_j^{\left(d\right)}\right)\gamma_{x_j}^{\left(d\right)}x_{j^\prime}^{\left(d\right)}+\ \ \left(1-m_j^{\left(d\right)}\right)(1-\gamma_{x_j}^{\left(d\right)}){\hat{x}}^{(d)}
\end{equation}

$x_j^{(d)}$ determines the value of the d-th variable at time $t_j$, The empirical mean of the d-th variable (${\hat{x}}^{(d)}$) and the previous value of the d-the variable $x_{j^\prime}^{\left(d\right)}$ are used as weight is an order to determine $x^{(d)}_j$ . The value for masking variable $m_j^{(d)}$ is either 1 in the case of the feature is not missing and 0 otherwise. Therefore, this masking variable controls if the measured value for the missing feature would be replaced or not. If $m_j^{(d)}$ =1, the observed value is used, otherwise, it is replaced by a weighted sum of   $x_{j^\prime}^{\left(d\right)}\ $ and ${\hat{x}}^{(d)}$.The decay rate is presented as a D-dimensional vector as in Eq.~(\ref{eq:4.13})

\begin{equation}\label{eq:4.13}
\gamma_{x_j}=\exp  {- \max(0,W_{\gamma_x}\delta_j+b_{\gamma_x})  }
\end{equation}

$W_\gamma$ and $b_\gamma$ are modelled parameters which are trained jointly. $\delta_j$ the time difference between two consecutive time sequences when the missing feature was observed and available. In this work, the decay rate also controls the change in the hidden state as shown in Eq.~(\ref{eq:4.14}), 

\begin{equation}\label{eq:4.14}
h_{j-1}=\gamma_{h_j}\ \odot\ h_{j-1}
\end{equation}
In addition to existing Phased-LSTM \cite{neil2016phased}, Phased-LSTM-D has used masking variables and time interval vectors for training additional hidden layers to estimate the decay coefficient. Therefore, the proposed work enable simultaneous decay in input and hidden state along with model training. This work introduces Phased-LSTM in the field of predictive modelling for continuous-time series with sampling irregularity. The benefits of this proposed work over another approach it that it introduces additional masking and time interval to impute missing feature instead of the direct imputation of the missing features. This significantly improves the performance and accuracy.

\subparagraph{ Time LSTM}
The recommendation system can demonstrate another important use of continuous-time series prediction. As recommendation systems need to analyze the insight of some feature in the continuous or discrete sequences of user's action, the different recurrent neural network architectures are recently being used for recommendation system. In the case problem domain such as recommendation system, it is not enough to only consider the order of different time steps as, like a traditional recurrent neural network, it is also necessary to establish a relationship between the sequence of time series. In a recommendation system, it is necessary to store the user's short-term actions. Such as if a user has purchased a flight ticket, he/she might need to have a recommendation for the hotel. This is an example of short-term dependencies. At the same time, some recommendations needed to consider a user's past action a long time ago. Time-LSTM \cite{timelstm} designed time gate of Phased-LSTM to capture both long-term as well as a short-term dependency to predict the next time sequence or recommendation for the user. However, Time-LSTM has adopted time gate of Phased-LSTM to models the timestamp time gate may or may not implicitly captures the time intervals and the consecutive data points, where Time-LSTM explicitly model time intervals and thereby capture the relation between two consecutive observations or data points. The main component of Time-LSTM is the time gate which has been designed in three different ways to capture the relation between time intervals. 

Based on the nature of the time gate, Time-LSTM has three different models as follows

\begin{enumerate}
	\item  Time-LSTM 1
	
	\item  Time-LSTM 2
	
	\item  Time-LSTM 3
\end{enumerate}

\subparagraph{ Time-LSTM 1}

This Time-LSTM architecture has an only one-time gate which is used to exploits time intervals simultaneously both long term and short term. The newly introduced time gate for Time-LSTM 1 for m-th sequence in the time series can be described as Eq.~(\ref{eq:4.15}).

\begin{equation}\label{eq:4.15}
T_m=\sigma_t\left(x_m W_{x t}+\sigma_{\Delta t}\left(\Delta t_m W_{t t}\right)+b_t\right)
\end{equation}

\begin{figure}[h]
	\centering\includegraphics[width=0.8\linewidth]{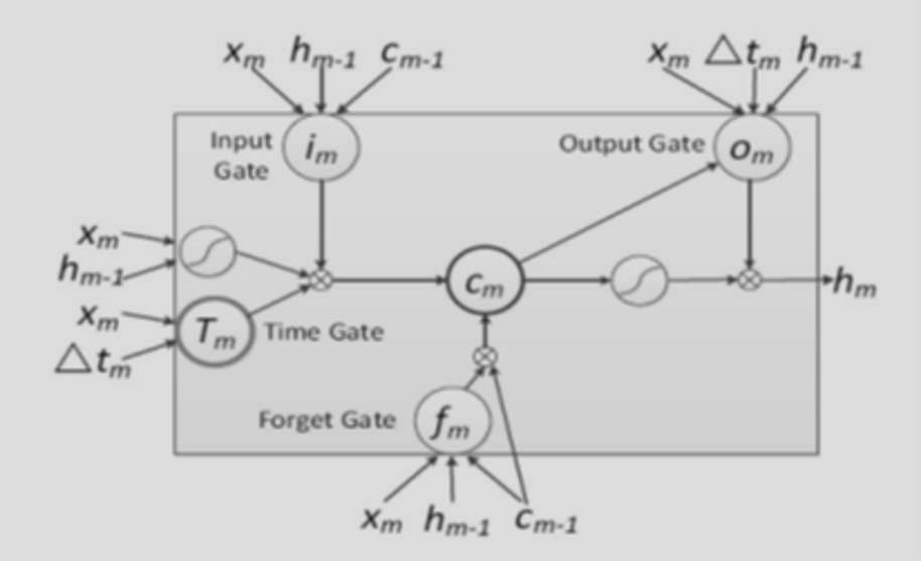}
	\caption{Architecture of Time-LSTM 1}\label{fig:tlstm1} 
\end{figure}

In this proposed architecture of LSTM, the cell state and the output gate has been modified as follows Eq.~(\ref{eq:4.16}) and Eq.~(\ref{eq:4.17})

\begin{equation}\label{eq:4.16}
c\left[m\right]=f\left[m\right]\ \odot c\left[m-1\right]+i\left[m\right]\ \odot{T\left[m\right]\ }\odot\ \sigma_c\left(x\left[m\right]W_{xc}+h\left[m-1\right]W_{hc}+b_c\right)
\end{equation}

\begin{equation}\label{eq:4.17}
\begin{aligned}
o_m= & \sigma_o\left(x_m W_{x o}+\Delta t_m W_{t o}+h_{m-1} W_{h o}+w_{c o} \odot c_m+b_o\right)
\end{aligned}
\end{equation}
Here $\vartriangle t$ is the time interval and $\sigma$  is the sigmoid function.  Therefore, the cell state c[m] is filtered by the input gate and the time gate. The time interval $\vartriangle t$ is stored in the time gate (T[m]) and later transferred to the cell to compute the cell state. Similar to the Phased-LSTM, time gate control the input vector for any m-th sequence. Fig[\ref{fig:tlstm1}] shows the architecture of Time-LSTM 1. 

\subparagraph{ Time-LSTM 2}

The second type of Time-LSTM architecture has two different time gates. One time gate is used to exploits time intervals to capture the short-term time value, and another time gate is used to save the time intervals to model long-term time values. The first-time gate for Time-LSTM 2 for m-th sequence in the time series can be described as Eq.~(\ref{eq:4.18}). T1 [m] deals with only short-term time values. 

\begin{equation}\label{eq:4.18}
T 1[m]=\sigma_{1}\left(x[m] W x 1+\sigma_{\Delta t}\left(\Delta t[m] W_{t 1}\right)+b_{1}\right), \text { s.t } W_{t 1} \leq 0
\end{equation}

Where the second gate can be described as follows Eq.~(\ref{eq:4.19}).

\begin{equation}\label{eq:4.19}
T 2[m]=\sigma_{2}\left(x[m] W x 2+\sigma_{\Delta t}\left(\Delta t[m] W_{t 2}\right)+b_{2}\right)
\end{equation}

Time-LSTM 2 also modifies the actual cell state, output gate and the hidden state equations (\ref{eq:4.6}, \ref{eq:4.7} and \ref{eq:4.8}) of the cell as follows in Eq.~(\ref{eq:4.20}, \ref{eq:4.21}, \ref{eq:4.22} and \ref{eq:4.23}). 

\begin{equation}\label{eq:4.20}
\begin{aligned} \widetilde{c_{m}}=& f_{m} \odot c_{m-1} \\ &+i_{m} \odot T 1_{m} \odot \sigma_{c}\left(x_{m} W_{x c}+h_{m-1} W_{h c}+b_{c}\right) \end{aligned}
\end{equation}

\begin{equation}\label{eq:4.21}
\begin{aligned} c_{m}=& f_{m} \odot c_{m-1} \\ &+i_{m} \odot T 2_{m} \odot \sigma_{c}\left(x_{m} W_{x c}+h_{m-1} W_{h c}+b_{c}\right) \end{aligned}
\end{equation}

\begin{equation}\label{eq:4.22}
\begin{aligned} o_{m}=& \sigma_{o}\left(x_{m} W_{x o}\right.\\ &+\triangle t_{m} W_{t o}+h_{m-1} W_{h o}+w_{c o} \odot \widetilde{c_{m}}+b_{o} ) \end{aligned}
\end{equation}

\begin{equation}\label{eq:4.23}
\begin{aligned} o_{m}=& \sigma_{o}\left(x_{m} W_{x o}\right.\\ &+\triangle t_{m} W_{t o}+h_{m-1} W_{h o}+w_{c o} \odot \widetilde{c_{m}}+b_{o} ) \end{aligned}
\end{equation}

\begin{equation}
h_{m}=o_{m} \odot \sigma_{h}\left(\widetilde{c_{m}}\right)
\end{equation}
The new cell state $\widetilde{c_{m}} $ is later aggregated with output state to compute hidden state. Time-LSTM 2 can distinguish the impact of current recommendation and future recommendation, therefore, it can give a better prediction for rare event prediction and where the interval between consecutive event can be longer. Along with the input gate i[m], the time gate $T1\left[m\right[$ is another filter for  ${\sigma }_c\mathrm{(}x\left[m\right]W_{xc}\mathrm{+}h\left[m\mathrm{-1}\right]W_{hc}\mathrm{+}b_c\mathrm{)}$.

The intermediate cell state $c\left[m\right[$ store the result which is later passed to output gate $o\left[t\right[$ to generate the hidden state $h\left[m\right[$ which is used for short term time sequence prediction.

On the other hand, $T2\left[m\right[$ stores the time interval $\mathrm{\delta}t\left[m\right[$ and later pass it to cell state $c\left[m\right[$ as the current cell state for computing the next cell state in a recurrent neural network, $\mathrm{\Delta}t\left[m\right[$ is transferred to $c\left[m+1\right[$, $c\left[m+\right[$ and so on. Therefore, $T2\left[m\right[$ influences the long-term time series prediction.

\begin{figure}[h]
	\centering\includegraphics[width=0.8\linewidth]{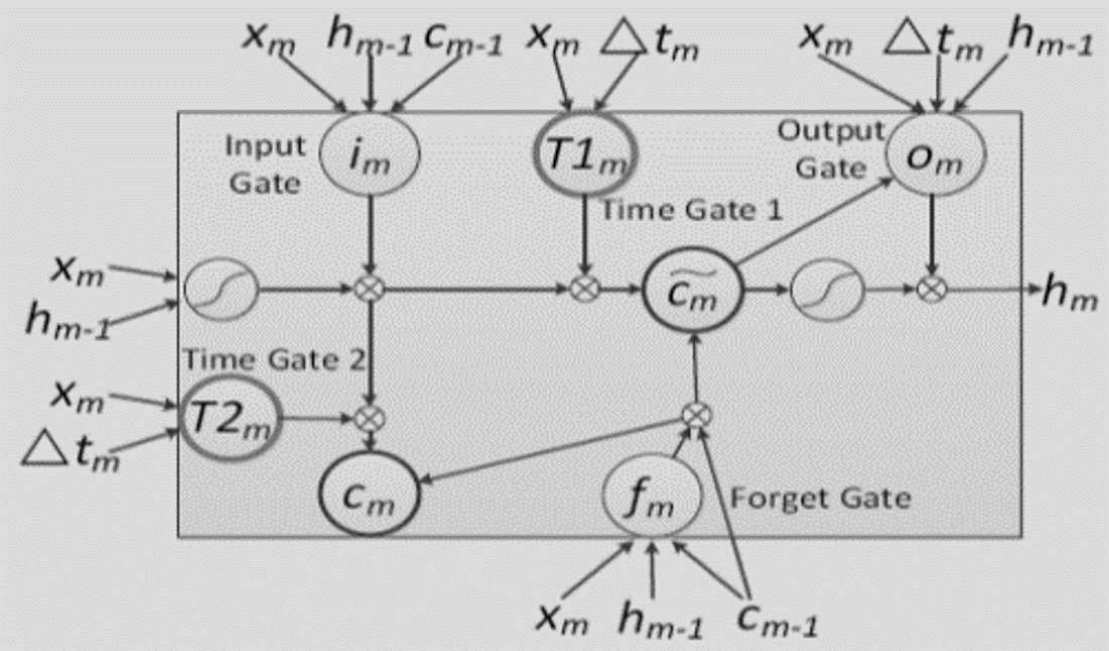}
	\caption{Architecture of Time-LSTM 2}
	\label{fig:tlstm2} 
\end{figure}

Therefore, if $\mathrm{\Delta}t\left[m\right[$ is smaller, according to Eq.~(\ref{eq:4.20}), $T1\left[m\right[$ would be larger and $T1\left[m\right[$, the cell state has a larger influence on short-term time intervals. On the other hand, if $\mathrm{\Delta}t\left[m\right[$ is a larger input vector  $x\left[m\right[$, the influence and correspondingly would be smaller. Fig [\ref{fig:tlstm2}] shows the architecture of Time-LSTM 2.    

\subparagraph{ Time-LSTM 3}

The third type of Time-LSTM architecture has two different time gate. One time gate is used to exploits time intervals to capture the short-term time value, and another time gate is used to save the time intervals to model long-term time values. The first-time gate for Time-LSTM 2 for m-th sequence in the time series can be described as Eq.~(\ref{eq:4.24} and \ref{eq:4.25}). T1 [m] deals with only short-term time values.

\begin{equation}\label{eq:4.24}
\begin{aligned} \widetilde{c_{m}}=&\left(1-i_{m} \odot T 1_{m}\right) \odot c_{m-1} \\ &+i_{m} \odot T 1_{m} \odot \sigma_{c}\left(x_{m} W_{x c}+h_{m-1} W_{h c}+b_{c}\right) \end{aligned}
\end{equation}

\begin{equation}\label{eq:4.25}
\begin{aligned} c_{m}=&\left(1-i_{m}\right) \odot c_{m-1} \\ &+i_{m} \odot T 2_{m} \odot \sigma_{c}\left(x_{m} W_{x c}+h_{m-1} W_{h c}+b_{c}\right) \end{aligned}
\end{equation}

\begin{figure}[h]
	\centering\includegraphics[width=0.8\linewidth]{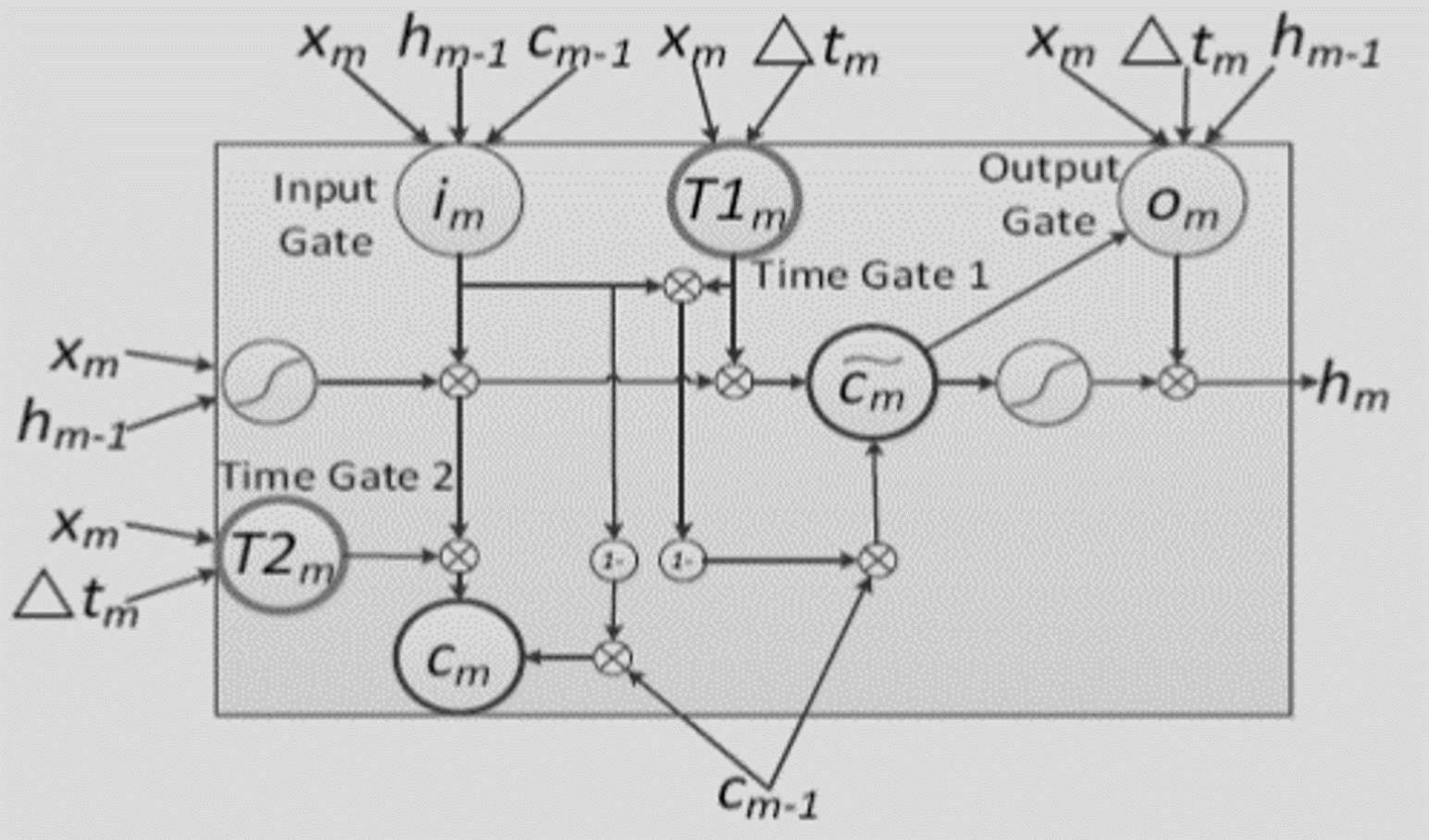}
	\caption{Architecture of Time-LSTM 3}
	\label{fig:tlstm3} 
\end{figure}

\subparagraph{ FCN-rLSTM}

FCN-rLSTM. \cite{fcnrlstm} proposed a spatial-temporal neural network by combining FCN \cite{fcn} and LSTM \cite{hochreiter1997long} to overcome the limitation of continuous sequence due to co-relation between the sequence. This model has used a residual learning framework. 

\begin{figure}[h]
	\centering\includegraphics[width=0.8\textwidth]{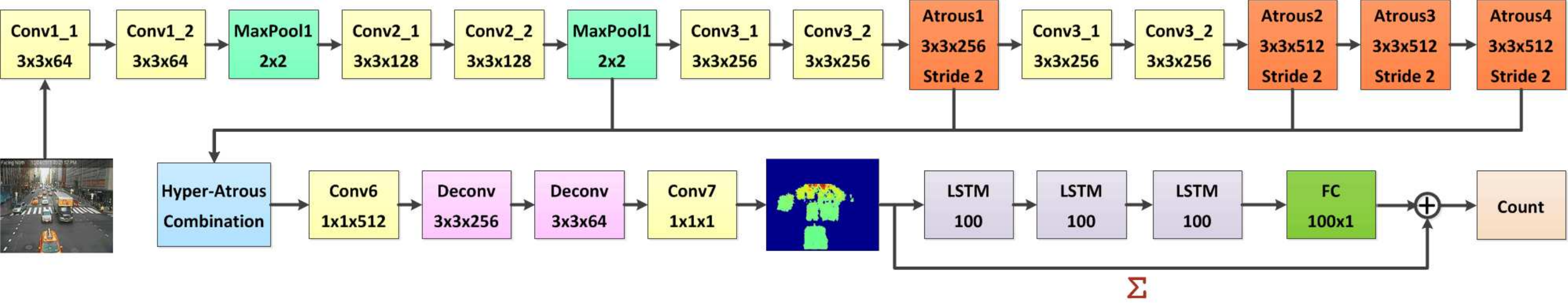}
	\caption{Network architecture and parameters of FCN-rLSTM}
	\label{fig:fcnrlstm} 
\end{figure}

\subsubsection{Grated Recurrent Unit (GRU)}
\begin{figure}[h]
	\centering\includegraphics[width=0.6\linewidth]{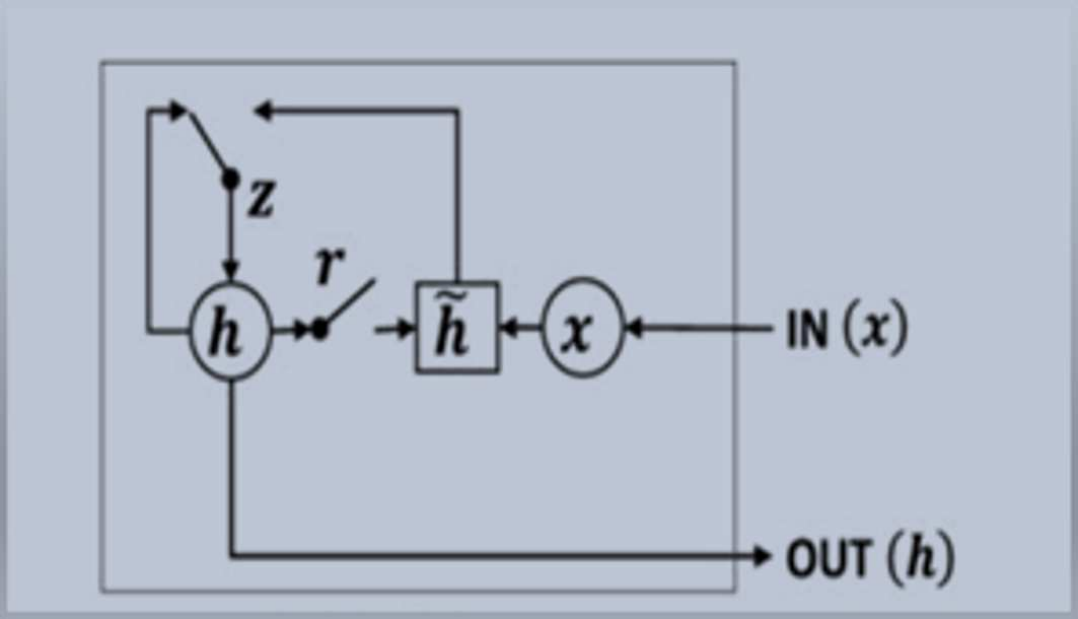}
	\caption{ Reset and update gate of GRU}
	\label{fig:gru2} 
\end{figure}

Grated Recurrent Unit (GRU) is the other recurrent neural network \cite{dey2017gate}. Fig \ref{fig:gru2} explains the architecture of the GRU cell.

The GRU cell usually has two gates reset gate (r) and update gate (z). Fig [\ref{fig:gru2}] explains the gate performance in a GRU cell.

There are several variations of GRU, as shown in Table\ref{tab:gru-gates}. The total number of the parameter (N) for different GRU cell differs from one to another, and it is dependent on the dimension of hidden state, n, and m, the dimension of the input vector. The reset gate ($ r_{t} $) decides how much data needs to be forgotten, where the update gate ($ z_{t} $) determines the amount of data to be passed to the future cell.

\begin{table}[h]
	\caption{\textsc{Different type of GRU cell}} \label{tab:gru-gates}
	\resizebox{\textwidth}{!}{%
	\begin{tabular}{lccc}
		\toprule
		& \multicolumn{2}{c}{Gates} \\
		\cmidrule(rl){2-3}
                \multicolumn{1}{c}{Cell}
		& Reset gate  & Update gate
                & Number of Parameters \\
                \cmidrule(rl){1-1}
		\cmidrule(rl){2-3}
                \cmidrule(rl){4-4}
		GRU          	 & $ \displaystyle r_{t}=\sigma\left(W_{r} x_{t}+U_{r} h_{t-1}+b_{r}\right) $ & $ \displaystyle z_{t}=\sigma\left(W_{2} x_{t}+U_{2} h_{t-1}+b_{2}\right) $  & $\displaystyle N=3\left(n^2+nm+n\right)$  \\
		GRU-1		 	 & $ \displaystyle r_{t}=\sigma\left(U_{r} h_{t-1}+b_{r}\right) $ & Gru-1  & $\displaystyle N=2\ \times nm $  \\
		GRU-2		 	 & $ \displaystyle r_{t}=\sigma\left(U_{r} h_{t-1}\right)  $ & $ \displaystyle z_{t}=\sigma\left(U_{2} h_{t-1}\right) $  & $ \displaystyle N=2\left(n\times \left(m + n\right)\right) $  \\
		GRU-3		 	 & $ \displaystyle r_{t}=o\left(b_{\mathrm{r}}\right) $ & $ \displaystyle z_{t}=0\left(b_{l}\right)  $  &   \\
		GRU-D		 	 & $ \displaystyle r_{t}=\sigma\left(W_{r} \hat{x}_{t}+U_{r} \hat{h}_{t-1}+V_{r} m_{t}+b_{r}\right) $ & $ \displaystyle z_{t}=\sigma\left(W_{z} \hat{x}_{t}+U_{z} \hat{h}_{t-1}+V_{z} m_{t}+b_{z}\right)  $  &   \\
		\bottomrule
	\end{tabular}}
\end{table}

GRU1 consider only previous hidden state and bias to determine the output, while GRU2 use the previous hidden state to compute the output and GRU3 uses only bias to compute the cell state. GRU1, GRU2 and GRU3 use less parameter than traditional GRU.

GRU-D \cite{grud} is a variant of GRU which is mainly focused on solving the informative missingness problem of recurrent neural network in general. This model uses a decay mechanism to decay the input over time towards the empirical mean. So, it does not use the value of the observation ($x_t$) as it is. Therefore, instead of $x_t$, it uses the vector defined by Eq.~\eqref{eq:4.24}. Here, $\hat{x}^d_t$ is the last observation of the $d$-th variable and $x^d$ is the empirical mean of the $d$-th variable.

\begin{equation} \label{eq:4.26} 
\hat{x}^d_t = m^d_t x^d_t + (1-m^d_t) (\gamma^d_{x_t} x^d_t +(1-\gamma^d_{x_t}) x^d)
\end{equation} 

Another decay mechanism used in this mode to understand the missingness. Eq.~\eqref{eq:4.27} defines the hidden state decay.
\begin{equation} \label{eq:4.27} 
\hat{h}_{t-1} = \gamma_{t-1} \odot h_{t-1}
\end{equation} 

The equivalent equation for Eq.~\eqref{eq:4.24} and \eqref{eq:4.25} for GRU-D can be defined using Eq.~\eqref{eq:4.28} and \eqref{eq:4.29}.

\begin{subequations}
\begin{align} \label{eq:4.28}
\tilde{h}_t &= \tanh\left(W_h\hat{x}_t+U_h\left(r_t\odot \hat{h}_{t-1}\right) + Vm_t + b_r\right)
\\
\label{eq:4.29}
h_t &= \left(1 - z_t\right) \odot \hat{h}_{t-1} + z_t \odot \hat{h}_t
\end{align}
\end{subequations}

In Table~\ref{tab:gru-gates} the equation for GRU-D has some new parameters, e.g., $V_r$, $V_z$ and $V$. The masking vector $m_{t}$ is fed into the model. This model succeeds to have a better result in terms of informative missingness with the help of different~interpolation and imputation methods, e.g., Mean,~Forward,~Simple, K- nearest neighbor \cite{batista2002study}, CubicSpline \cite{de1978practical}, matrix factorization (MF)\cite{koren2009matrix}, principal component analysis (PCA) model \cite{josse2012handling}, missforest \cite{stekhoven2011missforest}, softimpute \cite{mazumder2010spectral} and multiple imputations by chained equations (MICE)\cite{azur2011multiple}.

The main limitation is that it may fail or provide inefficient result in the absence of informative missingness or if the relationship between the missing patterns and the prediction tasks is not clear enough for the model. Therefore, this model is not suitable for unsupervised learning.  The significant contribution of the model is to use trainable decay to use the correlation between the missing pattern and the task to determine better prediction result. In terms of time and space complexity, this model is the same as RNN.

\subsection{Convolutional Neural Network (CNN)}
Convolutional neural networks are widely used for image processing. However, recently several works propose to use CNN for modelling continuous time series and related problems such as anomaly detection, structured time series classification. Usually, for time series problems, 1D CNN architecture is widely used. Another popular approach is to combine with LSTM and other RNN models and design and hybrid-CNN model for time series processing.  Most of the related works focus on modifying the existing traditional design of CNN to make it adjustable to work with the unique behaviour of continuous time series. Some other works are focused on combining the different useful feature from both CNN and RNN and develop an architecture of the hybrid neural network. Table~\ref{tab:cnn} describes an overall trend in CNN models for time series problems \\

Convolutional neural network (CNN) mainly used for classification and augmentation problem categories. In most of the cases, CNN is used in a hybrid model. CNN mainly suffers over-fitting less than other fully connected networks. Still, there are some specific time series augmentation techniques are used to improve the performance. One of the most used data augmentation technique for CNN model is Window Slicing (WS)\cite{le2016data} to slicing up the time series into a number of slices and train those slices in batches and each slice from a test time series is classified using the learned classifier. another data augmentation often used for CNN based time series classification algorithms are window wrapping (WW)\cite{le2016data} 

\begin{table}[ht]
\centering
\caption{CNN models and associated time series applications}
\label{tab:cnn}
\resizebox{\textwidth}{!}{%
\begin{tabular}{|p{0.15\textwidth}|p{0.22\textwidth}|p{0.2\textwidth}|p{0.22\textwidth}|}
\hline
CNN Model & Design Type & Application & Performance \\ \hline
TCN \cite{bai2018empirical} & Fully convolutional neural network (FCN) & Sequence Modelling & Demonstrate better performance than vanilla RNN  \\ \hline
CNN-LSTM \cite{livieris2020cnn} & Hybrid  & Prediction &  \\ \hline

CNN-RNN \cite{canizo2019multi} & Hybrid  & Anomaly Detection /Pattern Recognition & Works for multi-variate time series \\  \hline

CNN \cite{brunel2019cnn}& CNN & Classification & \\ \hline

CNN-FCM \cite{liu2020cnn}& Hybrid (CNN+ Fuzzy Cognitive Block) & Prediction & Shows promising performance for anomaly detection during training for handling data deviation\\ \hline

\end{tabular}%
}
\end{table}

As the non-stationary and dynamic data of any time series evolve, most of the deep learning models suffer from the uncertainty during training. One of the most challenging problems is maintaining the stability of the neural networks. Different CNN based models mentioned in table \ref{tab:cnn} shows that often hybrid approaches, where CNN can contribute to sequence modelling, data processing or classification phase with the help of other artificial intelligence techniques. Therefore, hybrid models are getting attention from different research area for time series problem domain. Combining LSTM and CNN is also a popular hybrid model for time series classification, anomaly detection or pattern recognition and prediction problems. 

In this section, some popular CNN model for time series problem domains is discussed. 

\subsubsection{ Temporal Convolutional Neural Network (TCN)}

Most of the recent deep learning models for the time series are mainly based on RNN architecture (LSTM or GRU). However, convolutional neural network architecture also has a contribution to learning time sequence for deep learning. TCN\cite{bai2018empirical} outperform standard recurrent architectures across a broad range of sequence modelling tasks.

\begin{figure}[h]
\centering\includegraphics[width=0.9\linewidth]{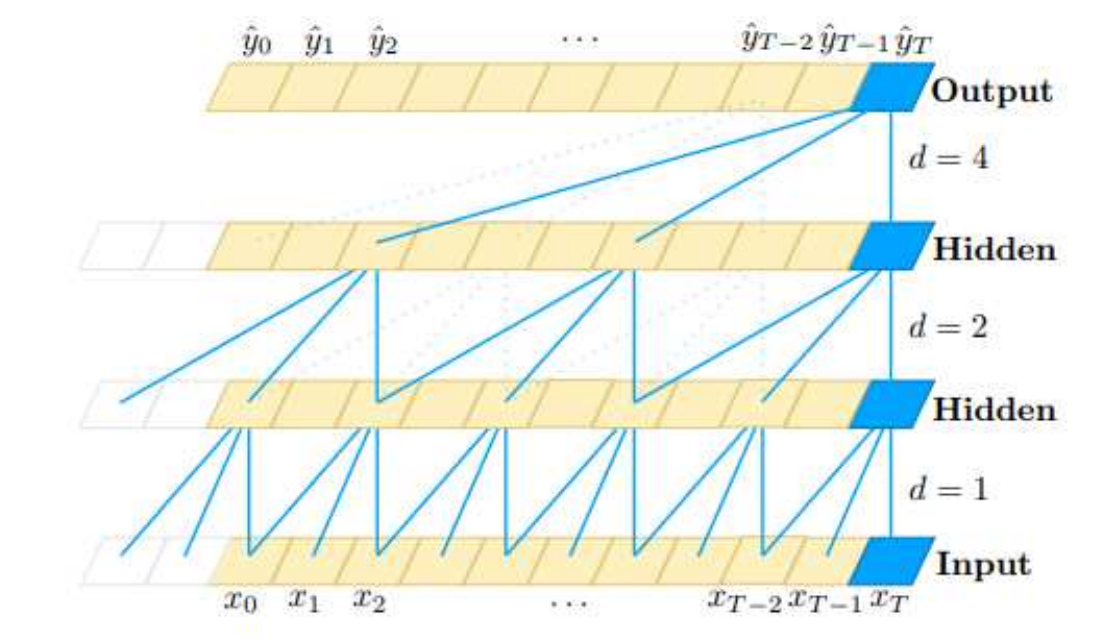}
\caption{Architectural element in TCN \cite{bai2018empirical}}\label{fig:tcn} 
\end{figure}

TCN is based on one fully convolutional neural network (FCN) where each hidden layer and the corresponding input layer are of the same length. TCN uses causal convolutions, where the elements from time t and earlier in the previous layer convolved the output at time t. Fig.[\ref{fig:tcn}] shows the architectural element of the TCN architecture. The input sequence $x_0$ … … $x_T$, and corresponding outputs $y_0$… … $yT$ at each time. d is the dilation factor, k is the filter size.
The main advantages of TCN over standard RNN are as follows
\begin{itemize}
\item	Any long-time sequence can be processed at the same time, unlike RNN which needs sequential processing.
\item	Authors claimed that TCN is better than RNN in memory size management. Filter size and dilation factor provide better control over memory usage.  In this work, it has also shown that RNNs likely to use up to a multiplicative factor more memory than TCNs.
\item	TCN battles against the vanishing gradient problem by imposing back-propagation path different from the temporal direction of the sequence.
\item	TCN also supports arbitrary input length.
\end{itemize}

Instead of all the advantages mentioned above of TCN over standard RNN, there is some limitation of TCN, which are it requires comparatively more memory during the evaluation period. Besides, TCN performs very poor in the absence of a large number of parameters regarding the time series. The primary concern of TCN is to minimise the memory size, reduce time by imposing parallel processing and overcome the vanishing gradient problem.

\subsubsection{ Spiking Neuron Neural Network(SNNN)}

Michael et al. \cite{pfeiffer2018deep} proposed a relatively new variation of convolutional neural network SNNN, which has a straightforward mechanism for ignoring data points. Instead of complex computation, additional gates like RNN or phase parameters like Phased-LSTM \cite{neil2016phased}, SNNN only computes whenever sudden bursts of activity are recorded and create more spikes, but when the recorded information is too little, it does not compute.

SNNNs are a suitable candidate for modelling Spatio-temporal event-based time sequence. As this model only compute data when there is a high spike of neurons, it is very power efficient. This model ignores irrelevant data when the spike is too low, therefore for real-time data analysis; the model can learn data in a more data-efficient training than other traditional neural networks model. Unlike DNN and DCN, it does not wait for the entire input sequence to be finished to start the learning phase.  As a relatively new model, there is no benchmark data to measure efficiency and accuracy vastly.
Moreover, designing training algorithms for this model is also very difficult. This model is not a suitable candidate for all sorts of problem. This model can achieve result faster in the case of classification problems.

\subsection{ Hybrid Deep learning model}

The recent trend in deep learning research is to find out an optimised way to design a model architecture that can have a combination of all useful attributes of RNN and CNN. Table [\ref{tab:hybridmodels}] shows some works in this field.

\begin{table}
\centering
\caption{Different type of hybrid model proposed for time series modelling}
\label{tab:hybridmodels}
\begin{tabular}{|p{0.25\textwidth}|p{.45\textwidth}|} \hline 
\textbf{Proposed Work} & \textbf{Used Neural Network Model} \\ \hline 
Shi et al. \cite{xingjian2015convolutional} & Convolutional  LSTM  \\ \hline 
Shaojie et a. l \cite{bai2018empirical} & TCN  \\ \hline 
Bradbury el al. \cite{quasirnn} & Quasi-RN \\ \hline 
 
\end{tabular}
\end{table}

\subsubsection{ Deep Belief Neural Network(DBN)}

\begin{figure}[h]
 \centering
  \begin{subfigure}[b]{0.45\textwidth}
    \includegraphics[width=\textwidth]{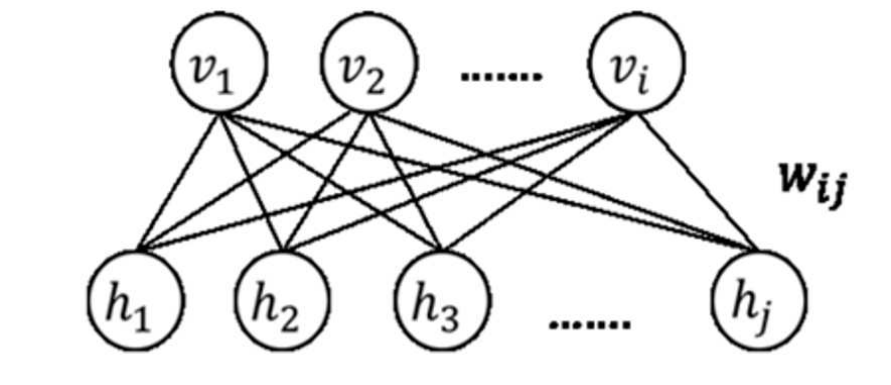}
    \caption{Architecture of RBM}
    \label{fig:rbm} 
  \end{subfigure}
  \begin{subfigure}[b]{0.45\textwidth}
    \includegraphics[width=\textwidth]{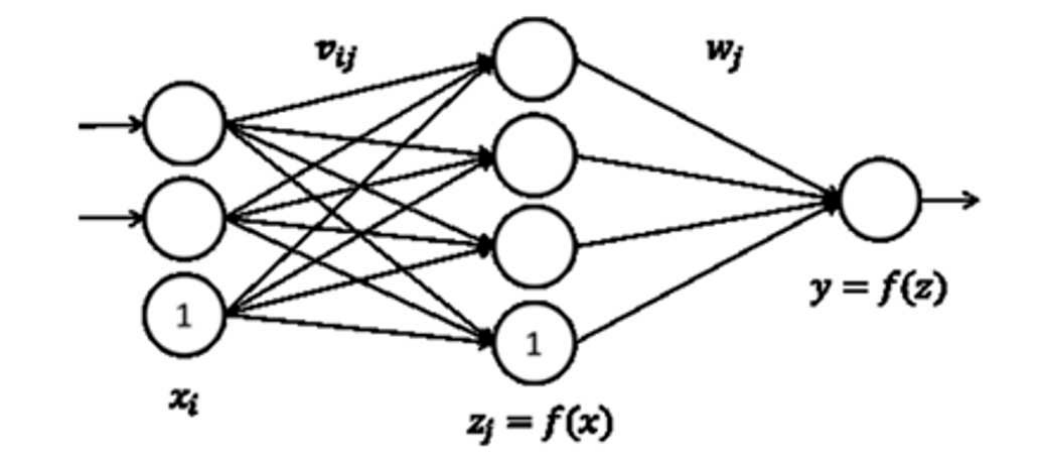}
    \caption{Architecture of MLP system}
	\label{fig:mlp} 
  \end{subfigure}
\end{figure}

T. Hirata et al. \cite{hirata2015time} has proposed a novel time series prediction model using ARIMA and DBN. There are some other works \cite{chandra2012cooperative,yao2017deepsense} based on DBN. DBN only takes care of the non-linear feature of time series. Therefore, it is not a suitable model for linear time series features. The fundamental component of DBN is Restricted Boltzmann machine (RBM) \cite{ref14} and Multi Later perceptron (MLP) \cite{yao2017deepsense}. DBM may also consist of only multiple RBM as well as \cite{chen2018neural}. RBM is one of the well-known mechanism to reduce the number of dimensions. Fig \ref{fig:rbm} shows the architecture of RBM, which consists of one visible layer (v) and one hidden layer (h). Each visible unit is connected with the asymmetric weight matrix (w${}_{ij}$).  One important feature of RBM is that the connection between units is bidirectional. Each unit outputs either 1 or 0.

Another core component of DBN is an MLP. MLP usually has three layers as like other neural networks, such as input, hidden, and output layer, as shown in Fig [\ref{fig:mlp}]. The activation function for MLP is a logistic sigmoid as shown in Eq.~(\ref{eq:4.32}) . If $w_ix_i\ >\ b_i$. the neuron of hidden layer get fired, where $x_i$ is the input, $w_i$ is the weight, and the threshold is the biases ($b_i$).

\begin{equation}\label{eq:4.32}
f\left(x\right)=\ \frac{1}{1+\exp{\left(-\frac{x}{\epsilon}\right)}}
\end{equation}

\begin{figure}[h]
\centering\includegraphics[width=0.8\linewidth]{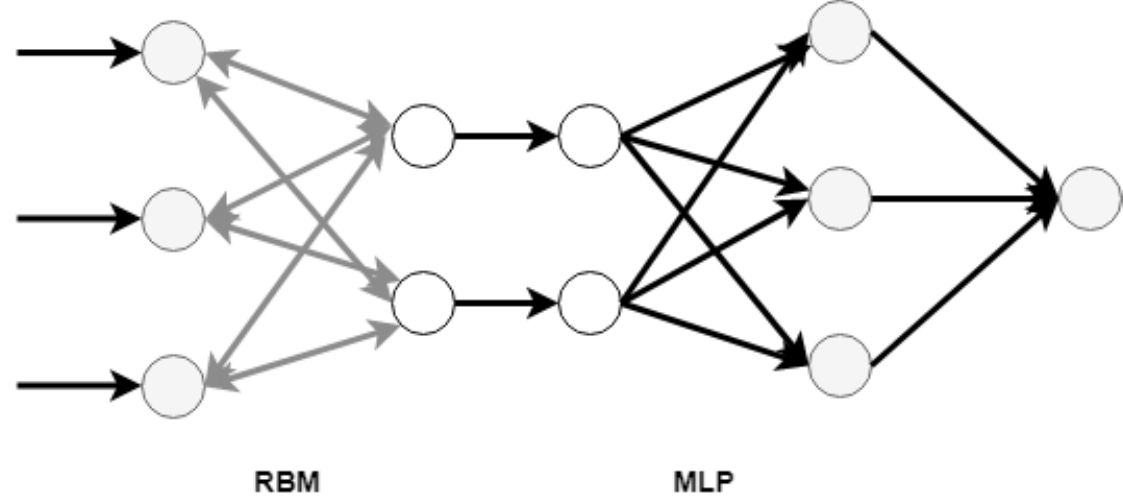}
\caption{Architecture of DBN}\label{fig:dbn3} 
\end{figure}

Here $\epsilon $ is the gradient.
Fig [\ref{fig:dbn3}] describes the architecture of DBN, where the inputs are feed into an RBM and output is learned by an MLP.

The time series prediction work proposed in \cite{kuremoto2014time} re-write the time series as the following equation Eq.~(\ref{eq:4.33}).

\begin{equation}\label{eq:4.33}
\ \ \ \ y\left(t\right)=L\left(t\right)+N(t)
\end{equation}

Here L is the linear part of the time series where N is the non-linear feature of the time series y(t). In \cite{kuremoto2014time}, an ARIMA has been used to determine the prediction value from the L(t), and a DBN predicted the prediction value from N(t). Therefore, the output time series from \cite{kuremoto2014time} can be described using Eq.~(\ref{eq:4.34}). Here $\hat{L}\left(t\right)$ is the predicted result from ARIMA and the $\hat{N}\left(t\right)$ is the predicted result of DBN. 

\begin{equation}\label{eq:4.34}
\hat{y}\left(t\right)=\hat{L}\left(t\right)+\hat{N}(t)
\end{equation}

This kind of hybrid neural network shows the highly precise result for chaotic time series. 

Cen et al. \cite{chen2018neural} have proposed another type of DBN which is a combination of DBN and a non-linear kernel-based parallel evolutionary support vector machine (ESVM). This model extracted different feature from different relevant parameters of the time series using DBN and combined them with the original attribute of the series; this combined attribute vector is imported to a non-linear SVM for training as well as solving classification or prediction problem. A parallel PSO algorithm is used to optimise SVMs. Different optimisation function has been used to overcome the curse of dimensionality caused by the high dimensionality of $\omega$, the weight vector.  

\subsubsection{ Deep-Sense Model}

\begin{figure}[h]
\centering\includegraphics[width=0.8\linewidth]{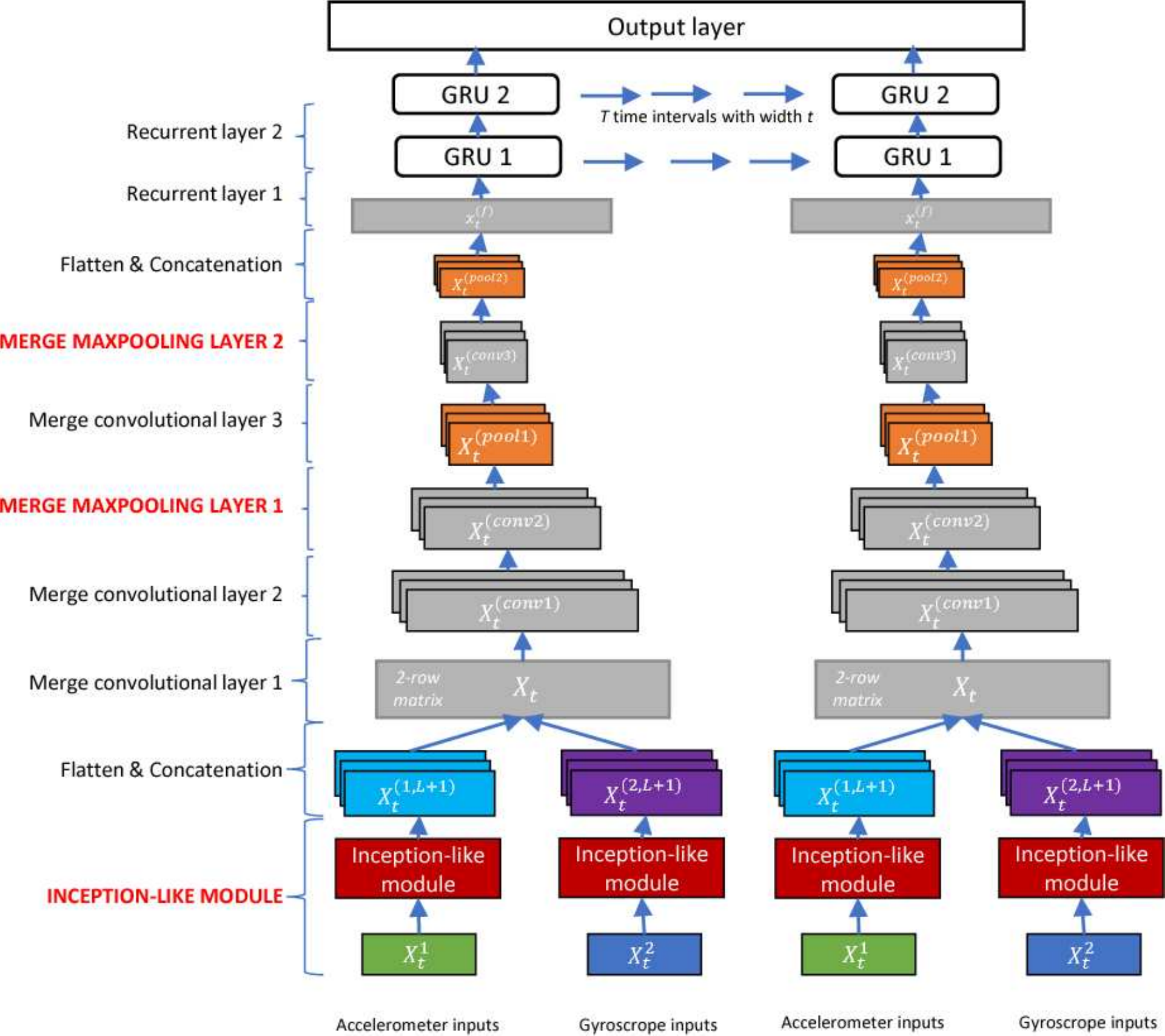}
\caption{Architecture of Deepsense - a hybrid neural network model \cite{chen2018neural}}
\label{fig:deepsense} 
\end{figure}

This neural network model is a perfect example of a hybrid model where the nature of recurrent and convolutional model have been used together for better performance in the case of solving the sequential temporal problem. This model has three different layers, the CNN layer, RNN layer and the output layer. Fig \ref{fig:deepsense} shows the architecture of this model.

This model successfully uses the strong feature of both of the general model CNN and RNN and provides an architecture where different design components can be added or removed as required.

\subsubsection{ Semi-Supervised Sequence Learning}

\begin{figure}[h]
\centering\includegraphics[width=0.8\linewidth]{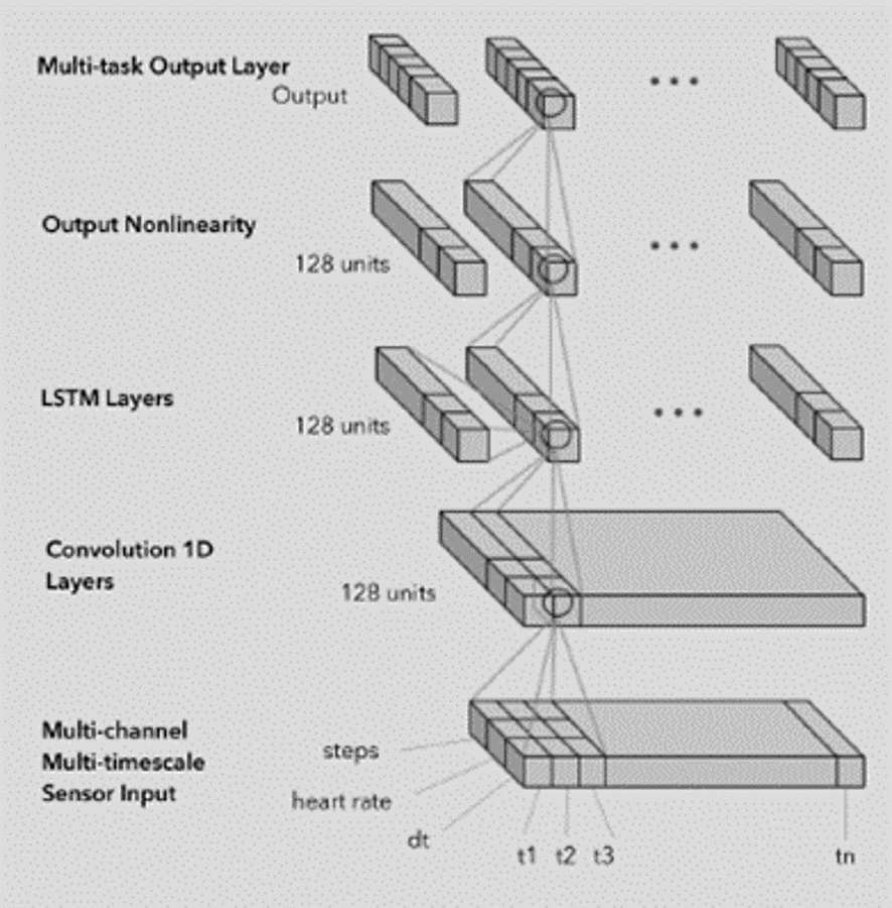}
\caption{Semi-supervised deep learning model for multi-channel input}
\label{fig:semisupervised} 
\end{figure}

B. Ballinger et al. \cite{ballinger2018deepheart} have recently proposed a neural network model architecture to process multi-channel input data through semi-supervised learning. The proposed architecture is a combination of the convolutional 1D neural network \cite{kiranyaz2016real} and a bidirectional LSTM \cite{huang2015bidirectional}. Fig \ref{fig:semisupervised} shows that the proposed model takes multi-channel as well as multi-timescale input from the various data source and pass them through three layers of temporal convolutional 1 D layer, where all the required features are efficiently extracted and model temporal translation invariance \cite{bai2018empirical}.

Once the features are extracted, the model is first trained with a purely-supervised four bi-directional layer of LSTM without pre-training. Then this supervised LSTM is initialised with weight from the heuristic pre-training. The author has demonstrated that for some type of data, the LSTM without pre-training performs significantly better than LSTM with pre-training. On the other hand, the insights of some data, discovered by the heuristic pre-training influence positively the corresponding output. This work is a significant contribution to processing multi-channel input data using semi-supervised learning.

\subsubsection{Graph based Neural Network model}

\cite{zhang2019predicting} proposed a graph based augmentation for continuous-time data with relation between multiple variables of the data. In this graphical augmentation method, a patent record D consists of U clinician and v patients. Therefore each record in a continuous sequence of individual patient's visit (S), each visit record (D) is represented as a graph, $ G = \left\{U,V,\mathcal{E}\right\} $ , where  $ \mathcal{E}=\left\{\left\{w_{i j}\right\}_{i=1}^{M}\right\}_{j=1}^{N} $, represents the set of edges connecting U and V; $w_{ij}$ is the count of patient j visiting clinician i.

\subsubsection{Runge-Kutta Neural Network}
\label{sec:rknet}
The core component of Runge-Kutta Neural Network \cite{wang1998runge} is higher order Runge-Kutta methods.Higher order RK methods are utilized to
build network models with higher accuracy. Generally, RK methods are widely-used procedures to solve ODEs in numerical analysis. RK methods can also be used for image data processing. If a system x can be described as ODE as Eq. \ref{eq:1} with initial condition $x(0)=x_{0}$ and the boundary value for t is [0,T]. This ODE represents the rate of change of the system states.
The rate of change is a function of time and the current system state

\begin{equation}
\label{eq:1}
x(t) = f(x(t))
\end{equation}

\subsection{Physics Informed Neural Network (PINN)}

As defined in \cite{raissi2017physics, raissi2017physics2,raissi2019physics} Physics Informed Neural Networks are model for supervised learning that follow any given law of physics described by general non-linear partial differential equations (PDE). PINN has successfully established the foundation for a new paradigm to model time series using physics informed mathematical model which is fully differentiable using partial differential equations. This model is suitable for both discrete as well as a continuous-time series. PINN uses a one-dimensional non-linear Schrodinger equation \cite{nelson1966derivation} to represent a continuous-time model and uses a large number of data points to enforce physics informed constraints. This large amount of data points can introduce a severe bottleneck in the processing for higher dimensional continuous time series. This work mainly focuses on the data-driven solution and data-driven discovery using partial differential equations. The general form of PINN model is f(t,x) as defined by Eq.~(\ref{eq:4.30})

\begin{equation}\label{eq:4.30}
f \coloneqq u _{t} + N\left[ t \right] 
\end{equation}
Here $u_t$ is the hidden latent and $N[.]$ is a non-linear differential operator. The chain rule of differentiating compositions of functions using automatic differentiation \cite{baydin2014automatic} is used to derive this neural network.  This novel neural network model has successfully demonstrated the success of using automatic differentiation to differentiate the network concerning corresponding input and model parameters and finally obtain a novel physics informed neural network $f$. The authors have improved their work to the next level by defining a deeply hidden physics model as described in Eq.~(\ref{eq:4.31})

\begin{equation}\label{eq:4.31}
f\mathrm{\coloneqq }u_t\mathrm{-}N\mathrm{(}t,x,u_x\mathrm{,\ }u_{xx}\mathrm{,\ \dots \dots \dots )} 
\end{equation}
In Eq.~(\ref{eq:4.31}) $u$ is the derivative of neural network work with respect to time $t$ and space $x$, which has been used the processes described in their earlier work which is simply applying the chain rule of differentiating compositions of functions using automatic differentiation.

This work is indeed a novel addition in the research field of continuous-time modelling using the neural network and automatic differentiation. Although, there are several questions open such as
\begin{itemize}
\item Is it possible to design the architectural components of LSTM neural network in the form of partial differential equations and can obtain an efficient model which is fully differentiable for a continuous-time sequence as input?
\item What are the potential relation between neural network and classical physics?
\item How to resolve the over-fitting and vanishing gradients problems?

\end{itemize}
 
PINNs are useful for an inverse problem where data is unstructured, noisy and multi-variate. PINN models are suitable for continuous data where the state can be at any time with an irregular time interval. However, this neural network does not deal with the variable parametrised model. Therefore, it requires further collaborative work to improve the performance for data-driven dynamics and infer the parametrised partial differential equations. Still, the model is subject to the high complexity associated with neural network for input with a higher dimension. This neural network model only supports fixed model parameters; therefore, in the case of the dynamic model parameters, this network is not efficient enough.

A generative adversarial networks (GAN) implementation \cite{yang2018physics} of PINN, encoded the governing physical laws of the dataset in the form of stochastic differential equations (SDEs) using automatic differentiation into the architecture of GANs. This model uses three generators; two of them are the feed-forward deep neural network, and the third one is induced by SDE to capture the inherent stochasticity and uncertain extrinsic dynamics of the system. This model, PI-GAN, is the best fit for the dynamic system where the underlying physics is partially known, but with the help of some measurement, e.g. initial as well as boundary conditions, it is possible to determine missing pattern and parameters in the PDE as well as the system. To avoid the instability of GAN, this model uses weight-clipped WGANs. Similar to DNN, PI-GAN trains it multiple generators and descriptors iteratively with the loss function for the optimiser. PI-GAN is suitable for forward, inverse and mixed problems but the computational cost of training is much higher than training a single feed-forward neural network or PINN \cite{raissi2017physics}. Both generator and discriminator exhibits over-fitting.
The structure of the model is relatively complex with multiple generators and descriptors with multiple levels of depth and width. 

The main focus of PINN-based model is to find the data-driven solution for PDEs. One of the significant limitations of PINN is that vanishing gradient issue, and it fails to represent the high-frequency sampling rate of the data. PINN also struggles to provide a representation of complex function; therefore, it uses additional constraints of continuity as global regulariser. Distributed PINN (DPINN) \cite{jia2020physics} model provides a solution to these challenges by using local regulariser instead of global regulariser, which helps DPINN to be comparatively more data-efficient. This model divides the computational domain into uniformly distributed non-overlapping cells and install a PINN
in each cell with local regulariser for initial and boundary conditions. each PINN in the corresponding cell compute loss, which is later aggerated to compute the total loss for the gradient descent algorithm of the optimiser. 

One of the hybrid PINN model \cite{qian2019physics} where Convolutional Neural Networks(CNN)
Furthermore, conditional Generative Adversarial Neural Networks(cGANs) are used to extract the hidden dynamics of the data simulated by PDE e.g.Shallow Water Equation (SWE). Due to the hybrid nature, this model provides real-time predictions of the flood with comparatively higher precision. The main challenges of this model are that it can ensure the stability of the dynamics of SWE. Another winning point of this model is that it can overcome the limitation of cGAN where a trivial missing point in the continuous-time series can result in noticeable divergence of the prediction result.   

Table \ref{tab:pinn} shows several application of PINN with real-time data.

\begin{table}
\centering
\caption{Different type of PINN}
\label{tab:pinn}
\begin{tabular}{|p{0.2\textwidth}|p{0.35\textwidth}|p{0.35\textwidth}|} \hline 
\textbf{PINN} & \textbf{Applications} & \textbf{Characteristics} \\ \hline 

\cite{yucesan2019wind}& Wind Turbine Main Bearing Fatigue Life Estimation &  \\ \hline 

PI-GAN \cite{yang2018physics} & Solving stochastic differential equation & GAN with multiple generators and discriminators \\ \hline 

\cite{kissas2020machine} & Non-invasive 4D flow modelling for MRI data & Solve conservation laws in graph topologies\\ \hline

(DPINN) \cite{jia2020physics}&  Solve Advection algorithm & Divide the computational domain and deploy PINN for each each division\\ \hline

fPINN \cite{pang2019fpinns} &  solve space-time fractional advection-diffusion equations (fractional ADEs)& \\ \hline 

\cite{qian2019physics}& Flood Prediction & CNN and cGANs\\ \hline

nPINNs \cite{pang2020npinns} &High-dimensional
parameter spaces. & Flexible and minimal implementation \\ \hline

\end{tabular}
\end{table}

\subsection{Differential Equation Neural Network (DiffNN)}

The main goal of deep learning is to find an unknown function (f) that can correctly predict the value in future time for any given observations based on their past values. Generally, this function f is the concatenation of the hidden layer and non-linear functions. This section reviews the recent trend in neural networks architecture based on both Ordinary Differential equations (ODE) as well as Partial Differential Equations (PDE). Different challenges, as well as strength, are identified for existing limitations in terms of this recent family of Differential Equations based neural network. One of the significant challenges of the neural network model is that they process a continuous time series as a discrete-time sequence with a fixed sampling rate and fixed sampling frequency \cite{neil2016phased}. This characteristic makes real-time continuous data processing difficult. It also imposes high computation load and memory usage. Besides, due to the dependency on the computation of previous states for some neural network, such as RNN models, they are prone to vulnerabilities such as if the time gap, between two consecutive observations, is too big, it can affect the efficiency of the model adversely. Therefore, the neural network performs better for discrete-time series of moderate length with a fixed sampling rate, few missing values, and short time intervals between observations. When it comes to continuous real-time series with multiple variables and irregular sampling rate, neural network model needs different additional techniques.

A differential equation can learn the underlying hidden dynamics of continuous-time data \cite{PEARLMUTTER89A} in real-time.  \cite{chen2018neural, rubanova2019latent} demonstrate the strength of Neural Ordinary Differential Equations (Neural-ODEs). Neural-ODEs mainly re-factored the design of Residual Neural Network (ResNet).

\subsubsection{Residual neural network}

ResNets can be re-designed as a dynamic system []. The architecture of a ResNet can be described using Euler Integration, as shown in Algorithm \ref{alg:1}. First, to take the initial step of the vector and then on every step, compute the update and add that with the hidden vector.  This is similar to Euler integration.

\begin{algorithm}[H]
\caption{ResNet as Euler Integration}
\label{alg:1}
\begin{algorithmic}[1]
\Procedure{resnet}{$z,\theta$}
 \While{$t<T$}
   \State $z = z + f(z, t, \theta)$
  \EndWhile \label{net loop}
  \State $return z$
\EndProcedure
 \end{algorithmic}
\end{algorithm}

Here $f$ is a ResNet as defined in Algorithm \ref{alg:2}. Simply feed the current depth to the ResNet block  and use the fixed set of parameters for the entire block. That refers the hidden dynamics can change between layer continuously by ODEs.  
\begin{algorithm}[H]
\caption{ResNet Function}
\label{alg:2}
\begin{algorithmic}[1]
\Procedure{f}{$z,t,\theta$}
 \State $return resnet(z, \theta[t])$
\EndProcedure
 \end{algorithmic}
\end{algorithm}

The while loop in Algorithm \ref{alg:1} can be replaced by an ODE solver.  Algorithm \ref{alg:1} is same as Algorithm \ref{reverse-ode}. As a result,  \cite{chang2017reversible,lu2018beyond} considers pre-active ResNets as forward Euler method.

\subsubsection{Ordinary Differential Equation as Neural Network }

\label{sec:ode}

Residual Neural Network is discritization of Euler method. In the residual network, the state of the layer $i+1$ is related to the state of the layer $i$ as shown in Eq.~\eqref{eq:ode1};
\begin{equation}
y_{i+1}=y_i+f(y_i,\theta_i)
\label{eq:ode1}
\end{equation}

However, if we consider the Euler method for solves ODE with step size $h=1$ then we obtain residual network. So ResNet is a special case of the Euler method. Briefly, the Euler method is a first-order numerical method for solving an ordinary differential equation with initial condition introduced by Leonhard Euler. Differential equations can model dynamic of continuous systems those are changed by time, since in physics, time is not countable, and it is a real number. Therefore, it is not feasible to use a differential equation for the dynamic of the systems. Nevertheless, for modelling time series for computer, discretising continuous dynamical system is essential. So if $x(t)$ be the state of a system at time $t$,  $\dot{x}(t)$ refers the derivative of $x$ for time, i.e., rate of change of $x$ with respect to time. For example, if $x(t)$ is a population of an entity at time $t$ then $\dot{x}(t)$ is the rate of change in population at time $t$. Differential equations are the relations of the rate of changes with the original state, i.e.,
\[
\dot{x}(t)=f(x(t))
\] 
Solving a differential equation means finding the function $x(t)$ that satisfies the differential equation with the initial condition(the starting point of the computation). So if the state of the system at the starting time is $x_0$, i.e., $x(t_0)=x_0$, where $t_0$ is the starting time of the computation, then, the initial condition for solving an ordinary differential equation(ODE) are as follows,
\[
\begin{aligned}
\label{ode-def}
&\dot{\mathbf{x}}(t)=\mathbf{f}(\mathbf{x}(t))\\
&\mathbf{x}(t_0)=\mathbf{x_0}
\end{aligned}
\]
The solution of \ref{ode-def} is given by the following formula,
\begin{equation}
\label{ode-solution}
\begin{aligned}
&\mathbf{x}(t)=\mathbf{x}(t_0)+\int_{t_0}^t\mathbf{f}(x(s))ds\\
&\forall t\in (a,b)
\end{aligned}
\end{equation}
where $(a,b)$ is the domain of the definition of solution and $t_0\in (a,b)$.
Euler used the first-order approximation of the function $f$ by Taylor's formula, so:
\[
x(t+h)=x(t)+h\dot{x}(t)+o(h^2)
\]
where $o(h^2)$ is the term containing $h^k$ with $k\geq 2$ such that $\lim_{h\to 0}\frac{o(h^2)}{h^2}=0$. If we replace $\dot{x}(t)$ with $f(x(t))$ in this formula and ignore $o(h^2)$ term we have Euler formula for solving ODEs, i.e.,
\[
\begin{aligned}
&x(t+h)=x(t)+hf(x(t))\\
&x(t_0)=x_0
\end{aligned}
\]
If the parameter of the system $\theta$ change with time, i.e., $\theta:\mathbb{R}\to \mathbb{R}$, we have Neural Ordinary Differential Equation with parameter $\theta$ as in Eq.~\eqref{eq:ode2}:
\begin{equation}
\begin{aligned}
x(t+h) &= x(t)+hf(x(t),\theta(t))\\
x(t_0) &= x_0
\end{aligned}
\label{eq:ode2}
\end{equation}

If $h=1$ it is a ResNet. Suppose the loss function is a real valued function of state, i.e., $L:\mathcal{C}\to\mathbb{R}$, where $\mathcal{C}$ is the class of functions $x:\mathbb{R}\to \mathbb{R}^N$ which are the solutions of ODE of the system. The goal is to minimize $L()$ with respect to the controls parameter $\theta(t)$. Lets consider a system of differential equations consisting $n$ differential equations with initial conditions is given as follow:
\begin{equation}
\begin{aligned}
&\dot{x}_i(t)=f_i(x_1(t),\dots,x_n(t),t,\theta(t)),i=1,\dots,n\\ 
&x_1(t_0)=\tilde{x}_0,\dots,x_n(t_0)=\tilde{x}_n
\end{aligned}
\label{ode}
\end{equation}

A specific choice of control function $\theta(t)$ yield a solution of \eqref{ode}, $x_i(t)$ for $t_0\leq t\leq t_1$, where we assume $\theta:\mathbb{R}\to\mathbb{R}$ is a piecewise continuous function. We call this solution a trajectory. The control function $\theta$ is chosen in such a way that at time $t_1$, a set of equations holds as like Eq.~\eqref{eq:terminal}:
\begin{equation}
\Psi_i(x_1(t_1),\dots,x_n(t_1),t_1)=0,\;\; i=1,\dots,p\leq n
\label{eq:terminal}
\end{equation}
This set of equations is called terminal conditions. The problem of Mayer is to find the optimal control function $\theta(t)$ such that minimize the function:
\begin{equation}
\Phi(x_1(t),\dots,x_n(t),t)
\label{objective}
\end{equation}
At the terminal time $t_{1}$ such that \eqref{eq:terminal} holds. This function is called criterion or objective function. Note that \ref{objective} is first calculated at $t_1$ and then minimization applied with respect to $\theta$ such that \eqref{eq:terminal} holds. The set of all point $(x_1,\dots,x_n,t)$ that satisfies in \eqref{eq:terminal} is called terminal manifold. The control function $\theta$ that yield trajectories reach to the terminal manifold is called admissible. The goal is to find an admissible control function $\theta$ such that minimize the objective function which known as  optimal control function ($\Phi$). The corresponding trajectories is called optimal trajectories. Let $(x_1,\dots,x_n,t)$ be a point in terminal manifold such that minimum value of $\Phi$ attained if the admissible trajectory satisfy the initial condition $(x_1,\dots,x_n)$ at time $t$. Let $L$ be a function of $(x_1,\dots,x_n,t)$ with this property, i.e.,
\begin{equation}
L(x_1,\dots,x_n,t)=\text{minumum value of $\Phi$ with the above conditions}
\label{eq:L}
\end{equation}
so the optimal control also depend on $(x_1,\dots,x_n,t)$, i.e., if $\theta^*$ is the optimal control value then,
\[
\theta^*(t)=u(x_1,\dots,x_n,t)
\]
for some real value function $u$, which is called optimal policy function.

If we expand $x_i(t)$ to the first order Taylor expansion in the neighbourhood of $t$ and put this value in $L$ and use optimality of $L$ we have,
\begin{equation}
\begin{aligned}
x_i(t+\Delta t)&=x_i(t)+\dot{x}_i(t)\Delta t+o(\Delta t^2)\\
               &=x_i(t)+f_i(x_1(t),\dots,x_n(t),t,\theta^*(t))\Delta t+o(\Delta t^2)
\end{aligned}
\label{eq:tylor-expansion}
\end{equation}
Now by calculating $L$ on $x_i(t+\Delta t)$ and use \eqref{eq:tylor-expansion} we have,
\begin{equation}
\begin{aligned}
&L(x_1(t+\Delta t),\dots,x_n(t+\Delta t),t+\Delta t)\\
&=L(x_1(t)+f_1(\mathbf{x}(t),t,\theta^*(t))\Delta t+o(\Delta t ^2),\dots,x_n(t)+f_n(\mathbf{x}(t),t,\theta^*(t))\Delta t+o(\Delta t^2),t+\Delta t)\\
&\geq L(x_1(t),\dots, x_n(t),t)
\end{aligned}
\end{equation}
where $\mathbf{x}(t)=(x_1(t),\dots,x_n(t))$. The last inequality holds because $(x_1(t),\dots,x_n(t),t)$ is the optimum trajectory in the interval $[t,t+\Delta t]$. Thus,
\begin{equation}
L(x_1(t),\dots,x_n(t),t)=\min_{\theta}[L(x_1(t)+f_1\Delta t+o(\Delta t ^2),\dots,x_n(t)+f_n\Delta t+o(\Delta t^2),t+\Delta t)]
\label{eq:min-1}
\end{equation}
Now if we apply first-order Taylor formula to $L$ we obtain,
\begin{equation}
\begin{aligned}
&L(x_1(t)+f_1\Delta t+o(\Delta t ^2),\dots,x_n(t)+f_n\Delta t+o(\Delta t^2),t+\Delta t)\\
&=L(x_1(t),\dots, x_n(t),t)+\sum_{i=1}^n\frac{\partial L}{\partial x_i}f_i\Delta t+\frac{\partial L}{\partial t}\Delta t+o(\Delta t^2)
\end{aligned}
\end{equation}
if we remove $L(x_1(t),\dots,x_n(t),t)$ from both hand side of 
\eqref{eq:min-1} we have,
\begin{equation}
0=\min_{\theta}[\sum_{i=1}^n\frac{\partial L}{\partial x_i}f_i\Delta t+\frac{\partial L}{\partial t}\Delta t+o(\Delta t^2)]
\label{min-2}
\end{equation}
by deviding both hand side of \ref{min-2} and letting $\Delta t\to 0$ we have,\cite{dreyfus1965dynamic},
\begin{equation}
0=\min_u\left[\sum_{i=1}^nL_{x_i}f_i+L_t\right]
\end{equation}
where $L_{x_i}=\frac{\partial L}{\partial x_i}$ and $L_t=\frac{\partial L}{\partial t}$. If $u^*=\mathrm{argmin}_u\left[\sum_{i=1}^nL_{x_i}f_i+L_t\right]$ then,
\begin{equation}
\sum_{i=1}^nL_{x_i}f_i(x_1,\dots,x_n,t,u^*)+L_t=0
\label{eq:argmin}
\end{equation}
since $L$ is not explicitly depend on $u$, then $\frac{\partial L}{\partial u}=0$. 
Our goal is to obtain an ODE to describe the dynamic of $L_{x_j}$ with respect to time. The first step is to compute $(\frac{dL_{x_j}}{dt})_u$ which subscribe $u$ means that $L$ is computed with respect to the control $u$. By the chain rule and the fact that $L$ is a function of $(x_1,\dots,x_n,t)$ and each $x_i$ is a function of $t$ we have,
\begin{equation}
\begin{aligned}
(\frac{dL_{x_j}}{dt})_u&=\sum_{i=1}^n\frac{\partial L_{x_j}}{\partial x_i}\frac{d x_i}{d t}+\frac{\partial L_{x_j}}{\partial t}\\
&=\sum_{i=1}^n\frac{\partial L_{x_j}}{\partial x_i}f_i+\frac{\partial L_{x_j}}{\partial t}
\end{aligned}
\label{eq:maineq1}
\end{equation}
The partial derivative in Eq.~\eqref{eq:argmin} with respect to $x_j$ can be described as Eq.~\eqref{eq:maineq2},
\begin{equation}
\begin{aligned}
\sum_{i=1}^n(\frac{\partial L_{x_i}}{\partial x_j}f_i+L_{x_i}\frac{\partial f_i}{\partial x_j})+\frac{\partial L_t}{\partial x_j}&=\sum_{i=1}^n\frac{\partial L_{x_i}}{\partial x_j}f_i+ \sum_{i=1}^nL_{x_i}\frac{\partial f_i}{\partial x_j}+\frac{\partial L_t}{\partial x_j}\\
&=\sum_{i=1}^n\frac{\partial L_{x_i}}{\partial x_j}f_i+\frac{\partial L_t}{\partial x_j}+ \sum_{i=1}^nL_{x_i}(\frac{\partial f_i}{\partial x_j}+\frac{\partial f_i}{\partial u}\frac{\partial u}{\partial x_j})\\
&=\sum_{i=1}^n\frac{\partial L_{x_i}}{\partial x_j}f_i+\frac{\partial L_t}{\partial x_j}+ \sum_{i=1}^nL_{x_i}\frac{\partial f_i}{\partial x_j}\\
&=0
\end{aligned}
\label{eq:maineq2}
\end{equation}
the last equality holds because of \ref{optimality}. If we combine Eq.~\eqref{eq:maineq1} and Eq.~\eqref{eq:maineq2} we derive the following differential equation,\footnote{If $L$ and its partial derivative are continuous then $\frac{\partial^2L}{\partial x_i\partial t}=\frac{\partial^2L}{\partial t\partial x_i}$ so $\frac{\partial L_{x_j}}{\partial t}=\frac{\partial L_t}{\partial x_j}$
}\begin{equation}
(\frac{dL_{x_j}}{dt})_u=-\sum_{i=1}^nL_{x_i}\frac{\partial f_i}{\partial x_j}
\label{eq:adjoint}
\end{equation}
Let $a_i(t)=\frac{\partial L}{\partial x_i}$ then Eq.~\ref{eq:adjoint} is the adjoin equation for Eq.~\eqref{eq:daj},
\begin{equation}
(\frac{da_j}{dt})_u=-\sum_{i=1}^na_i(t)\frac{\partial f_i}{\partial x_j}
\label{eq:daj}
\end{equation}
In the matrix notation, if $\mathbf{a}(t)=[a_1(t),\dots,a_n(t)]^\top$ and $\mathbf{f}(\mathbf{x}(t),t,\theta)=[f_1,\dots,f_n]^\top$ then Eq.~\eqref{eq:daj} can also be described as Eq.~\eqref{eq:daj2},
\begin{equation}
(\frac{da_j(t)}{dt})_u=-\mathbf{a}(t)^\top\frac{\partial \mathbf{f}(\mathbf{x}(t),t,\theta)}{\partial x_j}
\label{eq:daj2}
\end{equation}
If we know the value of $a_j(t)$ at time $t_1$ then we could compute the value of $a_j(t)$ at time $t_0$, where $t_0\leq t_1$, by,
\begin{equation}
a_j(t_0)=a_j(t_1)-\int_{t_1}^{t_0}\mathbf{a}(t)^\top\frac{\partial \mathbf{f}(\mathbf{x}(t),t,\theta)}{\partial x_j}dt
\label{eq:daj3}
\end{equation}
but to solve this integral, we need to know the value of $\mathbf{x}(t)$ at the entier time interval $[t_0,t_1]$ but since we know the initial condition of $\mathbf{x}(t_0)$ we could obtain the trajectory 
of $\mathbf{x}(t)$ for all $t\in[t_0,t_1]$. 

Now, we can obtain the sensitivity of $L$ with respect to the control parameter $u$. 
If we compute partial derivative of \eqref{eq:argmin} with respect to $u$, the control parameter, then we have,
\begin{equation}
\label{optimality}
\sum_{i=1}^n L_{x_i}(\frac{\partial f_i}{\partial u})_{u^*} +\frac{\partial L_t}{\partial u}= 0 
\end{equation}
or equivalently,
\begin{equation}
\sum_{i=1}^n a_i(t)(\frac{\partial f_i}{\partial u})_{u^*}+\frac{\partial L_t}{\partial u} = 0 
\end{equation}
since $a(t)=L_{x_i}(x_1(t),\dots,x_n(t),t,u)$, but we know $\frac{\partial L_t}{\partial u}=\frac{d L_u}{dt}$ so,
\begin{equation}
\frac{\partial L_u}{dt} = - \sum_{i=1}^n a_i(t)(\frac{\partial f_i}{\partial u})_{u^*}
\end{equation}
the solution of this equation at time $t_0$ is obtaine by taking integral from $t_1$ to $t_0$ backward in time, because we know the value of $L_u$ at time $t_1$, i.e.,
\begin{equation}
\frac{\partial L}{\partial u} = (\frac{\partial L}{\partial u})|_{t=t_1} -\int_{t_1}^{t_0} \sum_{i=1}^n a_i(t)(\frac{\partial f_i}{\partial u})_{u^*}
\end{equation}
but $(\frac{\partial L}{\partial u})|_{t=t_1}=0$ and in matrix notation we have,
\begin{equation}
\frac{\partial L}{\partial u} = -\int_{t_1}^{t_0} \mathbf{a(t)}^{\top}\frac{\partial \mathbf{f}(\mathbf{x}(t),t,u(t))}{\partial u}dt
\end{equation}

The value of $\frac{\partial \mathbf{f}(\mathbf{x}(t),t,u(t))}{\partial u}$ and the Jacobian $\frac{\partial \mathbf{f}(\mathbf{x}(t),t,u(t))}{\partial \mathbf{x}}$ can be computed by automatic differentiation \cite{PEARLMUTTER_LTUB}. If we consider the ODE solver as the black box then the initial value problem,
\begin{equation}
\begin{aligned}
&\dot{\mathbf{x}}(t)=\mathbf{f}(\mathbf{x}(t),t,u(t))\\
&\mathbf{x}(t_0)=x_0\\
&t\in[t_0,t_1]
\end{aligned}
\end{equation}
then this equation can be solve by calling an ODE solver $\mathtt{ODESOLVER}(\mathbf{x}(t_0),\mathbf{f},t_0,t_1,u)$ which get initial condition $\mathbf{x}(t_0)$, dynamic of the problem $\mathbf{f}$, start time $t_0$, end time of $t_1$ and control parameter $u$ as input and produce $\mathbf{x}(t_1)$ as the output. So the reverse mode derivative of an ODE can be computed by Algorithm \ref{reverse-ode}.

\begin{algorithm}
\caption{Reverse-mode derivative for ODE}
\label{reverse-ode}
\begin{algorithmic}[1]
\Procedure{Reverse-ODE}{$u,t_0,t_1,\mathbf{x}(t_1),\mathbf{a}(t_1)$}
    \State $\mathbf{a}(t_0)=\mathtt{ODESOLVER}(\mathbf{a}(t_1),-\mathbf{a}(t)\frac{\partial \mathbf{f}}{\partial\mathbf{x}},t_1,t_0,u)$
    \State  $\frac{\partial L}{\partial u}|_{t_0}=\mathtt{ODESOLVER}(0,-\mathbf{a}(t)\frac{\partial f}{\partial u},t_1,t_0,u)$
    \State Return $(\mathbf{a}(t_0),\frac{\partial L}{\partial u}|_{t_0})$
\EndProcedure
\end{algorithmic}
\end{algorithm}

\begin{figure}[h]
\centering\includegraphics[width=0.9\linewidth]{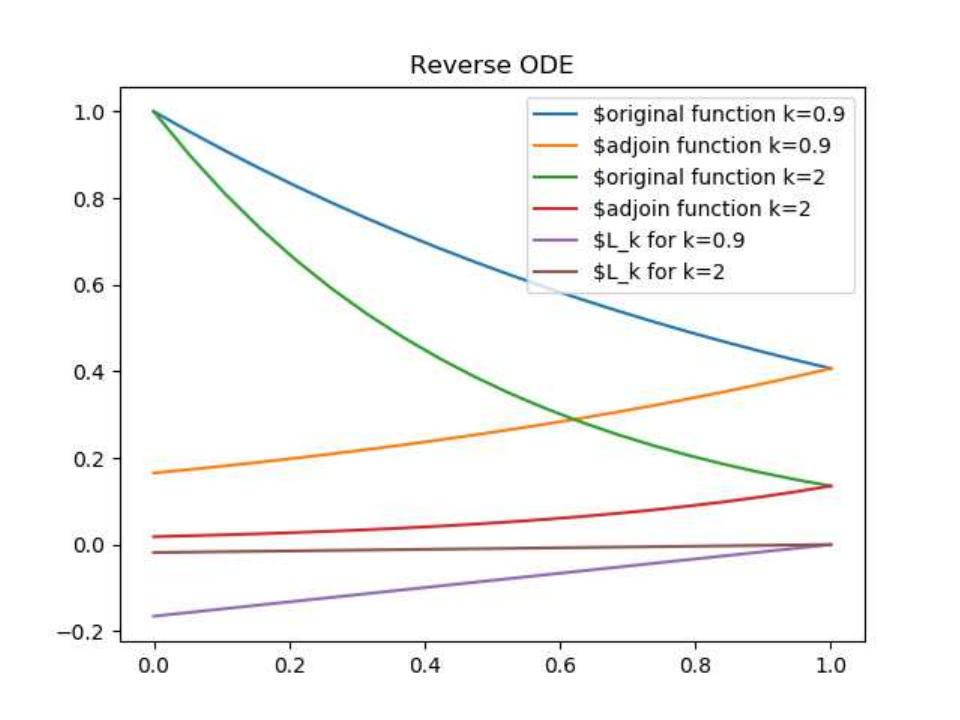}
\caption{Reverse ODE withODEsolver from scipy of Python} 
\label{fig:adjoin}
\end{figure}

As an example, consider the following ODE as the dynamic of the problem,
\begin{equation}
\begin{aligned}\label{original-ex}
&\frac{dx(t)}{dt}=-kx\\
&x(0)=1
\end{aligned}
\end{equation}
where the control parameter $k$ is constant. By calling the $\mathtt{odeint}$ from $\mathtt{scipy}$ we could obtain $x(1)$\footnote{In fact by calling theODEsolver we obtain $x(t)$ in all $t\in[0,1]$}. Then the dynamic of the lose function is as follow,
\begin{equation}
\begin{aligned}\label{adjoin-ex}
&\frac{da(t)}{dt}=ka(t)\\
&a(1)=x(1)
\end{aligned}
\end{equation}
i.e., we start from $x(1)$ backward in time until $t=0$. Note that by definition $a(t_i)=\frac{\partial L}{\partial x(t_i)}$. We use $\mathtt{grad}$ from $\mathtt{autograde}$(Python) to obtain $\frac{\partial f(x,t,k)}{\partial x}$ where $f(x,t,k)=-kx(t)$. By another call to theODEsolver we obtain $a(t)$ in $[1,0]$\footnote{since $a(t)$ is calculated backward we write the time interval in the opposit order}.The optimal control policy is a value of $k$ such that $L_k(t)$ is minimum at $t=0$.
But the dynamic of $L_k$ is as follow,
\begin{equation}
\begin{aligned}
&\frac{dL_k(t)}{dt}=a(t)x(t)\\
&L_k(1)=0
\end{aligned}
\end{equation}
where $a(t)$ and $x(t)$ are the solutions of \ref{original-ex} and \ref{adjoin-ex}, respectively. In the Figure \ref{fig:adjoin} we plot the graph of $x(t), a(t)$ and $L_k(t)$ for $k=0.9$ and $k=2$ in the intgerval $[0,1]$.\footnote{Note that the whole process is done automatically by obtaining dynamic of adjoin and dynamic of control function by automatic differentiation and solving with the ODE solver} 

\subsubsection{Continuous Normalising Flows and Neural ODE}

Now suppose we have a evolution of a random variable. Let $\mathbf{x}_0\in\mathbb{R}^N$ be a $N$-dimentional random variable and suppose it continously transform to another random variable $\mathbf{x}_T\in\mathbb{R}^N$ at time $T$ via the following dynamic,
\begin{equation}
\begin{aligned}
&\frac{d\mathbf{x}}{dt}=\mathbf{f}(\mathbf{x}(t),t)\\
&\mathbf{x}(t_0)=\mathbf{x}_0\\
&\mathbf{x}_0\sim \mathbf{p}_{\mathbf{x}_0}
\end{aligned}
\end{equation} 
where in the last line we mean the probability density function of the random variable $\mathbf{x}_0$ is $\mathbf{p}_{\mathbf{x}_0}$. The probability density function of a continous random variable $\mathbf{x}\in\mathbb{R}^N$ is a real valued function $p:\mathbb{R}^n\to\mathbb{R}$ such that,\cite{papoulis2002probability} ,
\begin{enumerate}
\item $p(\mathbf{z})\geq 0\text{ for all }\mathbf{z}\in\mathbf{R}^N$
\item $\int_{\mathbb{R}^N}p(\mathbf{z})d\mathbf{z}=1$
\end{enumerate}
To obtain the dynamic for $\mathbf{p}_{\mathbf{x}(t)}$ we need to undrestand the nature of transformation in one step. Suppose $\mathbf{x}\in\mathbb{R}^N$ is a random variable with probability density function $\mathbf{p}_{\mathbf{x}}$. Suppose $\mathbf{x}$ change continously to random variable $\mathbf{y}\in \mathbb{R}^N$ via the smooth function $\mathbf{f}:\mathbb{R}^N\to \mathbb{R}^N$, i.e., $\mathbf{y}=\mathbf{f}(\mathbf{x})$. For simplicity suppose $\mathbf{J}(\mathbf{f}(\mathbf{x}))$ is invertible for every value of $\mathbf{x}\in\mathbb{R}^N$. To obtain probability density function of $\mathbf{y}$ we use change of variable formula \cite{spivak2018calculus} we have,
\begin{equation}
\begin{aligned}
\int_{\mathbb{R}^N}\mathbf{p}_{\mathbf{x}}(x)dx&=\int_{\mathbb{R}^N}\mathbf{p}_{\mathbf{x}}(x)|\det(\frac{\partial x}{\partial y})|dy\\
&=\int_{\mathbb{R}^N}\mathbf{p}_{\mathbf{x}}(x)|\det(\mathbf{J}(\mathbf{f}(x)))^{-1}|dy\\
&=\int_{\mathbb{R}^N}\mathbf{p}_{\mathbf{x}}(x)|\det(\mathbf{J}(\mathbf{f}(x)))|^{-1}dy\\
&=1
\end{aligned}
\end{equation}

where $\mathbf{J}$ is the Jacobian\footnote{If $x=g(y)$ then $\frac{\partial x}{\partial y}$ is another notation for the Jacobian $\mathbf{J}g(y)$} operator\footnote{By the Inverse Function Theorem if $\mathbf{J}\mathbf{f}(\mathbf{x}_0)$ is invertible, then the function $f$ has an inverse in the neighborhood of $\mathbf{x}_0$ and $\mathbf{J}(\mathbf{f}^{-1}(\mathbf{y}_0))=(\mathbf{J}(\mathbf{f}(\mathbf{x}_0)))^{-1}$ where $\mathbf{y}_0=\mathbf{f}(\mathbf{x}_0)$,\cite{spivak2018calculus}}. Since $\mathbf{p}_{\mathbf{x}}(x)|\det(\mathbf{J}(\mathbf{f}(x)))|^{-1}\geq 0$ and its integral over $\mathbb{R}^N$ is equal to $1$ it is probability density function of $\mathbf{y}$, i.e., $\mathbf{p}_{\mathbf{y}}(y)=\mathbf{p}_{\mathbf{x}}(x)|\det(\mathbf{J}(\mathbf{f}(x)))|^{-1}$, where $y=\mathbf{f}(x)$.

 Now if we divide both side of this equation by $\mathbf{p}_{\mathbf{x}}$ and apply $\log$ function in both side we have,
\begin{equation}
\begin{aligned}
\log(\frac{\mathbf{p}_{\mathbf{y}}(y)}{\mathbf{p}_{\mathbf{x}}(x)})&=\log|\det(\mathbf{J}(\mathbf{f}(x)))|^{-1}\\
&=-\log|\det(\mathbf{J}(\mathbf{f}(x)))|
\end{aligned}
\end{equation}
hence,
\begin{equation}\label{log-dist}
\log(\mathbf{p}_{\mathbf{y}}(y))=\log(\mathbf{p}_{\mathbf{x}}(x))-\log|\det(\mathbf{J}(\mathbf{f}(x)))|
\end{equation}
Now suppose $\mathbf{x}_0\sim \mathbf{p}_{\mathbf{x}_0}$ and,
\begin{equation}
\begin{aligned}
\mathbf{x}_i=\mathbf{f}_i(\mathbf{x}_{i-1}), i=1,\dots,M
\end{aligned}
\end{equation}
,i.e., $\mathbf{x}_M=\mathbf{f}_M\circ\dots\circ \mathbf{f}_1(\mathbf{x}_0)$. We can compute the probability destribution of $\mathbf{p}_{\mathbf{x}_M}$ by applying equation \ref{log-dist} $M$ times.
So if we suppose $\mathbf{x}_i\sim \mathbf{p}_{\mathbf{x}_i}$ we have,
\begin{equation}
\begin{aligned}\label{M-step}
\log(\mathbf{p}_{\mathbf{x}_M}(x_M))&=\log(\mathbf{p}_{\mathbf{x}_{M-1}}(x_{M-1})-\log|\det(\mathbf{J}(\mathbf{f}_{M-1})(x_{M-1}))|\\
&=\log(\mathbf{p}_{\mathbf{x}_{M-2}}(x_{M-2})-\log|\det(\mathbf{J}(\mathbf{f}_{M-1}(x_{M-2})))|-\log|\det(\mathbf{J}(\mathbf{f}_M(x_{M-1})))|\\
\vdots&\\
&=\log(\mathbf{p}_{\mathbf{x}_0}(x_0))-\log|\det(\mathbf{J}(\mathbf{f}_1(x_0)))|-\dots-\log|\det(\mathbf{J}(\mathbf{f}_M(x_{M-1})))|\\
&=\log(\mathbf{p}_{\mathbf{x}_0}(x_0))-\sum_{i=1}^M\log|\det(\mathbf{J}(\mathbf{f}_i(x_{i-1})))|
\end{aligned}
\end{equation}

Now suppose $\mathbf{x}(t)$ change continuously by the following ODE,
\begin{equation}
\begin{aligned}
&\dot{\mathbf{x}}(t)=\mathbf{f}(\mathbf{x}(t),t)\\
&\mathbf{x}(t_0)=\mathbf{x}_0\sim \mathbf{p}_{\mathbf{x}_0}
\end{aligned}
\end{equation}

Now let $\mathbf{x}(t)$ transform continiously to $\mathbf{x}(t+\varepsilon)$ for small positive value of $\varepsilon$ via smooth function $\mathbf{F}$, i.e., $\mathbf{x}(t+\varepsilon)=\mathbf{F}(\mathbf{x}(t))$. If we use first order Taylor expansion of $\mathbf{x}(t+\varepsilon)$ we have,
\begin{equation}
\begin{aligned}
\mathbf{x}(t+\varepsilon)&=\mathbf{x}(t)+\varepsilon\dot{\mathbf{x}}(t)+o(\varepsilon^2)\\
&=\mathbf{x}(t)+\varepsilon\mathbf{f}(\mathbf{x}(t))+o(\varepsilon^2)\\
\end{aligned}
\end{equation}

So $\mathbf{F}(\mathbf{x}(t))=\mathbf{x}(t)+\varepsilon\mathbf{f}(\mathbf{x}(t))+o(\varepsilon^2)$. On the other hand by equation \ref{log-dist} we have,
\begin{equation}
\log(\mathbf{p}_{\mathbf{x}(t+\varepsilon)}(x(t+\varepsilon))-\log(\mathbf{p}_{\mathbf{x}(t)}(x(t))=-\log|\det(\mathbf{J}(\mathbf{F}(x(t))))|
\end{equation}
If we divide both hand side of this equation by $\varepsilon$ and take limit while $\varepsilon\to 0$ we have,
\begin{equation}
\begin{aligned}
\lim_{\varepsilon\to 0}\frac{\log(\mathbf{p}_{\mathbf{x}(t+\varepsilon)}(x(t+\varepsilon))-\log(\mathbf{p}_{\mathbf{x}(t)}(x(t)))}{\varepsilon}=-\lim_{\varepsilon\to 0}\frac{\log|\det(\mathbf{J}(\mathbf{F}(x(t))))|}{\varepsilon}
\end{aligned}
\end{equation}
The left hand side is the definition of $\frac{d\mathbf{p}_{\mathbf{x}(t)}(x(t))}{dt}$. We need to simplify the right hand side of the equation by unfolding $\mathbf{J}(\mathbf{F}(\mathbf{x}(t)))$,
\begin{equation}
\begin{aligned}
\mathbf{J}(\mathbf{F}(x(t)))&=\frac{\partial \mathbf{F}(x(t))}{\partial x(t)}\\
&=\frac{\partial x(t)}{\partial x(t)}+\varepsilon\frac{\partial \mathbf{f}(x(t),t)}{\partial x(t)}+\frac{\partial}{\partial x(t)}o(\varepsilon^2)\\
&=I+\varepsilon\frac{\partial \mathbf{f}(x(t),t)}{\partial x(t)}+\frac{\partial}{\partial x(t)}o(\varepsilon^2)\\
\end{aligned}
\end{equation}
so in limit we have,
\begin{equation}
\begin{aligned}
\lim_{\varepsilon\to 0}\frac{\log|\det(\mathbf{J}(\mathbf{F}(x(t))))|}{\varepsilon}&=\lim_{\varepsilon\to 0}\frac{\log|\det(I+\varepsilon\frac{\partial \mathbf{f}(x(t),t)}{\partial x(t)}|)}{\varepsilon}
\end{aligned}
\end{equation}
Before proceed, we need some results from linear algebra. If $A$ is a $N\times N$ positive definite matrix then all of its eigenvalues are positive real numbers \cite{halmos2017finite} and its characteristic polynomial is 
\begin{equation}
\begin{aligned}
\chi_A(\varepsilon)&=\det(\varepsilon I-A)\\
&=(\varepsilon_1-\varepsilon)\dots(\varepsilon_N-\varepsilon)
\end{aligned}
\end{equation}
where $\varepsilon_1,\dots,\varepsilon_N$ are eigenvalues of $A$. By applying the $\log$ function\footnote{here we mean $\log_e$ which is $\ln$ function} on both hand side of the charactristic polynomial and taking derivative with respect to $\varepsilon$ we have,
\begin{equation}
\begin{aligned}
\frac{\chi_A'(\varepsilon)}{\chi_A(\varepsilon)}=\frac{-1}{\varepsilon_1-\varepsilon}+\dots+\frac{-1}{\varepsilon_N-\varepsilon}
\end{aligned}
\end{equation}
now if $\varepsilon\to 0$ we have,
\begin{equation}
\begin{aligned}
\lim_{\varepsilon\to 0}\frac{\chi_A'(\varepsilon)}{\chi_A(\varepsilon)}&=\frac{-1}{\varepsilon_1}+\dots+\frac{-1}{\varepsilon_N}\\
&=-\mathtt{trace}(A^{-1})
\end{aligned}
\end{equation}
we have $\det(I+\varepsilon A)=\det(A)\cdot\chi_{-A^{-1}}(\varepsilon)$ since,
\begin{equation}
\begin{aligned}
\det(I+\varepsilon A)&=\det(A(A^{-1}+\varepsilon I))\\
&=\det(A)\cdot\det(\varepsilon I-(-A^{-1}))\\
&=\det(A)\cdot\chi_{-A^{-1}}(\varepsilon)
\end{aligned}
\end{equation}
so if $A$ is a positive definite matrix we have
\begin{equation}
\begin{aligned}
\lim_{\varepsilon\to 0}\frac{\log|\det(I+\varepsilon A)|}{\varepsilon}&=\lim_{\varepsilon\to 0}\frac{\log|\det(A)\cdot\chi_{-A^{-1}}(\varepsilon)|}{\varepsilon}\\
&=\lim_{\varepsilon\to 0}\frac{\chi_{-A^{-1}}'(\varepsilon)}{\chi_{-A^{-1}}(\varepsilon)}\\
&=\mathtt{trace}(A)
\end{aligned}
\end{equation}
Here we used L'Hopital's rule since the first limit is in the indeterminate forms $\frac{0}{0}$.
Hence if $\frac{\partial\mathbf{f}(x(t),t)}{\partial x(t)}$ is positive definit at $x(t)$ then,
\begin{equation}
\begin{aligned}
\lim_{\varepsilon\to 0}\frac{\log|\det(I+\varepsilon \frac{\partial\mathbf{f}(x(t),t)}{\partial x(t)})|}{\varepsilon}=\mathtt{trace}(\frac{\partial\mathbf{f}(x(t),t)}{\partial x(t)})
\end{aligned}
\end{equation}
and finaly,
\begin{equation}
\frac{d\log(\mathbf{p}_{\mathbf{x}(t)}(x(t))}{dt}=-\mathtt{trace}(\frac{\partial \mathbf{f}(x(t),t)}{\partial x(t)})
\end{equation}
by solving this ODE with respect to $\mathbf{p}_{\mathbf{x}(t)}(x(t))$ we obtain probability density function at each instant of time and we could compute density at the final time ,i.e., $\mathbf{p}_{\mathbf{x}(T)}$,\footnote{If $\mathbf{f}$ is Lipschits continious with respect to $\mathbf{x}$ then by Rademacher's theorem $\mathbf{f}$ is differentiable almost everywhere(the points where $\mathbf{f}$ is not differentiable is of measure zero). So we need a stronger condition for $\mathbf{f}$ to be sure that $\frac{\partial \mathbf{f}}{\partial \mathbf{x}}$ exists.}.

\begin{figure}[h]
\centering\includegraphics[width=0.9\linewidth]{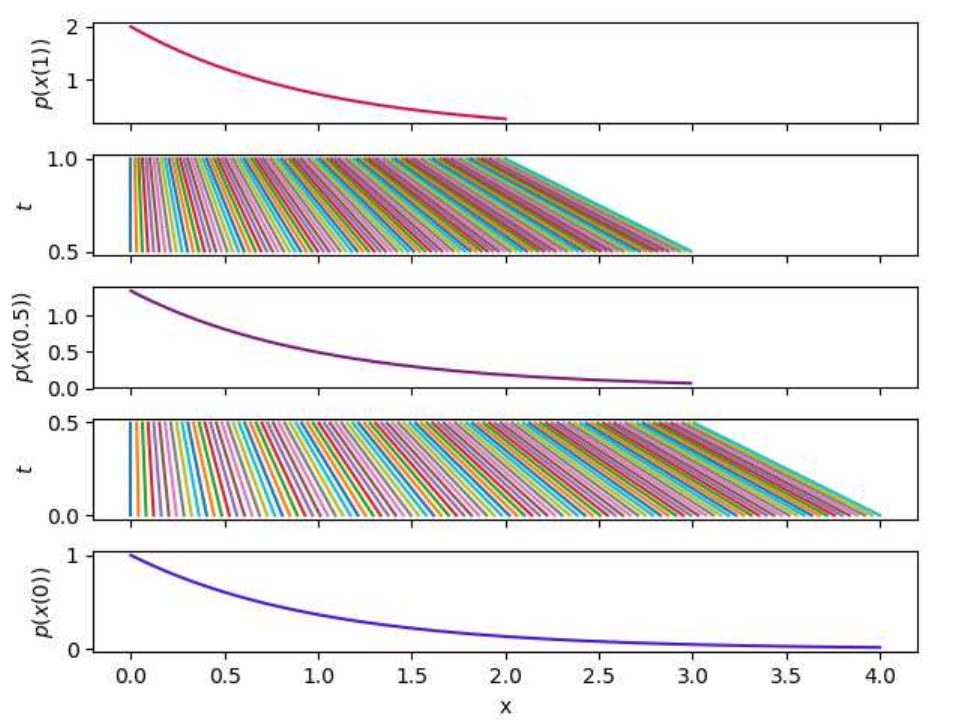}
\caption{Evolution of $p_{x(0)}(x)=e^{-x}$ by $\frac{dx}{dt}=\frac{x}{t-2}$} 
\label{fig:linear_flow}
\end{figure}

\begin{figure}[h]
\centering\includegraphics[width=0.9\linewidth]{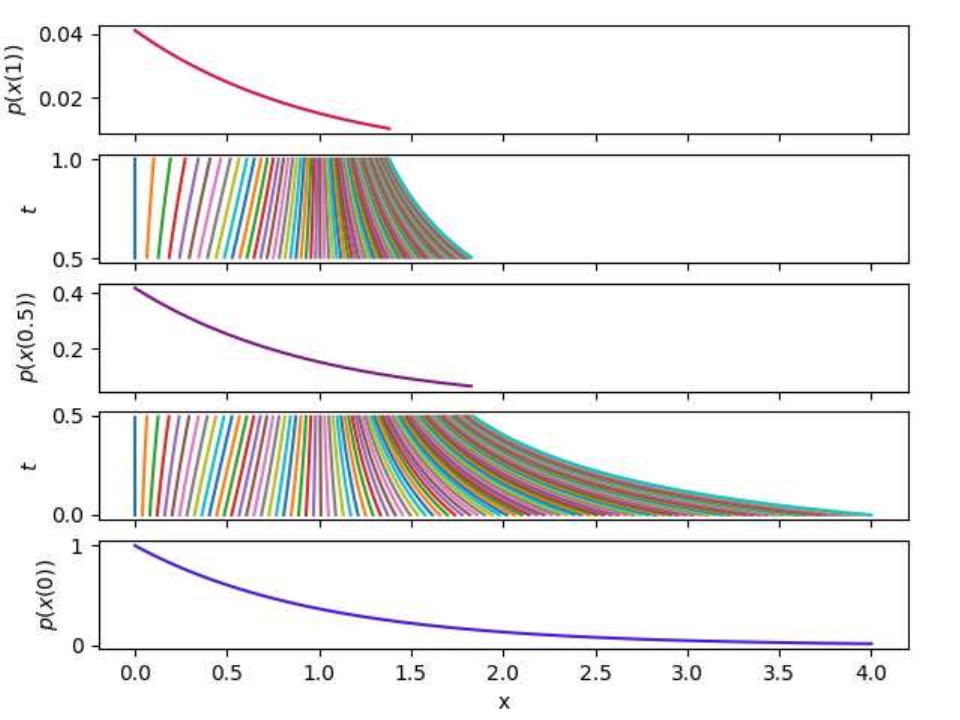}
\caption{Evolution of $p_{x(0)}=e^{-x}$ by $\frac{dx}{dt}=x(1-x)$} 
\label{fig:chaotic_flow_exp}
\end{figure}

\begin{figure}[h]
\centering\includegraphics[width=0.9\linewidth]{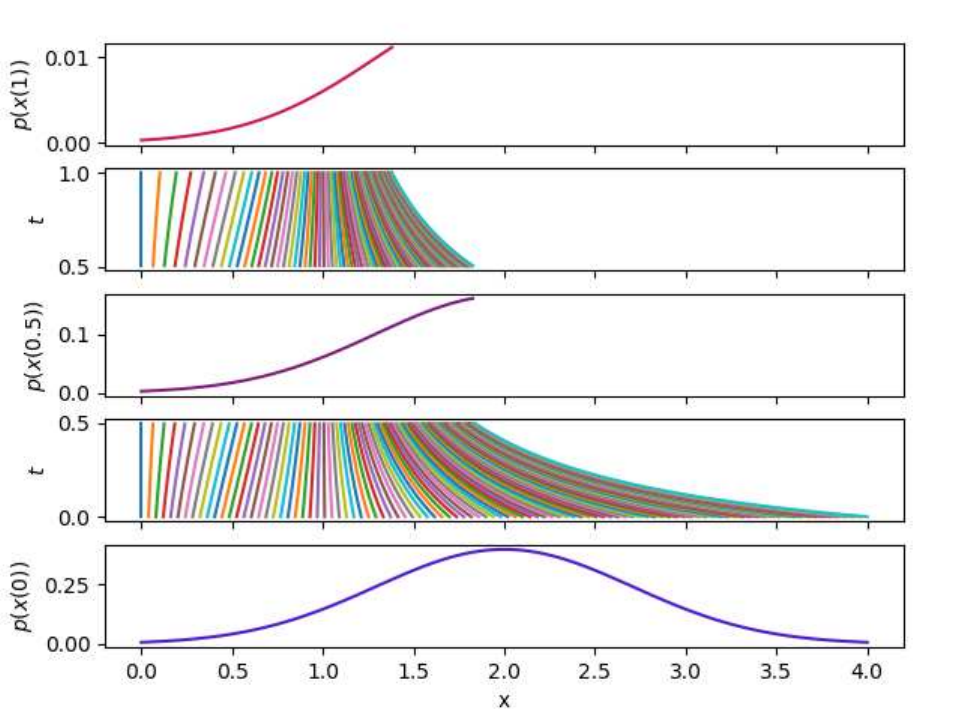}
\caption{Evolution of $x(0)\sim \mathcal{N}(2,1)$ by $\frac{dx}{dt}=x(1-x)$}
\label{fig:chaotic_flow_normal}
\end{figure}

In Figure \ref{fig:linear_flow} we show the evolution of the exponential distribution $p_{x(0)}(x)=e^{-x}$ for $x\geq 0$ under the dynamic $\frac{dx}{dt}=\frac{x(t)}{t-2}$. In Figure \ref{fig:chaotic_flow_exp} we have shown the evolution of the exponential distribution under the dynamic $\frac{dx}{dt}=x(t)(1-x(t))$ and finaly in Figure \ref{fig:chaotic_flow_normal} we show the evolution of the probability density function of a normal distributed random variable $x(0)\sim \mathcal{N}(2,1)$, i.e., $p_{x(0)}(x)=\frac{1}{\sqrt{2\pi}}e^{-\frac{(x-2)^2}{2}}$, under the dynamic $\frac{dx}{dt}=x(t)(1-x(t))$.

\paragraph{Time complexity comparsion}
For computing $p_{\mathbf{x}_M}(x_M)$ using equation \eqref{M-step} we need to compute $M$ determinant of $N\times N$ Jacobian matrices $\mathbf{J}\mathbf{f}_i(x_{i-1})$, $i=1,\dots M$. We could compute the partial derivative of the function $g:\mathbb{R}^N\to \mathbb{R}$ by automatic differentiation with cost $O(\mathbf{size}(\frac{\partial g}{\partial x_i}))=O(\mathbf{size}(g))$ where $\mathbf{size}(g)$ is the size of computational graph of $g$. But by Strassen algorithm we have an $O(N^3\cdot\mathbf{size}(\mathbf{f}_i))$ algorithm to compute $\mathbf{J}\mathbf{f}_i(x_{i-1})$. So the time complexity for computing $p_{\mathbf{x}_M}(x_M)$ is,
\begin{equation}
O(N^3\cdot(\mathbf{size}(\mathbf{f}_1)+\dots+\mathbf{size}(\mathbf{f}_M))
\end{equation}

On the other hand, for computing continious normalizing flow we need to compute $\mathtt{trace}(\frac{\partial \mathbf{f}(x(t),t)}{\partial x(t)})$ and a call to an ODE solver. We know that,
\begin{equation}
\mathtt{trace}(\frac{\partial \mathbf{f}(x(t),t)}{\partial x(t)})=\frac{\partial \mathbf{f}_1(x(t),t)}{\partial x_1(t)}+\dots+\frac{\partial \mathbf{f}_N(x(t),t)}{\partial x_N(t)}
\end{equation} 
then the time complexity for computing $\mathtt{trace}(\frac{\partial \mathbf{f}(x(t),t)}{\partial x(t)})$ is $O(N\cdot \mathbf{size}(f))$ which is less than standard normalizing flow with time complexity $O(N^3)$.

Table \ref{tab:odenn} shows several Neural-ODE and their characteristics. 

\begin{table}
\centering
\caption{Different type of Neural-ODE}
\label{tab:odenn}
\begin{tabular}{|p{0.2\textwidth}|p{0.75\textwidth}|} \hline 
\textbf{Neural-ODE}  & \textbf{Characteristics} \\ 
\toprule 

ODE-RNN \cite{chen2018neural} & \\ \hline

Latent-ODE \cite{rubanova2019latent} &  \\ \hline

ODE-GRU and ODE-LSTM \cite{2020arXiv200509807H} & Continuous-time version of the GRU and LSTM. \\ \hline

GRU-ODE-Bayes \cite{de2019gru} & A continuous-time version of the GRU, built on Neural-ODE, and a Bayesian update network that processes the sporadic observations of continuous-time series.\\ \hline

Continuous ODE-GRU-D  and Extended ODE-GRU-D \cite{2020arXiv200510693H} & \\ \hline

Augmented Neural ODEs (ANODEs)\cite{dupont2019augmented}& \\ \hline

Ordinary Differential Equation Variational Auto-Encoder ($ODE^{2}VAE$) \cite{yildiz2019ode}& A latent second order ODE model for high-dimensional sequential data\\ \hline

\bottomrule
\end{tabular}
\end{table}

Different applications are explored to understand the efficiency of Neura-ODE. Table \ref{tab:odenn-app}

\begin{table}
\centering
\caption{Different applications of Neural-ODE}
\label{tab:odenn-app}
\begin{tabular}{|p{0.2\textwidth}|p{0.35\textwidth}|p{0.35\textwidth}|} \hline 
\textbf{Neural-ODEs}  & \textbf{Application} & \textbf{Characteristics} \\ 
\toprule 
Continuous ODE-GRU-D  and Extended ODE-GRU-D \cite{2020arXiv200510693H} & Multi-Variate time series with irregular sampling rate & Leverage $\ODESOLVE$ similar to Neural-ODEs and compute hidden as well as input decay rate based on the dynamics of the data \\ \hline

\cite{fernandes2019predicting}&Detection of Heart Rate from Facial Videos & Train Neural-ODE using heart rate from original videos, and predict heart rates of deepfake videos using trained Neural-ODE.\\ \hline

\cite{sarkar2019improving} & Data Augmentation & Used the optimized parameters from a Neural ODE network to compute local gradient of the time series and later used for gereting perturbed adversarial samples. \\ 
\bottomrule
\end{tabular}
\end{table}

This new family of neural network has some substantial improvement over residual neural network as following 
\begin{itemize}
	\item Ordinary Differential Equations require one independent variable and one derivative of unknown function. The value of the independent variable at any time t can be computed by an unknown ODE solver with desired accuracy.	
	\item There are two different methods can be used for first order differential equations , e.g., the Eulers method \cite{alexander1990solving}and adjoint sensitive method. 
	\item The construction of neural network is much easier. it is no longer required to determine the number of layer and any optimizer can be used for learning. 
	\item It does not need to define the number of discrete layer, instead the network is a continuous function where the gradient can train itself within marginal error.
	\item  Most common neural network used for time series modelling are vulnarable for adversarial attack. Therefore, high robustness is essential for time series data augmentaion.  \cite{sarkar2019improving} shows that Neural ODE can generated local guided gradient which improves the robustnes of time series data augmentation significantly. 
	\item Finally, it would help to design the neural network model as function of time $t$ for continuous time learning problems. 
\end{itemize}

However, Neural ODE based neural network still has some limitations to overcome as follows

\paragraph{Less Accuracy:} The accuracy of Neural-ODE are less for long term prediction. Neuro-ODE is mainly a generative neural network. For parallel mode prediction, the past values in the series are usually not actual system output, and instead, they are the predicted value of the network. Therefore, the accuracy can be reduced over time for long term prediction. The main reason for this limitation is that a Nero-ODE network learns the state of the system over time, instead of the changing rate of system states over the training period. This causes potential obstacles for the network to learn long term behaviour of the system.
\paragraph{Training time:} Neural-ODEs are usually slower than other neural networks; at the same time, it requires longer training time.
\paragraph{High error rate:} The error rate is higher for Neural-ODE in comparison to the stranded neural network. One missing component of Neural-ODEs is various regularisation mechanisms, which are crucial for reducing generalisation error as well as improving the robustness of neural networks against adversarial attacks. The initial input for Neural SDE \cite{liu2019neural}  is the output of a convolutional layer, and the value at time t is passed to a linear classifier after average pooling. The unique characteristics of neural SDE are that it adds randomness in the continuous neural networks using a Stochastic Differential Equation (SDE) framework.

\paragraph{Difficult architecture:}The proper order of the neural identifier for identifying an ODE system is not easy
to know. 

\paragraph{Stabilization of the structure:}  ODE-GRU and ODE-LSTM \cite{2020arXiv200509807H}, GRU-ODE-Bayes \cite{de2019gru} models also contribute to the stabilization of the structure of the Neural-ODEs. Neural SDE \cite{liu2019neural} also uses stochastic noise injection in the SDE network to regularise the continuous ODEs,

\paragraph{Fixed time interval}: The existing neural identifier can only predict the system behaviour well at a fixed time interval (using a fixed, regular sampling rate). This is not the nature of an ODE system. Although a high-order discretisation is more accurate than the first-order discretisation, the resulting ordinary differential equations of the former are usually complex and intractable.

\paragraph{Informative Missingness}: Continuous ODE-GRU-D  and Extended ODE-GRU-D \cite{2020arXiv200510693H} leverage ordinary differential equation solver $(\ODESOLVE)$ to compute the hidden dynamics of the model. These models use $\ODESOLVE$ to replace the missing value in continuous time series $(T)$ by the derivative of the value of available observations of $T$. These models are evaluated against the multi-variate time series. Over time the decay in hidden dynamics, as well as input, has a significant impact on the final output on multi-variate time series. Both models compute the decay rate as the derivatives of time ($t$). Therefore, the decay rate can control the gradient optimisation of the model over time. Continuous ODE-GRU-D model use to generate the hidden dynamics of GRU-D model \cite{grud} as continuous-time dynamics using $\ODESOLVE$. Extended ODE-GRU-D model computes decay rate in addition to continuous hidden dynamics of GRU-D model and exhibits an efficient way to generate both hidden and input decay rate based on the dynamics of the data. Experiment on Physio net demonstrates that these models can successfully solve the informative missingness from continuous data. 

\paragraph{High Dimension datasets}: Neural-ODEs suffer from higher computation cost for time series with higher dimension. Unlike,black-box ODEs, $ODE^{2}VAE$ model explicitly decomposes the latent space $(\dot{\mathbf{z}}_{t})$ into momentum $(\dot{\mathbf{s}}_{t})$ and position components $\dot{\mathbf{v}}_{t}$ similar to variational auto-encoders (VAEs) and solves dynamical system governed by a second order Bayesian neural ODE model $\mathbf{f}_{\mathcal{W}}\left(\mathbf{s}_{t}, \mathbf{v}_{t}\right)$ as shown in Eq.~\eqref{eq:bode}. As first-order ODEs are incapable of modelling high-order dynamics, $ODE^{2}VAE$  uses three different components; (i) a distribution for the initial position $p(s_0)$ and velocity $p(v_0)$ in the latent space , (ii) hidden dynamics defined by an acceleration field, and (iii) a decoding likelihood $p(xi|si)$ , which is only computed by the velocity positions. .

\begin{equation}
\dot{\mathbf{z}}_{t}=(\dot{\mathbf{s}}_{t},\dot{\mathbf{v}}_{t})=\{\begin{array}{ccc}
\dot{\mathbf{s}}_{t} & = & \mathbf{v}_{t} \\
\dot{\mathbf{v}}_{t} & = & \mathbf{f}_{\mathcal{W}}\left(\mathbf{s}_{t}, \mathbf{v}_{t}\right)
\end{array}, \quad\left[\begin{array}{c}
\mathbf{s}_{T} \\
\mathbf{v}_{T}
\end{array}\right]=\left[\begin{array}{c}
\mathbf{s}_{0} \\
\mathbf{v}_{0}
\end{array}\right]+\int_{0}^{T} \underbrace{\left[\begin{array}{c}
\mathbf{v}_{t} \\
\mathbf{f}_{\mathcal{W}}\left(\mathbf{s}_{t}, \mathbf{v}_{t}\right)
\end{array}\right]}_{\overrightarrow{\mathbf{f}} w\left(\mathbf{z}_{1}\right)} d t
\label{eq:bode}
\end{equation}

\paragraph{Representation of continuous function}: \cite{dupont2019augmented} demonstrate that even some straightforward function can not be represented using  Neural-ODE as this model preserve the state of the input and follow continuous limit or boundary. This limitation often leads to expensive computation for Neural-ODE. To overcome such limitation, Augmented Neural ODEs (ANODEs)\cite{dupont2019augmented} add dimension to learn complex function using simple flow. Therefore, instead of $\mathbb{R}^{d} $ , ANODEs uses $ \mathbb{R}^{d+p}$ to move points to an additional dimension and avoid intersection among trajectories. Eq~\eqref{eq:ode1} is formulated for ANODEs as  Eq.~\eqref{eq:anode}.

\begin{equation}
\frac{\mathrm{d}}{\mathrm{d} t}\left[\begin{array}{l}
\mathrm{h}(t) \\
\mathrm{a}(t)
\end{array}\right]=\mathrm{f}\left(\left[\begin{array}{l}
\mathrm{h}(t) \\
\mathrm{a}(t)
\end{array}\right], t\right), \quad\left[\begin{array}{l}
\mathrm{h}(0) \\
\mathrm{a}(0)
\end{array}\right]=\left[\begin{array}{l}
\mathrm{x} \\
0
\end{array}\right]
\label{eq:anode}
\end{equation} 

\begin{figure}[h]
\centering\includegraphics[width=0.8\textwidth]{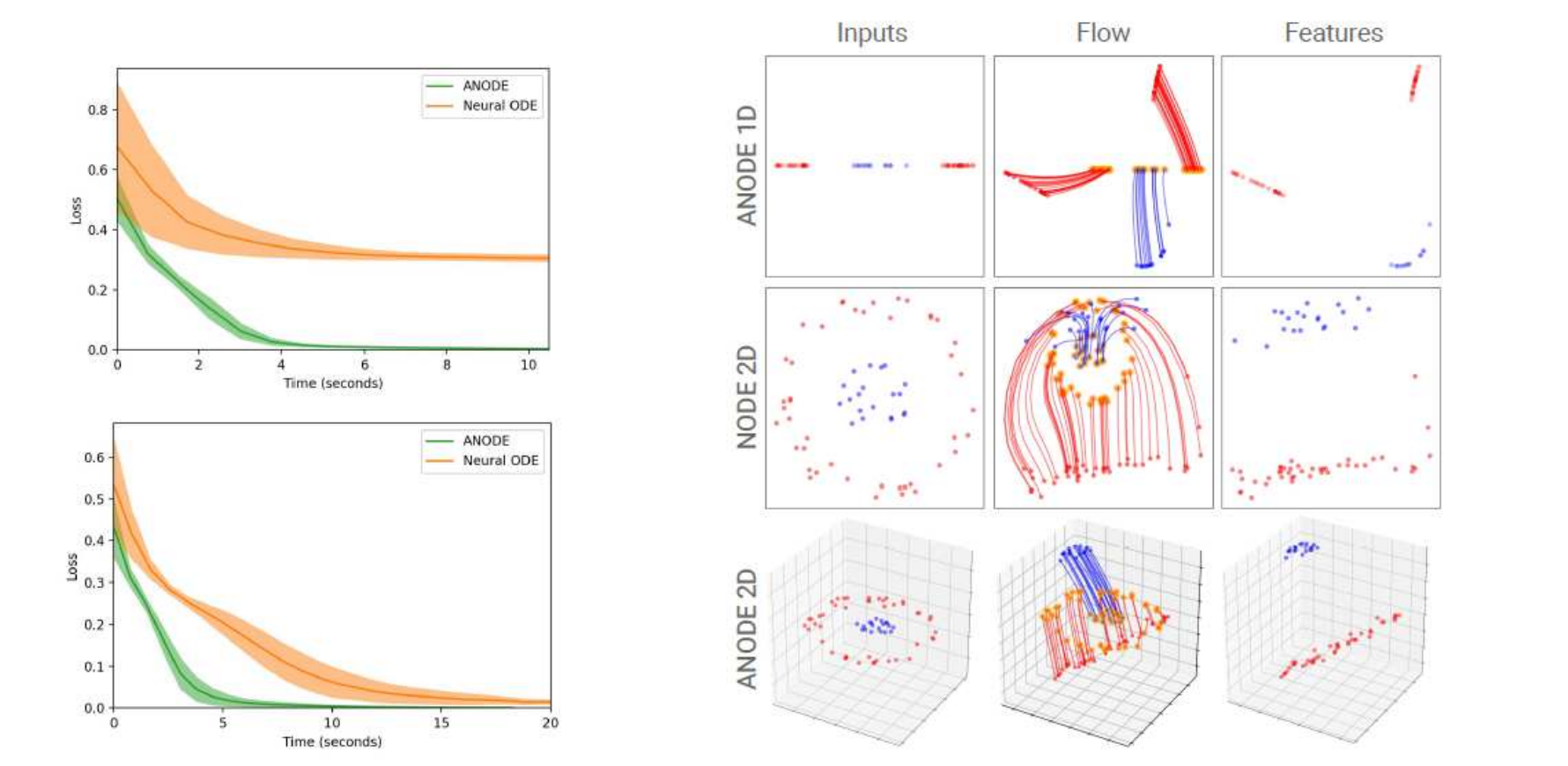}
\caption{(Left) Loss plots for Neural-ODEs and ANODEs  (Right)
Flows learned by NODEs and ANODEs.}
\label{fig:anode} 
\end{figure}

Fig.~\ref{fig:anode} describes the differences between Neural-ODE and ANODEs for the function $g(x)$. ANODEs achieve better precision for learning the functions and are consistently faster than Neural-ODEs. As ANODEs always learn simpler flow, the computation cost is also less than Neural-ODEs.  

\subsubsection{Partial Differential Equation as Neural Network}

Partial Differential Equation (PDE) contributes significantly to the advancement of the neural network. Convolution neural network (CNN) are highly appropriated in the area of image processing and computer vision. The basic architecture of a CNN model can be represented through mathematical equations. Recent works have been proposed to pre-process the image dataset as a dynamical system of continuous-time series. Therefore, a series of ODEs can be used to interpret these systems. \cite{cnnpde} shows the relation between PDE and CNN. Convolution neural network (CNN) are highly appropriated in the area of image processing and computer vision. For example, RKNN \cite{wang1998runge} as described in section \ref{sec:rknet} has some strong characteristics along with some limitations. In order to use  RKNN for a system like an image classification, it requires some adjustment. For example, each time period is modelled by a time-dependent first order dynamical system. Besides, all coefficient and time step size ($T$) can be computed implicitly. The time step size ($h$) of RK methods as shown in \eqref{eq:2} needs to be predefined in order to control the truncation error.  With these modifications, RKNet \cite{zhu2018convolutional} is proposed. 
\begin{equation}
\label{eq:Xn2}
X = {x_{1},x_{2}, ... ... ... , x_{n}} \in \mathbb{R}^n
\end{equation}

The main characteristics of RKNet are as following 

\begin{itemize}
	\item In an RKNet, a period is composed of iterations of time-steps.
	\item A particular RK method is adopted
	throughout the time-steps in a period to approximate the increment in each step
	\item The increment in
	each step is broken down to the increments in several stages according to the adopted RK method.
	\item Each stage is approximated by a convolutional subnetwork due to the versatility of neural networks on approximation.
	
\end{itemize} 

For PDE based CNN, the neural network function $f$ filters the input feature Y as shown in Eq.~\ref{eq:5}. Here $f$ consists of layers (F) of liner transformations and point wise non-linearities.

The core layer of deep neural network architecture can be simplified as described as Eq.~\eqref{eq:4}. Any deep neural network can be formalized as concatenation of many of the layers (F) given in Eq.~\eqref{eq:4}. The linear operations $ K_{2} $ and $K_{1}$ in Eq.~\eqref{eq:4}. Each of these two operators is parametrized $\theta^3 $ and $ \theta^1 $  respectively. \textit{N} is the  normalization layer parametrized with  $ \theta^2 $. A common choice for N in (1) is the batch normalization
layer.  $ \theta^2 $ represents the scaling factors and biases. of a neural network  For activation function any activation method can be used, e.g. $ \sigma $.  
\begin{equation}
\label{eq:4}
F(\theta, Y) = K_{2}(\theta^3)\sigma(\textit{N}(K_{1}(\theta^1)Y, \theta^2)
\end{equation}   

Most of the cases, the input Y is a time despondent data and can be represented as can be seen as a discretization of a continuous function $ Y (x) $. For example, time series forecasting, weather prediction, image data speech and other applications \cite{cnnpde} has modified Eq.~\eqref{eq:4}.  The linear operators $ K_{1} = K_{1} (\theta) \epsilon R^(wxw_{in})$  and $ K_{2} = K_{2} (\theta) \epsilon R^(w_{out}xw) $ are considered as convolution operators as shown in Eq.~\eqref{eq:5}. 

\begin{equation}
\label{eq:5}
\mathbf{F}_{\mathrm{sym}}(\boldsymbol{\theta}, \mathbf{Y})=-\mathbf{K}(\boldsymbol{\theta})^{\top} \sigma(\mathcal{N}(\mathbf{K}(\boldsymbol{\theta}) \mathbf{Y}, \boldsymbol{\theta}))
\end{equation}

In this section, we discuss Physics Informed Neural Network, which use a combination of neural networks and partial differential equation(PDE) to obtain the hidden dynamic of the model. In this model, we should consider a penalty for the loss function of a neural network that deviates from the dynamic of the model. In \cite{raissi2017physics}, authors introduced Physics Informed Neural Network to obtain not only the solution of the PDE but also obtain the parameters of the PDE for an optimization task. 
Assume we want to obtain the solution of a PDE by a neural network. Here there is no training set. By converting the PDE to an optimization problem, you can solve the PDE by a neural network using initial conditions and boundary value condition and use the new solutions as the training set of the neural network. This means as the new value of approximation of the solution obtained, a neural network used them to learn and produce the new solution and so on. 

We consider the general form of a class of PDE in the form:
\begin{equation}
\left\{
\begin{aligned}
&u_t(x,t)+\mathcal{N}[u(x,t),\lambda]=0,x\in\Omega\subset \mathbb{R}^N, t\in [0,T]\\
&u(x,0)=g(x), x\in \Omega\text{  Initial Condition}\\
&u(x,t)=h(x,t), x\in \partial \Omega\text{ Boundary Value Condition}\\
&\frac{\partial u}{\partial n}(x,t)=k(x,t), x\in\partial \Omega\text{ Boundary Value Condition} 
\end{aligned}
\right.
\label{pde}
\end{equation}

where $\mathcal{N}$ is a nonlinear differential operator, $\lambda$ is the parameter, $\partial \Omega$ is the boundary of $\Omega$, and boundary condition is the Cauchy boundary condition. In this setting, $g:\Omega\subset \mathbb{R}^n\to \mathbb{R}$, $h:\partial\Omega\times \mathbb{R}\to\mathbb{R}$ and $k:\partial\Omega\times \mathbb{R}\to\mathbb{R}$ are known function and we want to find unknown function $u:\Omega\times \mathbb{R}\to \mathbb{R}$.

Assume there is a neural network $neural\_net$ that approximate the function $u(x,t)$, i.e., on input $(x,t)$ it compute approximation of $u(x,t)$. So if we assume $\hat{u}=neural\_net(x,t)$ then by reverse mode Automatic Differentiation we could compute partial derivatives of $\hat{u}$ with respect to $t$ and $x$, \cite{PEARLMUTTER_LTUB}. But, what is the loss function? Our goal is to find an approximation of $u$ such that it is the solution of the Eq~\eqref{pde}. So,
\begin{equation}
\hat{u}_t+\mathcal{N}[\hat{u},\lambda]
\end{equation}
is not exactly zero. Thus we need to minimize the following function,
\begin{equation}
Minimize [\hat{u}_t+\mathcal{N}[\hat{u},\lambda]]^2
\end{equation}
since $\hat{u}$ has been obtain by weights and biases, we need to find optimum value of weights and biases to find the solution. But we need also our approximation respect the initial condition and boundary value conditions, so,
\begin{equation}
\begin{aligned}
L(\theta,x,t,\lambda)=&[\hat{u}_t+\mathcal{N}[\hat{u},\lambda]]^2+[\hat{u}(x,0)-g(x)]^2+\\
&\chi_{\partial \Omega}(x).([\hat{u}(x,t)-h(x,t)]^2+[\frac{\partial \hat{u}}{\partial n}(x,t)-k(x,t)]^2)
\end{aligned}
\end{equation}

where $\theta$ is the parameter of the neural network and $\chi_{\partial\Omega}$ is the characteristic function of $\partial \Omega$, i.e.,
\begin{equation}
\chi_{\partial\Omega}(x)=
\left\{
\begin{aligned}
& 1\, \text{ if }x\in\partial \Omega \\
& 0\, \text{ otherwise}
\end{aligned}
\right.
\end{equation}

The loss function is not just depend on the parameter of the neural network, it also depend on the space and time. So for every instant $(x,t)$ we have a loss function. Suppose we fixed the parameter $\lambda$. Let $N_b$ be the number of boundary and initial training data and $N_c$ be the number of collocation points then the cost function is as follow:
\begin{equation}
MSE=MSE_b+MSE_c
\end{equation}
where,
\begin{equation}
MSE_b=\frac{1}{N_b}\sum_{i=1}^{N_b}[\hat{u}(x_i^b,0)-g(x_i^b)]^2+\chi_{\partial \Omega}(x_i^b).([\hat{u}(x_i^b,t_i^b)-h(x_i^b,t_i^b)]^2+[\frac{\partial \hat{u}}{\partial n}(x_i^b,t_i^b)-k(x_i^b,t_i^b)]^2)
\end{equation}
and
\begin{equation}
MSE_c=\frac{1}{N_c}\sum_{i=1}^{N_c}[\hat{u}_t(x_i^c,t_i^c+\mathcal{N}[\hat{u}(x_i^c,t_i^c,\lambda]]^2
\end{equation}
  
\begin{figure}[h]
\includegraphics[width=0.5\linewidth]{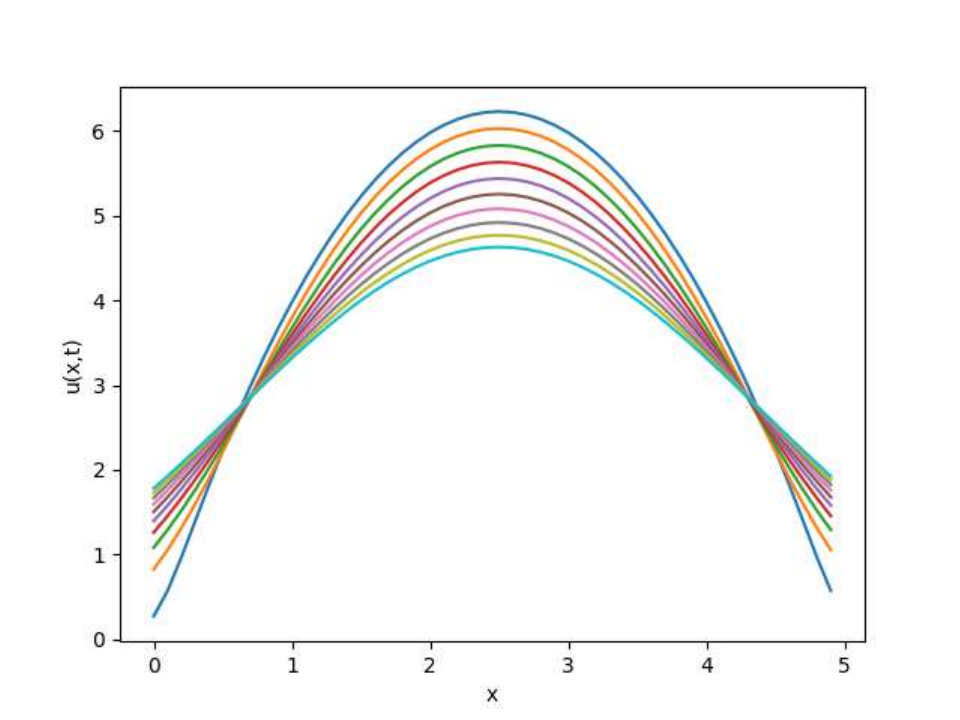}
\includegraphics[width=0.5\linewidth]{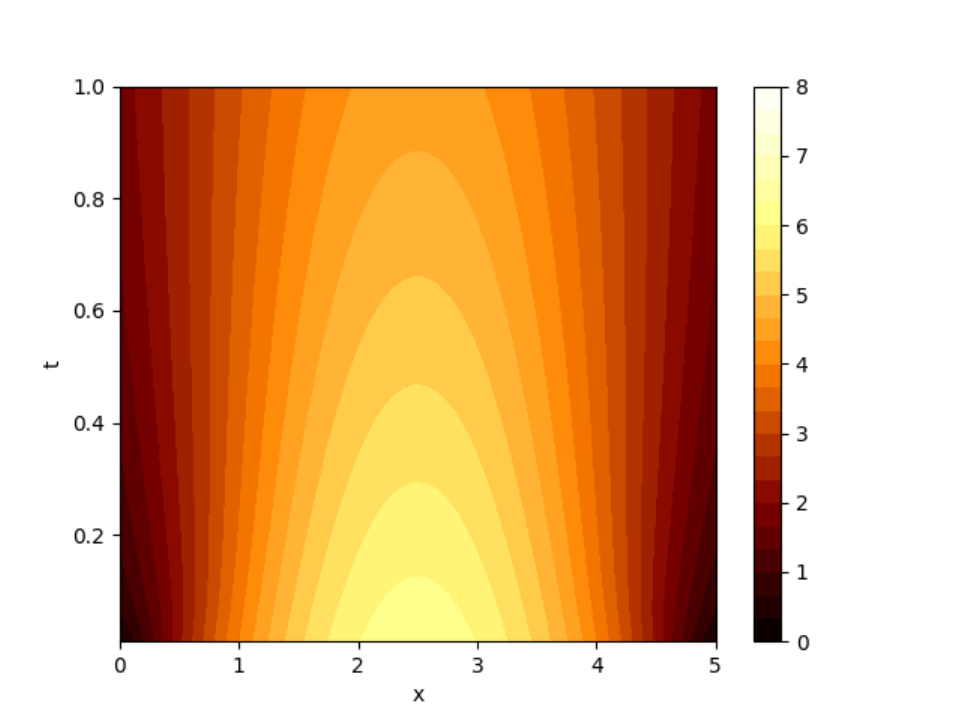}
\caption{Heat equation on $[0,5]\times[0,1]$}
\label{fig:heat}
\end{figure}

As an example consider one-dimentional heat equation with homogeneous Dirichlet boundary conditions on $(x,t)\in [0,5]\times [0,1]$
\begin{equation}
\left\{
\begin{aligned}
&u_t=u_{xx}\\
&u(x,0)=x(5-x)\\
&u(0,t)=u(5,t)=0
\end{aligned}
\right.
\end{equation}

we know the analytic solution of this equation is as follow,
\begin{equation}
u(x,t)=\frac{1}{\sqrt{4\pi t}}\int_0^5e^{-\frac{(x-y)^2}{4t}}y(5-y)dy
\end{equation}
the solution and the evolution of initial condition is depicted in Fig.~\ref{fig:heat}.

If we represent $MSE_0$ as mean squar error of the initial condition, $MSE_b$ the mean squar error of the boundary points and $MSE_c$ as the mean squar error of the collocation points then we have,
\begin{equation}
MSE_0=\frac{1}{N_0}\sum_{i=1}^{N_0}(\hat{u}(x_0^i,0)-x_0^i(5-x_0^i))^2
\end{equation}
where $\{x_0^i\}_{1\leq i\leq N_0}$ are training data of initial condition in $[0,5]$.
\begin{equation}
MSE_b=\frac{1}{N_b}\sum_{i=1}^{N_b}(|\hat{u}(0,t_b^i)|^2+|\hat{u}(5,t_b^i)|^2)
\end{equation}
where $\{t_b^i\}_{1\leq i\leq N_b}$ are training data of boundary conditions in $[0,1]$.
\begin{equation}
MSE_c=\frac{1}{N_c}\sum_{i=1}^{N_c}(\hat{u}_t(x_c^i,t_c^i)-\hat{u}_{xx}(x_c^i,t_c^i))^2
\end{equation}
where $\{(x_c^i,t_c^i)\}_{1\leq i\leq N_c}$ are training data of the collocation points in $[0,5]\times [0,1]$.

We know we can compute partial derivative of a neural network by reverse mode automatic differentiation, \cite{PEARLMUTTER_LTUB}, i.e.,
\begin{equation}
\begin{aligned}
&\hat{u}(x^i,t^i)=neural\_net(x^i,t^i)\\
&\hat{u}_t(x^i,t^i)=\frac{\partial}{\partial t}(neural\_net(x^i,t^i))\\
&\hat{u}_{xx}(x^i,t^i)=\frac{\partial^2}{\partial x^2}(neural\_net(x^i,t^i))
\end{aligned}
\end{equation}
so the cost function of the heat equation with the given initial and boundary condition is as follow;
\begin{equation}
MSE(\theta)=MSE_0+MSE_b+MSE_c
\end{equation}
by minimizing $MSE(\theta)$ with respect to $\theta$ we could find the learning parameter of the heat equation.

\section{Future Direction for Potential Research Area}
\label{sec:future}
Therefore, the main concern of this survey work is “How can we ensure efficient event-based sensor’s signal processing that generates an asynchronous data-driven output on a continuous-time sequence? “   
To have a better solution for this raised question, we also need to answer some other questions:
\begin{itemize}
	\item	How the existing neural network model can be improved to provide a solution. 
	\item	How does open data helps in risk assessment, fault tolerance, failure detection, real-time accuracy? 
	\item	How can output be updated asynchronously to be mapped with data from one or multiple events on a continuous-time series? 
	\item	How energy consumption and computation time can be optimised?
\end{itemize}
The main aim of the proposed work is to define a recurrent neural network-based (RNN) deep learning model for processing and modeling event-based sensor signal with a short and dynamic time gap in continuous time series and generate an asynchronous output at higher frequencies to deal with the following challenges:

\begin{itemize}
	\item	To reduce the high amount of computation for the short time gap in data sampling. 
	\item	To design the LSTM model using an ordinary differential equation solver.  
	\item	To develop asynchronous output in continuous time series. Asynchronous output can be beneficial for learning continuous series data generated by event-based sensors of IoT devices. For example, sensors in an autonomous vehicle are synthesised and processed in real-time with more accuracy. 
	\item	To immolate human brain work-flow to generating asynchronous and event-triggered a spike in continuous time.
\end{itemize}

Recent emerging technologies, such as Internet of things (IoT), cloud computing,  edge computing and 5G are using continuous-time data analysis and time-frequency analysis as a core fundamental technology for the corresponding ecosystem. For example, different cloud providers are extensively using deep learning algorithm for processing continuous-time data for monitoring performance and activity within a software-defined network (SDN), dynamic resource allocation, scaling up or down the memory and power of compute engine and other resources, activity tracking for security and privacy purpose. Table \ref{tab:emmerging-tech}  shows the use of Deep learning algorithm for continuous-time data analysis in different emmerging technology.

\begin{table}
	\centering
	\caption{Different Projects in Health care with time series}
	\label{tab:emmerging-tech}
	\begin{tabular}{|p{2.9in}|p{2.9in}|} \hline 
		\textbf{Applications} & \textbf{Popular neural Network}   \\ \hline 
		Resource Scheduling & LSTM\cite{yan2019intelligent}     \\ \hline 
		Edge Intelligence &  RF \cite{polese2020machine}    \\ \hline 
		Software defined Network (SDN)  & DNN\cite{sun2015intelligent}     \\ \hline 
		Self-protective framework & DNN\cite{vidal2019framework}    \\ \hline 
		Network Management & CNN\cite{bega2019deepcog}    \\ \hline
	\end{tabular}
\end{table}

\section{Conclusion} \label{sec:conclusion}

In this paper, we have reviewed several recent research works focusing on time series modeling and prediction using a neural network and providing the mathemaical background of these models. We have investigated all different challenges in the case of continuous-time series modeling with neural networks.
In this paper, we have reviewed recent works mainly focused on dealing with the different challenges faced in modeling the continuous-time series using deep learning. We investigate several promising types of research that have established the foundation to introduce a new paradigm for modeling continuous-time series overcoming most of the associated challenges. In recent days, there is an enormous attention to the sensory event containing temporal information. Events happening in continuous time for the different real-time autonomous scenario are significant for different research practitioner. For example, an automobile vehicle with different types of sensor can collect information from the context continuously in real-time. Most cases, the information is a sequence of events that occurs in either a long or short time duration. Events occurred in continuous time are interrelated by either longer-term or short-term dependency. This dependency needs to be recognised, processed, stored and analysed in order to predict the next event with similar behaviour correctly. For example, a sequence of events can detect the lane change behaviour of the driver or detect potential system failure or accident.  There are also much other similar research areas as like mobile sensor information, where the temporal sequence plays a significantly important role. 
Research with time series is not a new thing. Different applications are using continuous time series classification, prediction and generation. However, in most cases, in order to avoid the complexity due to the massive length of series or higher dimensions, real-time data are presented as a discrete-time model with the fixed sampling rate. This cause lost in precision and accuracy of the result. In that context, most existing NN model is ill-suited to model a real-time continuous time series with higher dimension and a long length. The continuous-time series modeling is one of the most promising research areas in the field of deep learning for future research.
Based on the comparative analysis, some of the significant research areas are as follows

\begin{itemize}
	\item	Sampling irregular data
	\item	Reduce energy consumption 
	\item	Reduce the amount of computation
	\item	Optimize performance
	\item	Improve memory capacity
	\item	Modelling long time sequence
	\item	Efficient feature extracting 
	\item	Represent time sequence as a differentiable function 
	\item	Implementation of the deep learning algorithm in mobile sensor data 
	\item	Reduce the imbalance between positive and negative event
	\item	Optimise overlapping events
\end{itemize}

\bibliographystyle{unsrt}
\bibliography{referecenes.bib}

\end{document}